\documentclass[journal]{IEEEtran}
\usepackage{multirow}
\usepackage{amsmath,graphicx}
\usepackage{algpseudocode}
\usepackage{multirow}
\usepackage{array}
\usepackage{colortbl}
\usepackage{booktabs}
\usepackage{rotating}
\usepackage{caption}
\usepackage{subcaption}
\usepackage[normalem]{ulem} 
\usepackage{gensymb}
\usepackage{pifont}
\usepackage{makecell}

\begin{document}

\title{Robust Uncalibrated Stereo Rectification with Constrained Geometric 
Distortions (USR-CGD)}
\author{Hyunsuk Ko, Han Suk Shim, Ouk Choi,~\IEEEmembership{Member,~IEEE,}
and~C.-C.~Jay~Kuo,~\IEEEmembership{Fellow,~IEEE}}% <-this % stops a space
%\thanks{Manuscript received November XX, 2014.}
%\thanks{Hyunsuk Ko, Han Suk Shim and C.-C. Jay Kuo are with Ming Hsieh Department of
%Electrical Engineering, Signal and Image Processing Institute,
%University of Southern California, Los Angeles, CA 90089, USA (e-mails:
%kosu9980@gmail.com; cckuo@sipi.usc.edu).}%
%\thanks{Ouk Choi is with Samsung Advanced Institute of Technology, 
%Gyeonggi-Do 446-712, Korea (e-mail: ouk.choi@samsung.com).}}%

%\markboth{IEEE TRANSACTIONS ON IMAGE PROCESSING,~Vol.~XX, No.~X, XXXXXX~2014}%
%{KO \MakeLowercase{\textit{et al.}}: ParaBoost to Stereo Image Quality Assessment}

\maketitle

\begin{abstract}
A novel algorithm for uncalibrated stereo image-pair rectification under
the constraint of geometric distortion, called USR-CGD, is presented in
this work.  Although it is straightforward to define a rectifying
transformation (or homography) given the epipolar geometry, many
existing algorithms have unwanted geometric distortions as a side
effect. To obtain rectified images with reduced geometric distortions
while maintaining a small rectification error, we parameterize the
homography by considering the influence of various kinds of geometric
distortions. Next, we define several geometric measures and incorporate
them into a new cost function for parameter optimization. Finally, we
propose a constrained adaptive optimization scheme to allow a balanced
performance between the rectification error and the geometric error.
Extensive experimental results are provided to demonstrate the superb
performance of the proposed USR-CGD method, which outperforms existing
algorithms by a significant margin. 
\end{abstract}


\section{Introduction}\label{sec:intro}

\IEEEPARstart{S}{tereo} vision is used to reconstruct the 3D structure
of a real world scene from two images taken from slightly different
viewpoints. Recently, stereoscopic visual contents have been vastly used
in entertainment, broadcasting, gaming, tele-operation, and so on. The
cornerstone of stereo analysis is the solution to the correspondence
problem, which is also known as stereo matching. It can be defined as
locating a pair of image pixels from left and right images, where two
pixels are the projections from the same scene element. A pair of
corresponding points should satisfy the so-called epipolar constraint.
That is, for a given point in one view, its corresponding point in the
other view must lie on the epipolar line.  However, in practice, it is
difficult to obtain a rectified stereo pair from a stereo rig since it often
contains parts of mechanical inaccuracy. Besides, thermal dilation
affects the camera extrinsic parameters and some internal parameter. For
example, the focal length can be changed during shooting with different
zoom levels. For the above-mentioned reasons, epipolar
lines are not parallel and not aligned with the coordinate axis and,
therefore, we must search through the 2D window, which makes the
matching process a time-consuming task. 

The correspondence problem can however be simplified and performed
efficiently if all epipolar lines in both images are in parallel with
the line connecting the centers of the two projected views, which is
called the baseline. Since the corresponding points have the same
vertical coordinates, the search range is reduced to a 1D scan line.
This can be embodied by applying 2D projective transforms, or
homographies, to each image. This process is known as ``image
rectification" and most stereo matching algorithms assume that the left
and the right images are perfectly rectified. 

If intrinsic/extrinsic camera parameters are known from the calibration
process, the rectification process is unique up to trivial
transformations.  However, such parameters are often not specified and
there exists a considerable amount of uncertainty due to faulty
calibration. On the other hand, there are more degrees of freedom to
estimate the homography of uncalibrated cameras. The way to parameterize
and optimize these parameteres is critical to the generation of
rectified images without unwanted warping distortions. 

As pioneering researchers on image rectification, Ayache and
Francis~\cite{cit:Ayache1991} and Fusiello {\it et al.}
\cite{cit:Fusiello2000} proposed rectification algorithms based on known
camera parameters. The necessity of knowing calibration parameters is
the major shortcoming of these early methods. To overcome it, several
researchers proposed a technique called projective (or uncalibrated)
rectification that rectifies images without knowing camera parameters
but estimating homography using epipolar geometry under various
constraints.  Hartley and Andrew~\cite{cit:Hartley1999, cit:Hartley2003}
developed a theoretical foundation for this technique, where they
derived a condition for one of two homographies to be close to a rigid
transformation while estimated the other by minimizing the difference
between the corresponding points. Loop and Zhang \cite{cit:Loop1999}
estimated homographies by decomposing them into a projective transform
and an affine transform. 

There exist rectification algorithms that take geometric distortions
into account to prevent them from being introduced in the rectified
images. For example, Pollefeys~{\it et al.} \cite{cit:Pollefeys1999}
proposed a simple and efficient algorithm for a stereo image pair using
a polar parametrization of the image around the epipole while Gluckman
and Nayar~\cite{cit:Gluckman2001} presented a rectification method to
minimize the resampling effect, which corresponds to the loss or
creation of pixels due to under- or over-sampling, respectively.  A
similar approach was developed by Mallon and Whelan
in~\cite{cit:Mallon2005}, where perspective distortions are reduced by
applying the singular vector decomposition (SVD) to the first order
approximation of an orthogonal-like transfrom.  Isgro and
Trucco~\cite{cit:Isgrò1999} proposed a new approach to estimate
homographies without explicit computation of the fundamental matrix by
minimizing the disparity as done in~\cite{cit:Hartley1999}.  However,
such a constraint on the disparity sometimes generates significantly
distorted image. For further improvement, Wu and Yu \cite{cit:Wu2005}
combined this approach with a shearing transform to reduce the
distortion. 

More recently, Fusiello and Luca~\cite{cit:Fusiello2011} proposed a
Quasi-Euclidean epipolar rectification method that approximates the
Euclidean (calibrated) case by enforcing the rectifying transformation
to be a collineation induced by the plane at the infinity, which does
not demand a specific initialization step in the minimization process.
Zilly {\it et al.} \cite{cit:Zilly2010} proposed a technique to estimate
the fundamental matrix with the appropriate rectification parameter
jointly.  However, their method is limited to the case with almost
parallel stereo rigs and a narrow baseline. Georgiev {\it et
al.}~\cite{cit:Georgiev2013} developed a practical rectification
algorithm of lower computational cost, yet it is only applicable in the
probable stereo setup. 

In this work, we propose a novel rectification algorithm for
uncalibrated stereo images, which demands no prior knowledge of camera
parameters. Although quite a few methods are proposed to reduce unwanted
warping distortions in rectified images with different homography
parameterization schemes, there is no clear winner among them.
Additionally, only two geometric measures (namely, orthogonality and the
aspect-ratio) are used as geometric distortion criteria while they are
not sufficient in characterizing the subjective quality of all rectified
images. Here, we analyze the effect of various geometric distortions on
the quality of rectified images comprehensively, and take them into
account in algorithmic design.  The proposed USR-CGD algorithm minimizes
the rectification error while keeping errors of various geometric
distortion types below a certain level. 

The main contributions of this work are summarized below. 
\begin{itemize}
\item An uncalibrated stereo rectification algorithm is proposed to
minimize the rectification error with constrained geometric distortions.
A variety of geometric distortions such as the aspect-ratio, rotation,
skewness and scale-variance are introduced and incorporated in the 
new cost function, then it is minimized by our novel optimization scheme.
\item A parameterization scheme for rectifying transformation is
developed. The parameters include the focal length difference between
two cameras, the vertical displacement between optical centers, etc.
This new scheme helps reduce the rectification error by adding more
degrees of freedom to the previous Euclidean
model~\cite{cit:Fusiello2011}. 
\item We provide a synthetic database that contains six geometric
distortion types frequently observed in uncalibrated stereo image pairs.
It allows a systematic way to analyze the effect of geometric
distortions and parameterize the rectifying transformations accordingly.
We also provide a real world stereo database with various indoor and outdoor
scenes of full-HD resolution. The performance of several algorithms
can be easily evaluated with these two databases.
\end{itemize}

The rest of this paper is organized as follows.  The mathematical
background on uncalibrated rectification is reviewed in
Section~\ref{sec:background}. The proposed USR-CGD algorithm is
elaborated in Section~\ref{sec:proposed}.  Two existing and two new
databases for unrectified stereo images are described in Section
\ref{sec:databases}.  Experimental results and discussions are provided
in Section \ref{sec:performance}.  The superiority of the proposed
USR-CGD algorithm are demonstrated by extensive experiments in the four
above-mentioned databases. Finally, concluding remarks are given in
Section \ref{sec:conclusions}. 

\section{Uncalibrated Rectification}\label{sec:background}

\begin{figure}[t]
\centering
\includegraphics[width=0.40\textwidth]{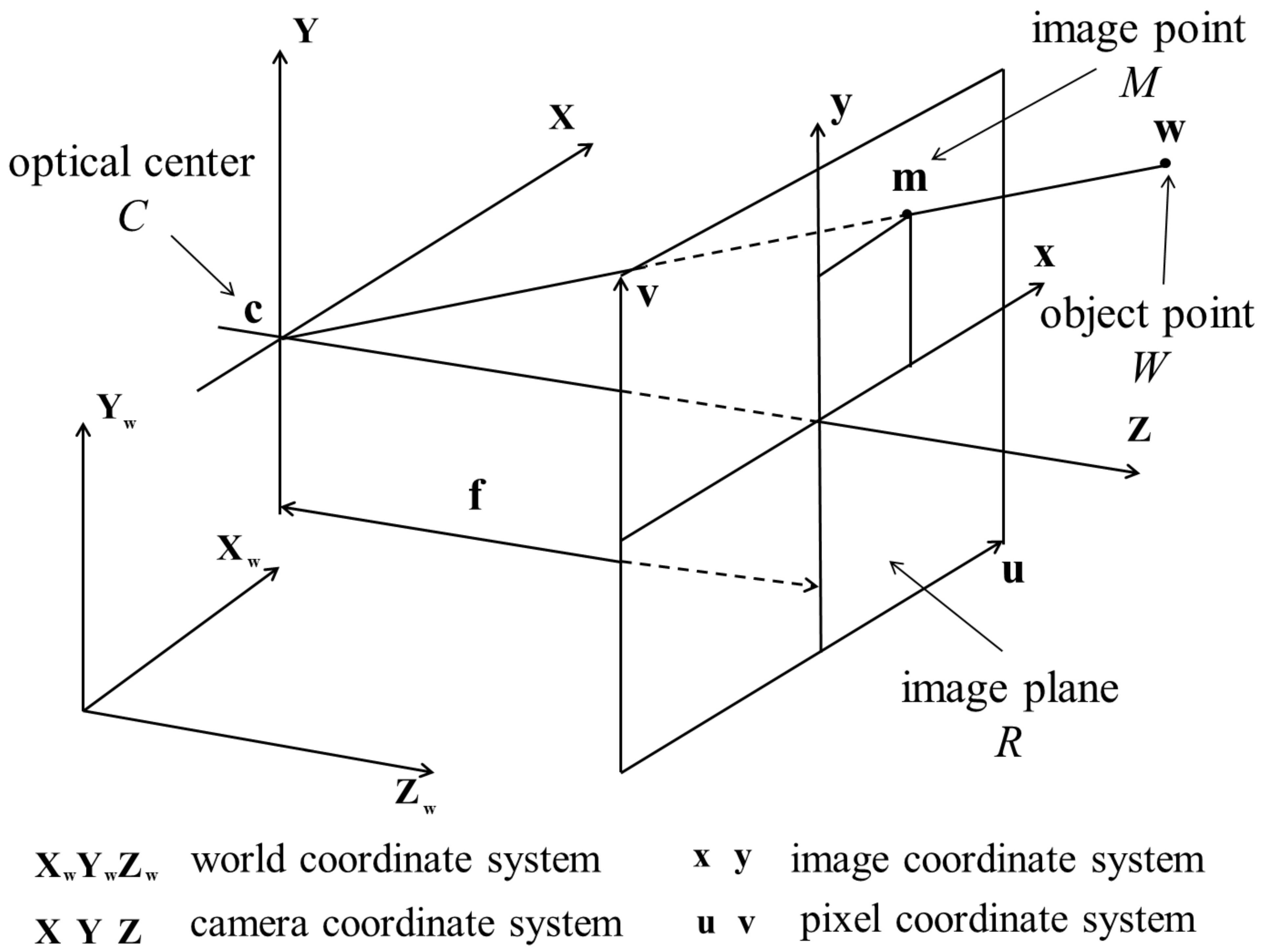}
\caption{Illustration of a pinhole camera model.}\label{fig:Pinhole}
\end{figure}

We briefly review the mathematical background on perspective projection
and epipolar geometry for uncalibrated rectification in this section.
As shown in~Fig.~\ref{fig:Pinhole}, the pinhole camera model consists of
optical center $C$, image plane $R$, object point $W$, and image point
$M$ that is the intersection of $R$ and the line containing $C$ and $W$.
The focal length is the distance between $C$ and $R$, and the optical
axis is the line that is orthogonal to $R$ and contains $C$, where its
intersection with $R$ is the principal point. 

%%%%%%%%%%%%%%%%%%%%%%%%%%%%%%%%%%%%%%%%%%%%%%%%%%%%%%%%%%
%%%%%%%%%%%%%%%%%%%%%%%%%%%%%%%%%%%%%%%%%%%%%%%%%%%%%%%%%%

Let $\mathbf{w}$ and $\mathbf{m}$ be the coordinates of $W$ and $M$,
respectively. They are related by a perspective projection matrix 
$\mathbf{P}$ of dimension $3\times4$ as
\begin{eqnarray} \label{eq:ES1}
\mathbf{m} = 
\left[ \begin{array}{ccc}
u\\
v\\
1
\end{array} \right]
&\simeq& 
\left[ \begin{array}{cccc}
p_{11} & p_{12} & p_{13} & p_{14} \\
p_{21} & p_{22} & p_{23} & p_{24} \\
p_{31} & p_{32} & p_{33} & p_{34}
\end{array} \right]
\left[ \begin{array}{cccc}
x\\
y\\
z\\
1
\end{array} \right] \\
& = & \mathbf{P}\mathbf{w},
\end{eqnarray}
where $\simeq$ indicates the equal up to scale. Matrix $\mathbf{P}$ can
be decomposed into
\begin{equation} \label{eq:ES1}
\mathbf{P}  = \mathbf{K}\left[ \mathbf{R} \mid \mathbf{t}\right],
\end{equation}
where $\mathbf{K}$ and $\left[ \mathbf{R} \mid \mathbf{t}\right]$ are
called the camera intrinsic matrix and the camera extrinsic matrix,
respectively. Matrix $\mathbf{K}$ is in form of
\begin{equation} \label{eq:ES1}
\mathbf{K} = 
\left[ \begin{array}{ccc}
\alpha_{u} & \gamma & u_{0} \\
0 & \alpha_{v} & v_{0} \\
0 & 0 & 1
\end{array} \right],
\end{equation}
where $\alpha_{u}=s_uf$ and $\alpha_v=s_vf$ are focal lengths in the $u$
and $v$ axes, respectively ($f$ is the physical focal length of the
camera in the unit of millimeters while $s_u$ and $s_v$ are the scale
factors), and $(u_{0}, v_{0})$ are the coordinates of the principal
point, $\gamma$ is the skew factor when the $u$ and the $v$ axes of the
model are not orthogonal. For simplicity, it is often assumed that the
horizontal and vertical focal lengths are the same and there is no skew
between $u$ and $v$ axes. Thus, we have
\begin{equation} \label{eq:ES1}
\mathbf{K} = 
\left[ \begin{array}{ccc}
\alpha & 0 & \frac{w}{2} \\
0 & \alpha & \frac{h}{2} \\
0 & 0 & 1
\end{array} \right],
\end{equation}
where $w$ and $h$ are the width and the height of the image,
respectively. The camera extrinsic matrix of dimension $3 \times 4$ is
concerned with camera's position and orientation. It consists of two
parts: a rotation matrix $\mathbf{R}$ of dimension $3\times 3$ and a
displacement vector $\mathbf{t}$ of dimension $3\times 1$. 

The plane that contains optical center $C$ and is parallel to the image
plane is the focal plane. According to~\cite{cit:Fuageras1993}, the
Cartesian coordinates $\mathbf{\tilde{c}}$ of $C$ is given by
\begin{equation}\label{eq:ES1}
\mathbf{\tilde{c}} = \mathbf{R}^{-1}\mathbf{t}.
\end{equation}
Then, any optical ray that passes through $M$ and $C$ can be represented
by the set of points $\mathbf{w}$:
\begin{equation}\label{eq:opticalray}
\mathbf{\tilde{w}} = \mathbf{\tilde{c}} +\alpha\mathbf{R}^{-1} \mathbf{K}^{-1} \mathbf{m},
\end{equation}
where $\alpha$ is a constant.

%%%%%%%%%%%%%%%%%%%%%%%%%%%%%%%%%%%%%%%%%%%%%%%%%%%%%%%%%%
\begin{figure}[t]
\centering
\includegraphics[width=0.35\textwidth]{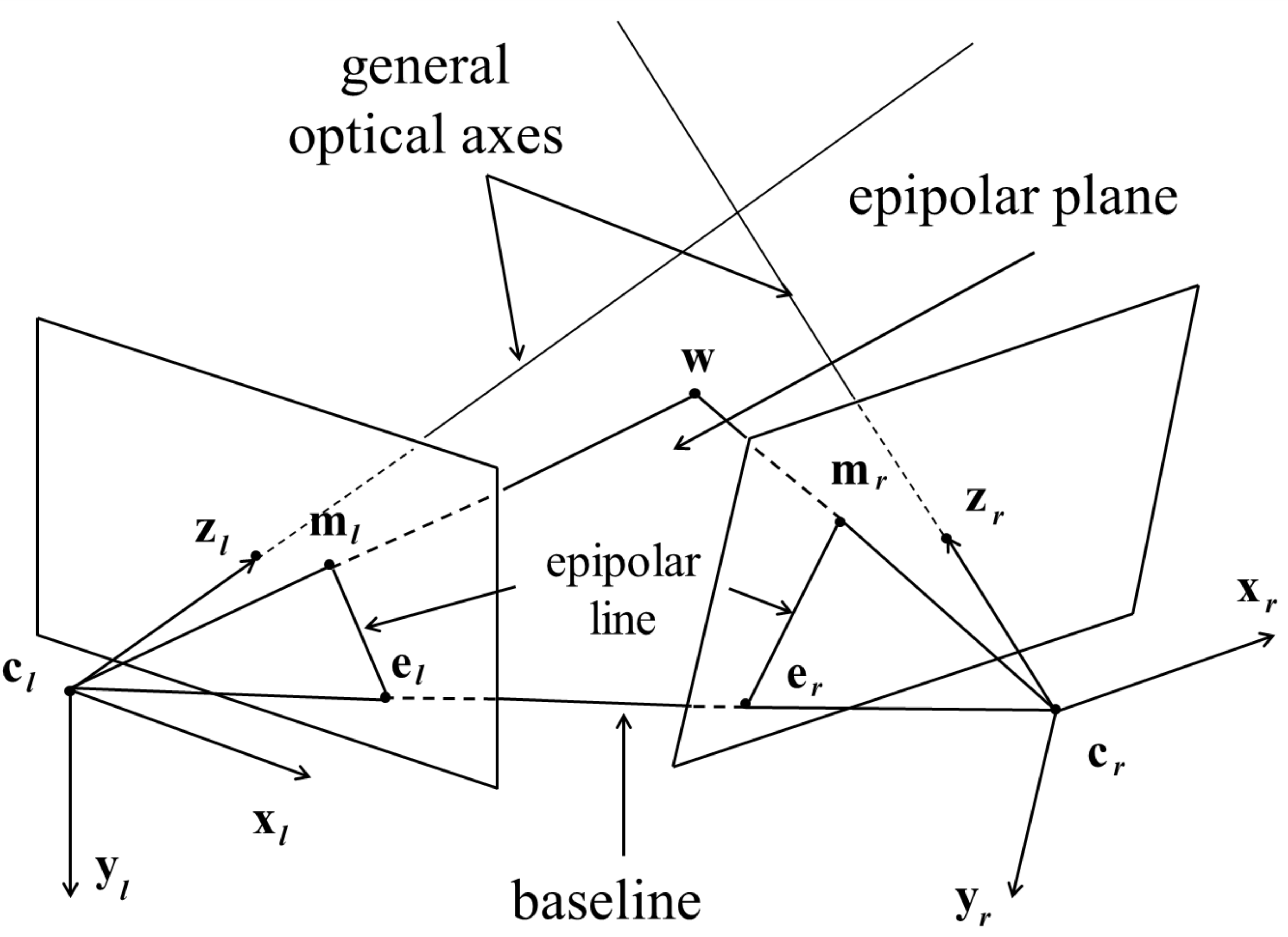}
\caption{The epipolar geometry of a pair of stereo images.}\label{fig:epipolar}
\end{figure}
%%%%%%%%%%%%%%%%%%%%%%%%%%%%%%%%%%%%%%%%%%%%%%%%%%%%%%%%%%

Next, we consider two stereo pinhole cameras as shown in Fig.
\ref{fig:epipolar}. Let $\mathbf{m_l}$ and $\mathbf{m_r}$ be point
correspondences that are the projections of the same 3D object point
$\mathbf{w}$ on images $I_l$ and $I_r$, respectively. $\mathbf{e_{l}}$
and $\mathbf{e_{r}}$ are called epipoles that are intersection points of
the baseline with the left and right image planes. The plane containing
the baseline and object point $\mathbf{w}$ is called the epipolar plane,
and the intersection lines between the epipolar plane and each of the
two image planes are epipolar lines. 

The intrinsic projective geometry between the corresponding points in the 
left and right images can be described by the epipolar constraint as
\begin{equation} \label{eq:epipolargeometry}
\mathbf{m}_{l}^{T}\mathbf{F} \mathbf{m}_{r} = \mathbf{m}_{r}^{T}
\mathbf{F}^{T} \mathbf{m}_{l} = \mathbf{0},
\end{equation}
where $\mathbf{F}$ is the fundamental matrix, which is a $3 \times 3$ 
matrix with rank 2, and $\mathbf{0} = \left[0~0~0\right]^T$ is a zero column 
vector. The epipole, which is the null space of $\mathbf{F}$, satisfies
the following condition:
\begin{equation} \label{eq:epipole}
\mathbf{F} \mathbf{e}_{l} = \mathbf{F}^{T}\mathbf{e}_{r} = \mathbf{0}.
\end{equation}
Fundamental matrix $\mathbf{F}$ maps a point, $\mathbf{m_l}$, in one
image to the corresponding epipolar line, $\mathbf{Fm_l=l_r}$, in the
other image, upon which the corresponding point $\mathbf{m_r}$ should
lie. Generally, all epipolar lines are not in parallel with each other
and passing through the epipole (namely, $\mathbf{l_l^T} \mathbf{e}_{l}
= \mathbf{l_r}^{T} \mathbf{e}_{r} = \mathbf{0}$). 

%%%%%%%%%%%%%%%%%%%%%%%%%%%%%%%%%%%%%%%%%%%%%%%%%%%%%%%%%%
\begin{figure}[t]
\centering
\includegraphics[width=0.35\textwidth]{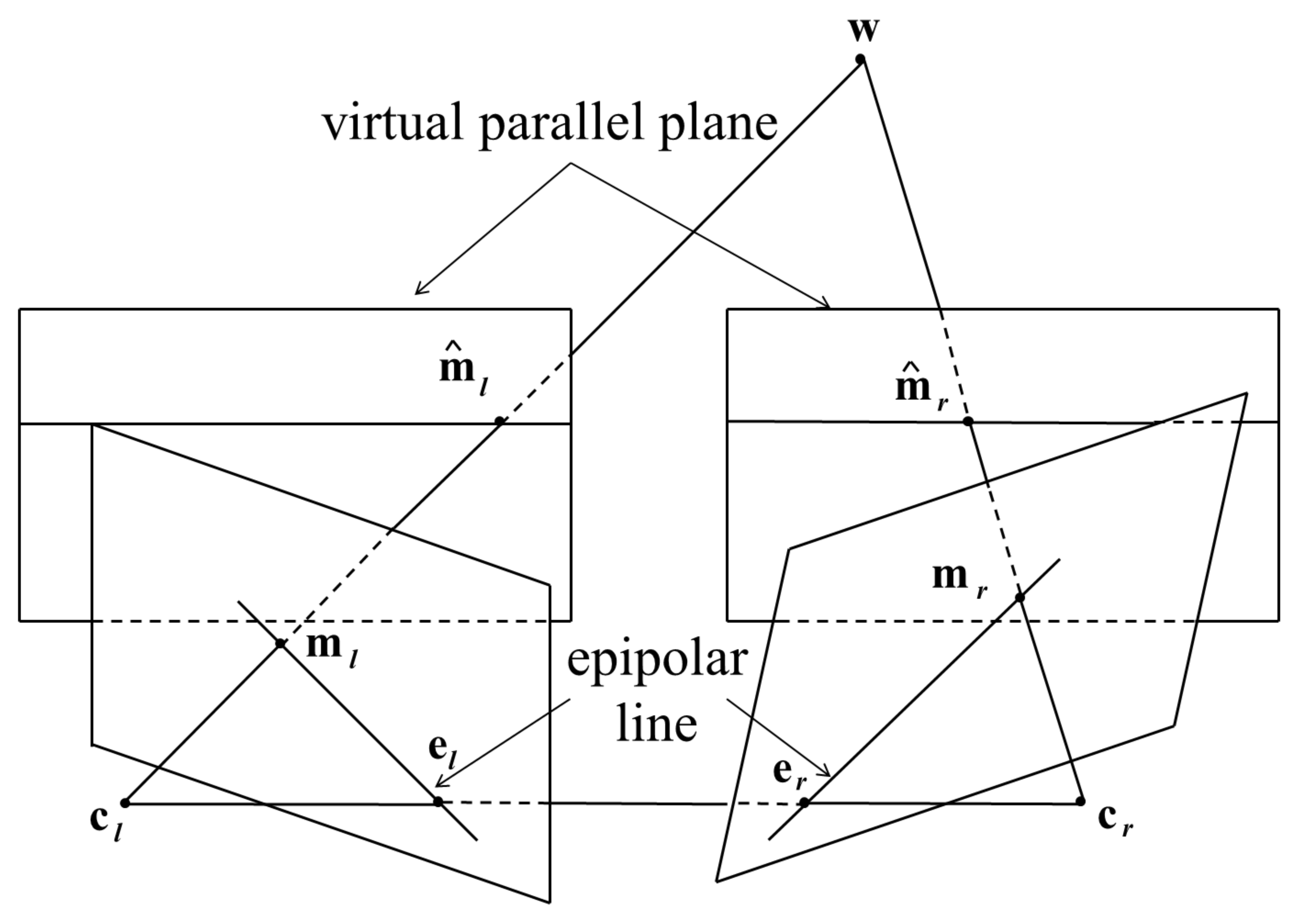}
\caption{Illustration of image rectification.}\label{fig:rectification}
\end{figure}
%%%%%%%%%%%%%%%%%%%%%%%%%%%%%%%%%%%%%%%%%%%%%%%%%%%%%%%%%%

Image rectification as shown in Fig.~\ref{fig:rectification} is the
process of converting the epipolar geometry of a given stereo image pair
into a canonical form that satisfies two conditions: 1) all epipolar
lines are parallel to the baseline, 2) there is no vertical disparity
between the corresponding epipolar lines in both images. This can be
done by applying homographies to each of image planes or, equivalently,
mapping the epipoles to a point at infinity as $\mathbf{e}_{\infty}=[1~0
~0]^T$.  Especially, the fundamental matrix of a pair of rectified
images can be expressed in form of
\begin{equation} \label{eq:rectifiedF}
\mathbf{F}_{\infty}  = 
\left[ \begin{array}{ccc}
0 & 0 & 0 \\
0 & 0 & -1 \\
0 & 1 & 0
\end{array} \right].
\end{equation}
Let $\mathbf{H_l}$ and $\mathbf{H_r}$ be two rectifying homographies
of the left and right images, respectively, and $(\mathbf{\hat{m}_l},
\mathbf{\hat{m}_r})$ be the corresponding points in the rectified images.
Then, we have 
\begin{equation} \label{eq:rectpoint}
\mathbf{\hat{m}_l} = \mathbf{H}_{l}\mathbf{m}_{l}, \phantom{1111} 
\mathbf{\hat{m}_r} = \mathbf{H}_{r}\mathbf{m}_{r}.
\end{equation}
According to Eq.~(\ref{eq:epipolargeometry}),
\begin{equation} \label{eq:idealconstraint}
\mathbf{\hat{m}_l}^T \mathbf{F}_{\infty} \mathbf{\hat{m}_r} = \mathbf{0}.
\end{equation}
By incorporating Eq.~(\ref{eq:rectpoint}) in Eq.~(\ref{eq:idealconstraint}), we
obtain
\begin{equation} \label{eq:ES1_1}
\mathbf{m}_{l}^{T} \mathbf{H}_{l}^{T} \mathbf{F}_{\infty} \mathbf{H}_{r} 
\mathbf{m}_{r} =\mathbf{0}.
\end{equation}
As a result, the fundamental matrix of the original stereo image pair
can be specified as $\mathbf{F=\mathbf{H}_{l}^{T} \mathbf{F}_{\infty}
\mathbf{H}_{r}}$.  The fundamental matrix is used to calculate the
rectification error in the process of parameter optimization. Thus, the
way to parameterize $\mathbf{H_l}$ and $\mathbf{H_r}$ is critical to the
generation of good rectified images. 

\section{Proposed USR-CGD Rectification Algorithm}\label{sec:proposed}

\subsection{Generalized Homographies}\label{sec:4.1}

We begin with a brief review on the parameterization of the homography
as proposed by Fusiello {\em et al.}~\cite{cit:Fusiello2000,cit:Fusiello2011}.  
The rectification procedure is a process that defines new virtual
cameras $\mathbf{P}_{n_{l}}$ and $\mathbf{P}_{n_{r}}$ given old cameras
$\mathbf{P}_{ol}$ and $\mathbf{P}_{or}$. New virtual cameras are
obtained by rotating old cameras around their optical centers until two
focal planes contain the baseline and become coplanar. The rectifying
transformation is used to map the image plane of $\mathbf{P}_{o}$ onto
the image plane of $\mathbf{P}_{n}$. Without loss of generality, we use
the left camera as the example in the following discussion. The same
derivation applies to the right camera as well. 

For the left camera, we have
\begin{equation} \label{eq:ES1_2}
\mathbf{H}_{l} = \mathbf{P}_{nl_{1:3}} \mathbf{P}^{-1}_{ol_{1:3}},
\end{equation}
where the subscript denotes a range of columns. By Eq. (\ref{eq:opticalray}),  
the optical rays of each images can be represented by
\begin{eqnarray}\label{eq:optray1} 
\mathbf{\hat{w}}_{l}&=&\mathbf{\hat{c}}_{ol}+\alpha_{ol}
\mathbf{R}_{ol}^{-1}\mathbf{K}_{ol}^{-1}\mathbf{m}_{ol}, \\
\label{eq:optray2} 
\mathbf{\hat{w}}_{l}&=&\mathbf{\hat{c}}_{nl}+\alpha_{nl}
\mathbf{R}_{nl}^{-1}\mathbf{K}_{nl}^{-1}\mathbf{m}_{nl}.
\end{eqnarray}
Since it is assumed that the optical center does not move in old and new camera; namely, $\mathbf{\hat{c}_{ol}=\hat{c}_{nl}}$, in \cite{cit:Fusiello2011},
the homography is expressed as
\begin{equation} \label{eq:ES1} 
\mathbf{H}_{l}=\mathbf{K}_{nl}\mathbf{R}_{l}\mathbf{K}_{ol}^{-1},
\end{equation}
where $\mathbf{K}_{ol}$ and $\mathbf{K}_{nl}$ are intrinsic matrices of
the old and the new cameras, respectively, and $\mathbf{R}_{l}$ is the
rotation matrix that transform the old camera to the new camera for rectification. 

%%%%%%%%%%%%%%%%%%%%%%%%%%%%%%%%%%%%%%%%%%%%%%%%%%%%%%%%
\begin{figure}[t]
\centering
\begin{subfigure}[h!]{0.45\textwidth}
\centering
\includegraphics[width=3.2in]{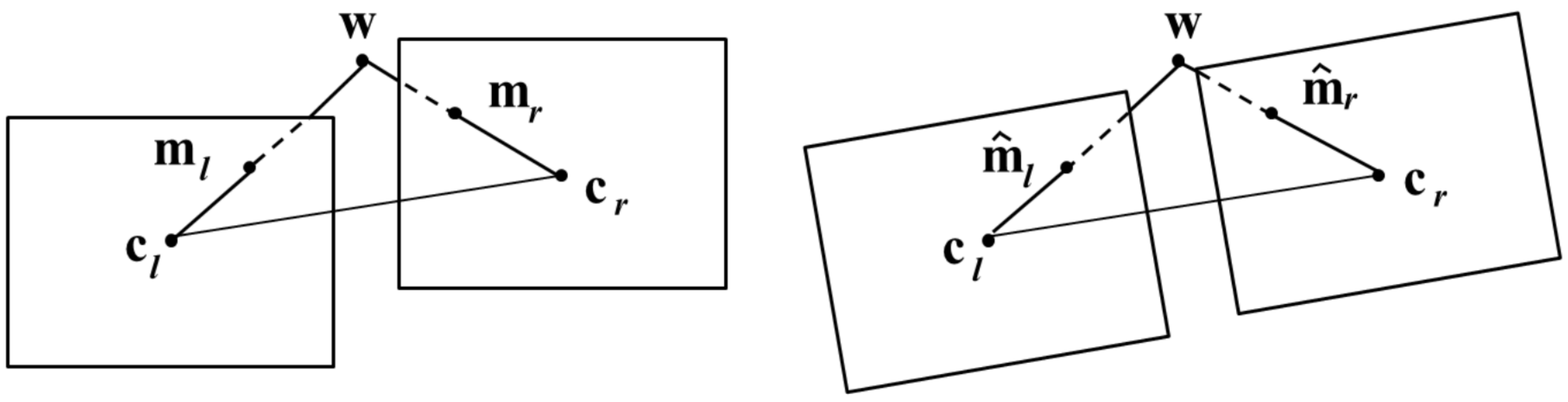}
\caption{Without parameter for y-translation}
\label{fig:ytranslation1}
\end{subfigure}\\
\begin{subfigure}[h!]{0.45\textwidth}
\centering
\includegraphics[width=3.2in]{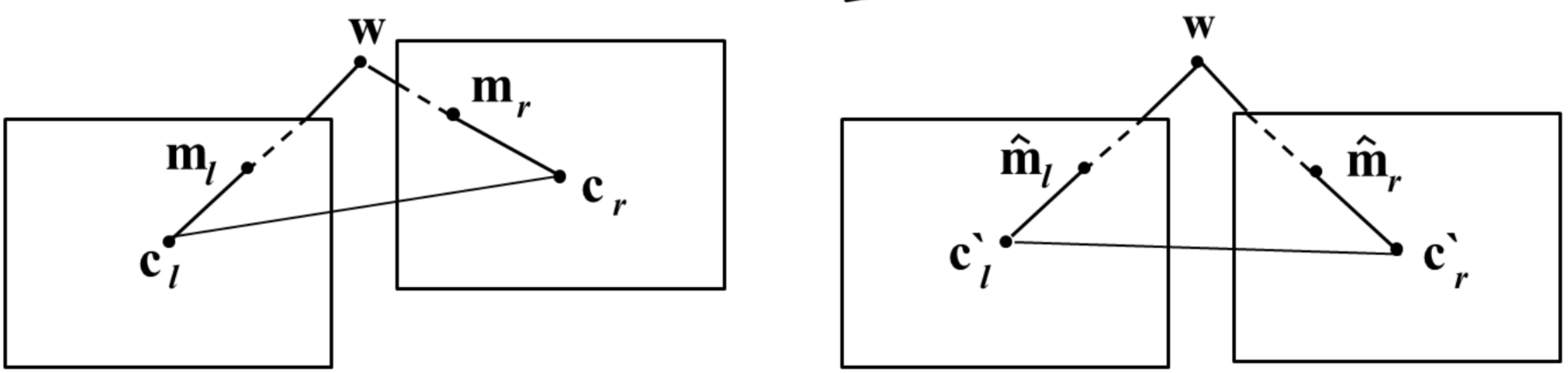} 
\caption{With parameter for y-translation}
\label{fig:ytranslation2}
\end{subfigure}
\caption{The effect of introducing a parameter for the y-translation: the original 
and rectified stereo pairs are shown in the left and right of subfigures (a) and (b).}
\label{fig:ytranslation}
\end{figure}
%%%%%%%%%%%%%%%%%%%%%%%%%%%%%%%%%%%%%%%%%%%%%%%%%%%%%%%%

The homography model as described above has its limitations since it
only allows the rotation around camera's optical center during the
rectification process. However, if there is only vertical disparity
between two images, the camera would still be rotated to make focal
planes coplanar, which inevitably introduces a warping distortion to the
rotated rectified image as shown in Fig.~\ref{fig:ytranslation1}.  Better
rectified images can be obtained by allowing an extended set of camera
parameters such as the displacement between optical centers and
different focal lengths. This is elaborated below.

First, we consider a new homography that includes the translation of the
optical center.  Based on Eqs.~(\ref{eq:optray1}) and
(\ref{eq:optray2}), we get
\begin{equation} \label{eq:optray3}
\mathbf{m}_{nl}=\mathbf{K}_{nl}\mathbf{R}_{nl}[\mathbf{c}_{tl}
+\mathbf{R}_{ol}^{-1}\mathbf{K}_{ol}^{-1}\mathbf{m}_{ol}],
\end{equation} where $\mathbf{c_{tl}=c_{ol}-c_{nl}}$ denotes the
movement between the optical centers of the old and the new cameras.
Since the horizontal disparity does not affect the rectification
performance, we focus on the case of camera's vertical translation.  The
corresponding homography can be derived by simplifying
Eq.~(\ref{eq:optray3}) as
\begin{equation} \label{eq:ES1} 
\mathbf{H_l}=\mathbf{K}_{nl}\mathbf{T_l}\mathbf{R_l}\mathbf{K}_{ol}^{-1}
\end{equation}
where 
\begin{equation} \label{eq:ES1}
\mathbf{T_l}  = 
\left[ \begin{array}{ccc}
1 & 0 & 0 \\
0 & 1 & t_{yl} \\
0 & 0 & 1
\end{array} \right]
\end{equation}
denotes a vertical translation matrix, which compensates the displacement as 
shown in Fig.~\ref{fig:ytranslation2}.

\begin{figure}[t]
\centering
\begin{subfigure}[h!]{0.45\textwidth}
\centering
\includegraphics[width=3in]{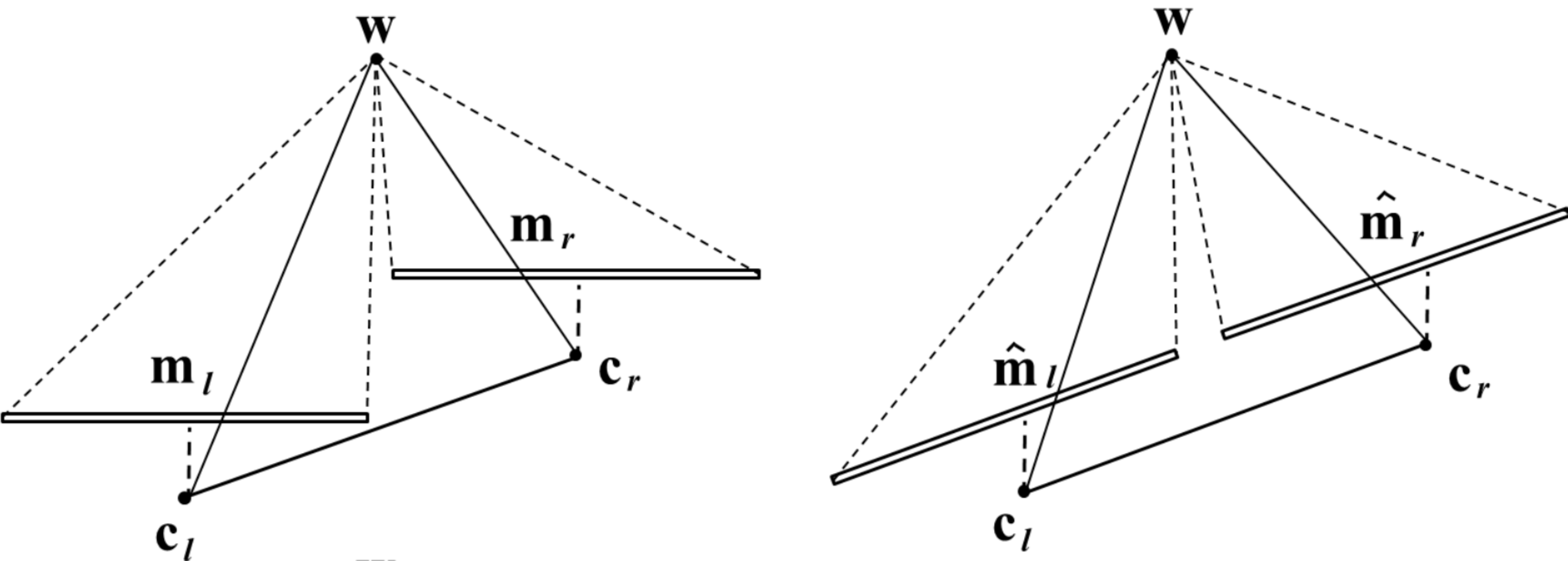}
\caption{Without parameter for z-translation}
\label{fig:ztranslation1}
\end{subfigure}\\

\begin{subfigure}[h!]{0.45\textwidth}
\centering
\includegraphics[width=3in]{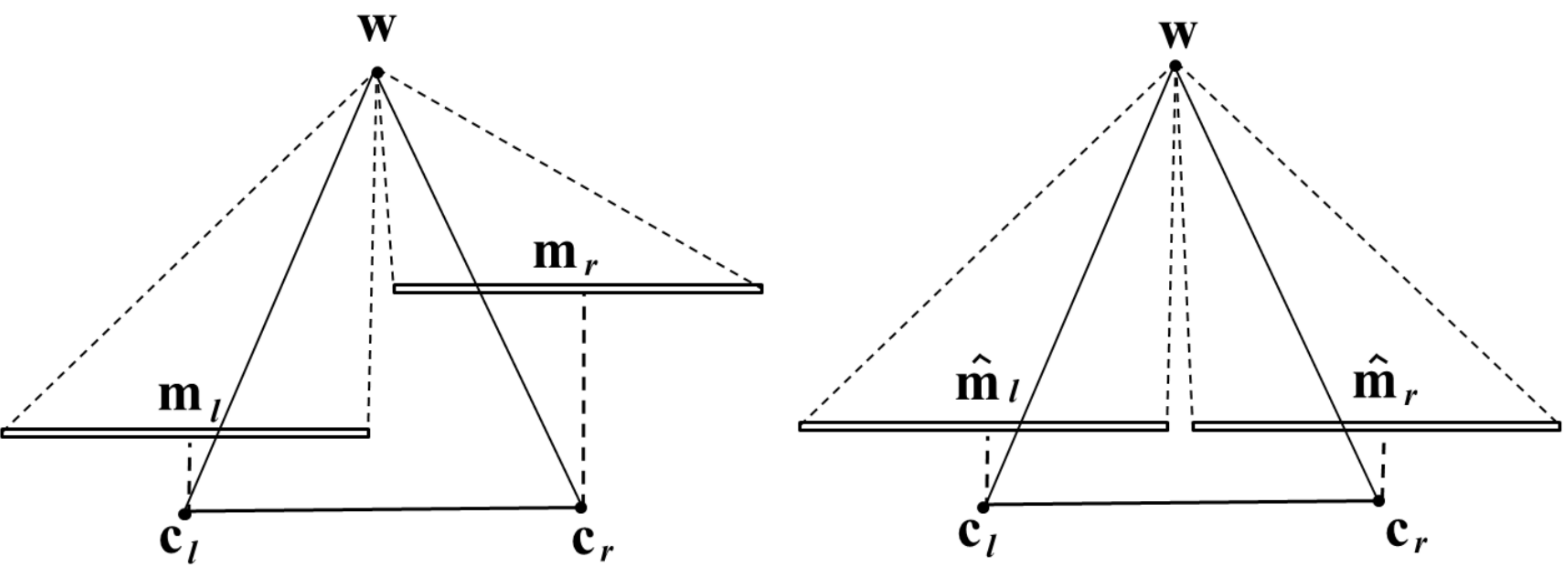} 
\caption{With parameter for z-translation}
\label{fig:ztranslation2}
\end{subfigure}
\caption{The effect of introducing a parameter for different zoom levels: the original 
and rectified stereo pairs are shown in the left and right of subfigures (a) and (b).}
\label{fig:ztranslation}
\end{figure}

Next, we examine another generalization by allowing different focal
lengths in two intrinsic matrices. In~\cite{cit:Fusiello2011}, intrinsic
matrices $\mathbf{K}_{ol}$ and $\mathbf{K}_{or}$ are set to be the same, 
{\em i.e.,}
\begin{equation} \label{eq:ES1}
\mathbf{K}_{ol} = \mathbf{K}_{or} = 
\left[ \begin{array}{ccc}
\alpha & 0 & \frac{w}{2} \\
0 & \alpha & \frac{h}{2} \\
0 & 0 & 1
\end{array} \right],
\end{equation}
where the same focal length $\alpha$ is adopted.  However, if there
exists an initial difference in the zoom level between the two cameras,
the rectification performance will degrade since two cameras should be
rotated on Z-axis to put both image planes to be coplanar
as shown in Fig.~\ref{fig:ztranslation1}, which reults in the up-scaled 
rectified images in horizontal direction.

%%%%%%%%%%%%%%%%%%%%%%%%%%%%%%%%%%%%%%%%%%%%%%%%%%%%%%%%
%%%%%%%%%%%%%%%%%%%%%%%%%%%%%%%%%%%%%%%%%%%%%%%%%%%%%%%%

To compensate the initial zoom difference, the left and the right
cameras are allowed to have different focal lengths. Thus, we have
\begin{equation} \label{eq:ES11}
\mathbf{K}_{ol} = 
\left[ \begin{array}{ccc}
\alpha_{l} & 0 & \frac{w}{2} \\
0 & \alpha_{l} & \frac{h}{2} \\
0 & 0 & 1
\end{array} \right], \quad \\
\mathbf{K}_{or} = 
\left[ \begin{array}{ccc}
\alpha_{r} & 0 & \frac{w}{2} \\
0 & \alpha_{r} & \frac{h}{2} \\
0 & 0 & 1
\end{array} \right].
\end{equation}
Geometrically, the two image planes are brought to a normalized image
plane without loss of perpendicularity with respect to the object point
as shown in Fig.~\ref{fig:ztranslation2} so that
the viewing angle of each camera is not changed. 

To summarize, we derive a generalized rectification homograhpy model
that consists of nine parameters in this section: 
\begin{itemize}
\item five rotational parameters: $\theta_{yl}$, $\theta_{zl}$, 
$\theta_{xr}$, $\theta_{yr}$, and $\theta_{zr}$; 
\item two y-translational parameters: $t_{yl}$ and $t_{yr}$; 
\item two focal length parameters: $\alpha_l$ and $\alpha_r$. 
\end{itemize}
The x-axis rotational parameter ($\theta_{xl}$), which affects the
portion of the scene that is projected, is set to zero.  Furthermore,
since images are projected to the virtual camera plane, we can
arbitrarily choose new intrinsic matrices as long as both cameras have
the same vertical focal length and the same vertical coordinate of the
principal point to meet the epipolar constraint in
Eq.~(\ref{eq:epipolargeometry}).  Thus, we demand $\mathbf{K}_{nl}$ and
$\mathbf{K}_{nr}$ to be the same as the old left one:
\begin{equation}
\mathbf{K}_{nl} = \mathbf{K}_{nr} = \mathbf{K}_{ol}.
\end{equation}
 
\subsection{Geometric Distortions}\label{sec:4.2}

The parameters of the rectification homography model are updated through
an optimization process to generate a pair of rectified images.
There are two measures on the quality of the rectified image: 1) the
rectification error and 2) errors of various geometric distortions. 

The rectification error is the average vertical disparity in pixels
between two corresponding points in the left and right images.  It is an
objective (or quantitative) measure, and it should be less than 0.5
pixels for a stereo matching algorithm to be performed in the 1-D
scanline. In the literature, the Samson error ($E_s$)
\cite{cit:Hartley2003,cit:Sampson} is widely used. It is the approximated 
projection error defined by
\begin{equation} \label{eq:Es}
E_s = \frac{1}{N}\sqrt{\sum_{j=1}^{N}(E_s)_j^2},
\end{equation}
where $N$ is the number of corresponding pairs and $(E_s)_j$ is the
$j$th component of normalized vector $E_s$, which is computed via
\begin{equation}
(E_s)_j^2 = \frac{({m_r^j}^TFm_l^j)^2}{(Fm_l^j)_1^2+(Fm_l^j)_2^2
+({m_r^j}^TF)_1^2+({m_r^j}^TF)_2^2}.
\end{equation}

In the image rectification process, we need to define a cost function
that takes both the objective quality (as represented by the
rectification error $E_s$) and the subjective quality (as represented by
geometric distortions) into account. The latter is needed since we
cannot avoid warping distortions in a perspective transformation.  Here,
the geometrical distortion is measured by the dissimilarity of a
rectified image from its original unrectified one. Several rectification
algorithms \cite{cit:Loop1999}-\cite{cit:Wu2005} have been proposed to
control the perspective distortion in rectified images using auxiliary
affine transforms ({\em e.g.}, the Jacobian or SVD of $H$, etc.), yet 
none of them is clearly superior to others in reaching well-balanced 
rectification quality. 

Mallon and Whelan~\cite{cit:Mallon2005} introduced two geometric
measures for performance evaluation: 1) orthogonality and 2) aspect
ratio. Let $a=(\frac{w}{2},0,1)^T$, $b=(w,\frac{h}{2},1)^T$,
$c=(\frac{w}{2},h,1)^T$ and $d=(0,\frac{h}{2},1)^T$ be four mid points
of an image before the transform, where $w$ and $h$ are the image width
and height. After the transform, the new positions of $a$, $b$, $c$ and
$d$ are denoted by $\hat{a}$, $\hat{b}$, $\hat{c}$ and $\hat{d}$,
respectively.  We examine vectors $\hat{x}=\hat{b}-\hat{d}$ and
$\hat{y}=\hat{c}-\hat{a}$. The orthogonality measure is defined as the
angle between $\hat{x}$ and $\hat{y}$:
\begin{equation}
E_O=\cos^{-1}{\frac{\hat{x}\hat{y}}{\lvert x\lvert \lvert y \rvert}}.
\end{equation}
It is desired that the orthogonality measure is as close to $90\degree$
as possible.  The aspect ratio is used to measure the ratio of image width to
image height before and after the rectification transform. Let
$a=(0,0,1)^T$, $b=(w,0,1)^T$, $c=(w,h,1)^T$ and $d=(0,h,1)^T$ be the
four corner points.  Again, we examine vectors $\hat{x}=\hat{b}-\hat{d}$
and $\hat{y}=\hat{c}-\hat{a}$ after the transform. The aspect ratio is
defined as
\begin{equation}
E_A=(\frac{\hat{x}^T\hat{x}}{\hat{y}^T\hat{y}})^{1/2}.
\end{equation}
The aspect ratio should be as close to unity as possible.

%%%%%%%%%%%%%%%%%%%%%%%%%%%%%%%%%%%%%%%%%%%%%%%%%%%%%%%%%%
\begin{figure}[t]
\centering
\includegraphics[width=0.25\textwidth]{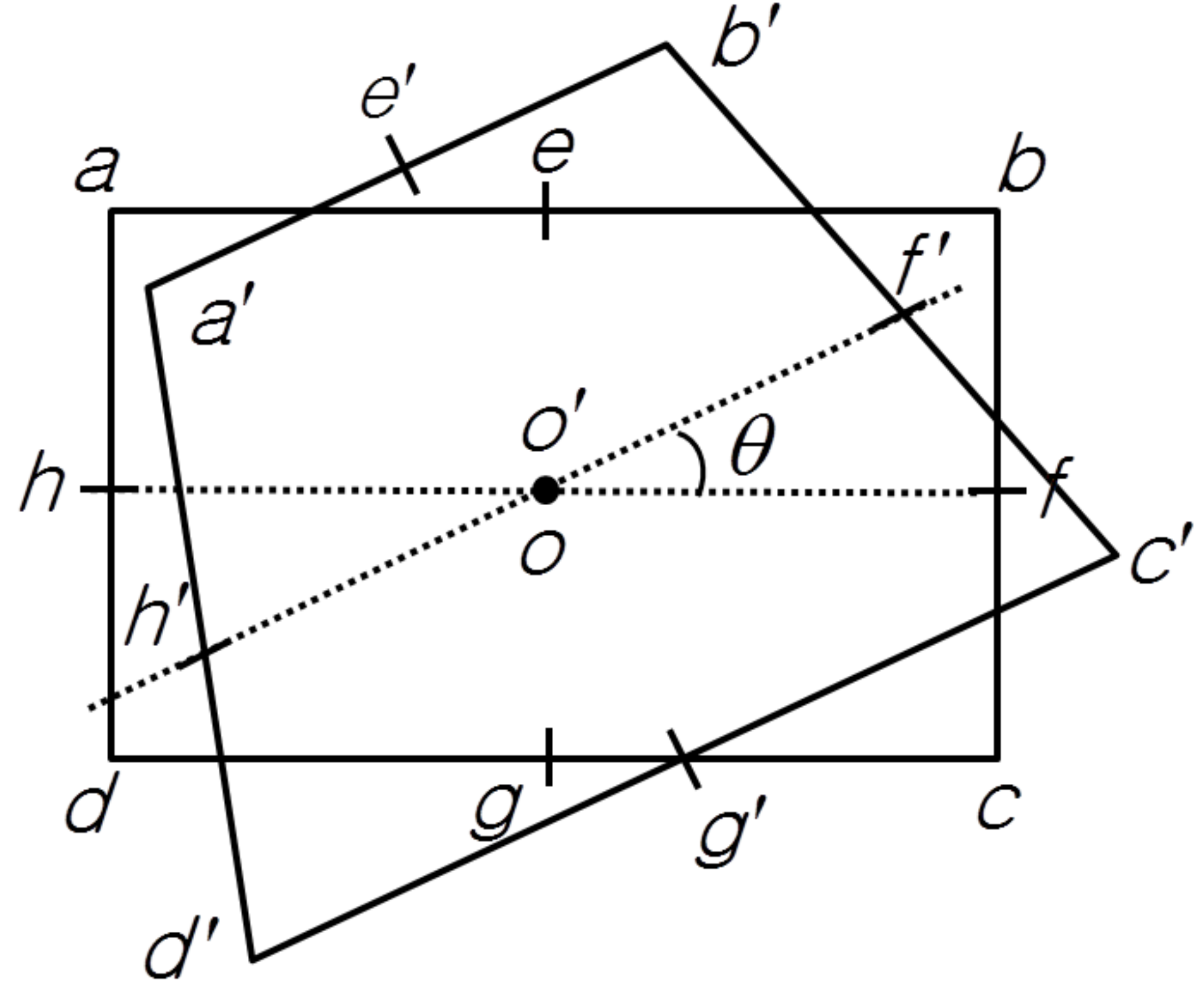}
\caption{The original and rectified images used to define four new
geometric measures.}\label{fig:geoerrors}
\end{figure}
%%%%%%%%%%%%%%%%%%%%%%%%%%%%%%%%%%%%%%%%%%%%%%%%%%%%%%%%%%

Although $E_O$ and $E_A$ are useful, they are still not sufficient to
characterize all possible geometric distortions in rectified images.
For example, $E_A$ is sometimes close to the unity in the presence of
large skewness. For this reason, we propose four new geometric
distortion measures, which includes a modified aspect ratio measure, and
use them in the regularization term of the cost function in the optimization 
process.

We define four corner points ($a$, $b$, $c$ and $d$) and four mid points
($e$, $f$, $g$ and $h$) in the original image as shown
in~Fig.~\ref{fig:geoerrors}.  The points in the rectified image are
represented by adding a prime to their symbols. The four new measures
are given below.
\begin{itemize}
\item Modified aspect ratio (ideally 1): \\
\begin{equation}\label{eq:AR} 
E_{AR}=\frac{1}{2}(\frac{\overline{a'o'}}{\overline{c'o'}}+\frac{\overline{b'o'}}
{\overline{d'o'}}).
\end{equation}
It gives a more accurate measure of the aspect ratio change.
\item Skewness (ideally 0\degree): \\
\begin{equation} \label{eq:Skewness} 
E_{Sk}=\frac{1}{4}\sum_{i=1}^{4}(\lvert 90\degree-\angle{CA_i} \rvert).
\end{equation}
It is used to measure how corner angles (denoted by $\angle CA_i$, $i=1,2,3,4$) 
deviate from 90\degree after rectification.
\item Rotation angle (ideally 0\degree): \\
\begin{equation} \label{eq:Rotation} 
E_{R}= \cos^{-1}(\frac{\overline{of} \cdot \overline{o'f'}}{\lvert \overline{of} 
\rvert \lvert \overline{o'f'} \rvert}).
\end{equation}
It is the angle between $\overline{of}$ and $\overline{\tilde{o}\tilde{f}}$,
which is used to measure the rotation degree after rectification. 
\item Size Ratio (ideally 1): \\
\begin{equation} \label{eq:Skewness} 
E_{SR}=\frac{Area_{rec}}{Area_{orig}}.
\end{equation}
It is used to measure to the size change of the input rectangle after rectification. 
\end{itemize}
The usefulness of the above four geometric distortion measures will be
shown in Section~\ref{sec:performance} based on extensive experiments.

Generally speaking, the cost function can be written as
\begin{equation} \label{eq:CostFunc} 
C(\phi)=E_s+\rho_{AR}E_{AR}+\rho_{Sk}E_{Sk}+\rho_{R} E_R+\rho_{SR}E_{SR},
\end{equation}
where $\phi$ denotes a set of rectification parameters to be selected in
the optimization process, $E_s$ is the Samson error to characterize the
rectification error, and $\rho_{AR}$, $\rho_{Sk}$, $\rho_{R}$ and
$\rho_{SR}$ are weights for the modified aspect ratio distortion, the
Skewness distortion, the rotation angle distortion and the size ratio
distortion, respectively. 

\subsection{Iterative Optimization}\label{sec:4.3}

To find an optimal solution to Eq. (\ref{eq:CostFunc}) is actually a
very challenging task. First, there is a tradeoff between the
rectification error and geometric errors. That is, when we attempt to
reduce the rectification error by adjusting the parameters of
generalized rectification homograhpy $H$, the geometric error tends to
increase, and vice versa.  As a result, the cost is a non-convex
function that has many local minima. Furthermore, the choice of proper
weights $\rho_{AR}$, $\rho_{Sk}$, $\rho_{R}$ and $\rho_{SR}$ is an open
issue remaining for future research.  To proceed, we simply adopt equal
weight solution.  That is, each weight takes either 0 or 0.25 two values
since there is one rectification error term and four geometric error
terms in Eq. (\ref{eq:CostFunc}), respectively.  For a specific type of
geometric distortion, if its error is less than a preset threshold, its
weight is set to 0. Otherwise, it is set to 0.25.  Mathematically, we
have
\begin{equation} \label{eq:rhotest} 
\rho_{X} = \left\{
\begin{array}{ll}
0.25, \quad & \mbox{if } E_{X} \ge T_{X}, \\
0, \quad & \mbox{if } E_{X} < T_{X}, \\
\end{array}
\right.
\end{equation}
where $X$ can be $AR$, $Sk$, $R$ or $SR$ and $T_X$ is the corresponding
preset threshold.

%%%%%%%%%%%%%%%%%%%%%%%%%%%%%%%%%%%%%%%%%%%%%%%%%%%%%%%%%%
\begin{figure}[t]
\centering
\includegraphics[width=0.48\textwidth]{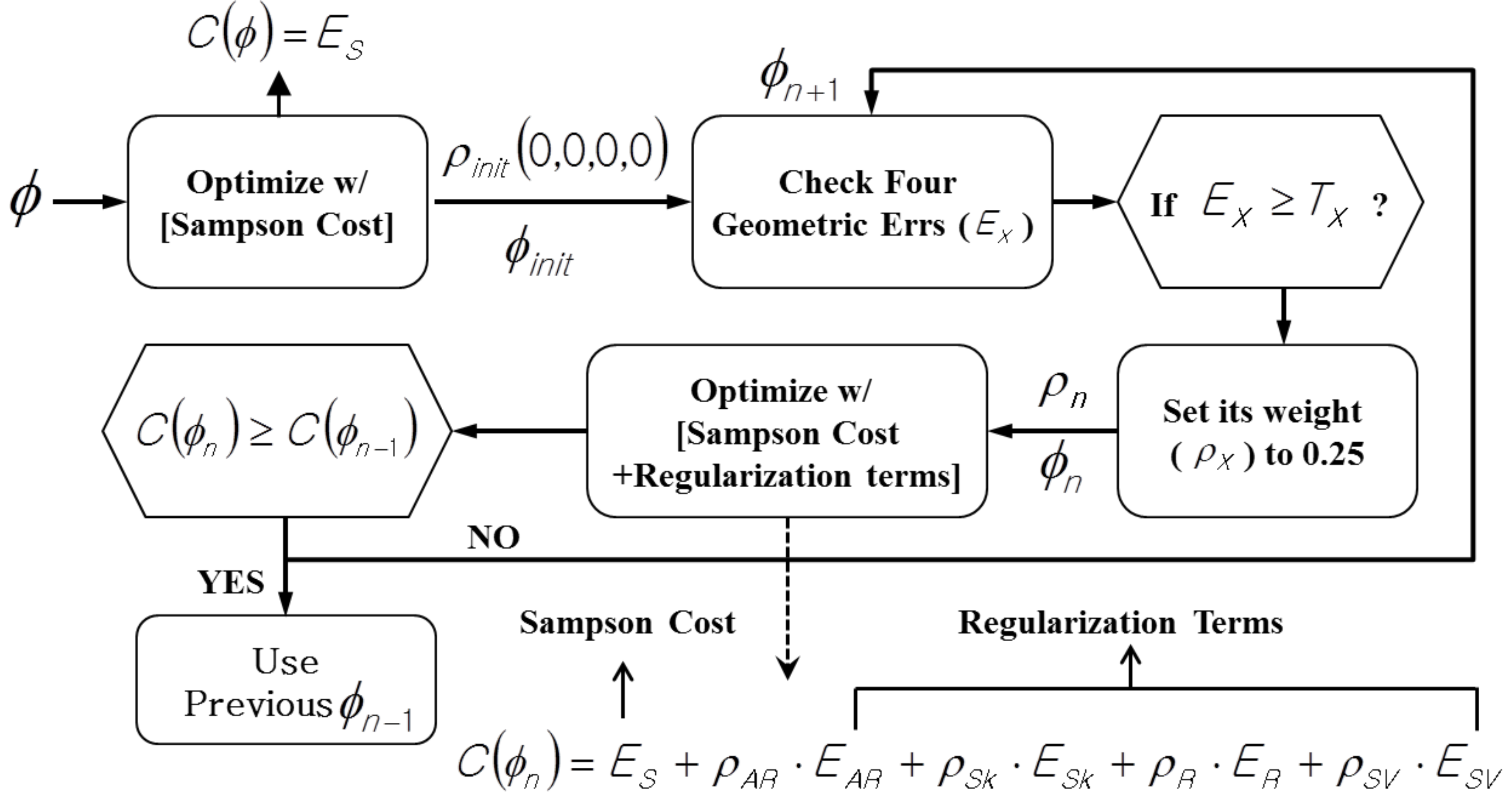}
\caption{The block diagram of the proposed iterative optimization procedure.}
\label{fig:constrainOpt}
\end{figure}
%%%%%%%%%%%%%%%%%%%%%%%%%%%%%%%%%%%%%%%%%%%%%%%%%%%%%%%%%%

Based on this design, the cost is a dynamically changing function that
always contains the rectification error. A geometric distortion term
will be included in the cost function (or "being turned-on") only when
it is larger than a threshold.  Otherwise, it is turned off. To solve
this problem, we propose an iterative optimization procedure as shown in
Fig.~\ref{fig:constrainOpt}. It consists of the following steps. 

\begin{enumerate}
\item We begin with a cost function that contains the Sampson error
($E_s$) only.  The optimization procedure offers the initial set of
rectification parameters, which is denoted by $\phi_{init}$.  
\item We update four weight $\rho=(\rho_{AR}, \rho_{Sk}, \rho_{R},
\rho_{SV})$ from $\phi_{init}$, which is initially set to $\rho_{init}=(0,0,0,0)$.
\item Under the current set of rectification parameters, $\phi$, and the
current set of weights, $\rho$, we solve the optimization problem with
respect to Eq. (\ref{eq:CostFunc}). Then, both rectificatioin parameters
and weights are updated accordingly.  Step 3 is iterated until the the cost 
of the current round is greater than or equal to that of the previous round.
Mathematically, if $C(\phi_{n}) \ge C(\phi_{n-1})$, we choose $\phi_{n-1}$
as the converged rectification parameters.
\end{enumerate}

When we compare the current cost $C(\phi_{n})$ with the cost in the
previous round $C(\phi_{n-1})$ in Step 3, the number of geometric terms
in Eq. (\ref{eq:CostFunc}) may change. If this occurs, we should
compensate it for fair comparison. That is, the cost should be
normalized by the sum of weight via $C_{normalized}(\phi_{n})=
C(\phi_{n})/(1+\sum_X{\rho_X})$. 

The choice of a proper threshold value for each geometric error is
important in reaching balanced performance. In this work, we set
threshold values of geometric errors to the following:
\begin{eqnarray}
0.8 \leq E_{AR} \leq 1.2, & \quad & E_{Sk}\leq 5\degree, \\
0.8 \leq E_{SV} \leq 1.2, & \quad & \lvert E_{R} \rvert \leq 30\degree
\end{eqnarray}
Furthermore, we normalize four geometric errors by considering their
value range. The normalizing factors are:
\begin{equation}
N_{AR}=1.5, \quad N_{Sk}=6.5, \quad N_{R}=18.5, \quad N_{SV}=2.5,
\end{equation}
which can be absorbed in their respective weight; {\em i.e.} the new
weight becomes $\rho_X=0.25/N_X$ when the term is on.  Last, the
minimization of the cost function is carried out using the nonlinear
least square method, \emph{Trust Region}~\cite{cit:Yuan2010}, starting
with all unknown variables set to zero. 

%%%%%%%%%%%%%%%%%%%%%%%%%%%%%%%%%%%%%%%%%%%%%%%%%%%%%%%%%%
\begin{figure}[t]
\centering
\includegraphics[width=0.5\textwidth]{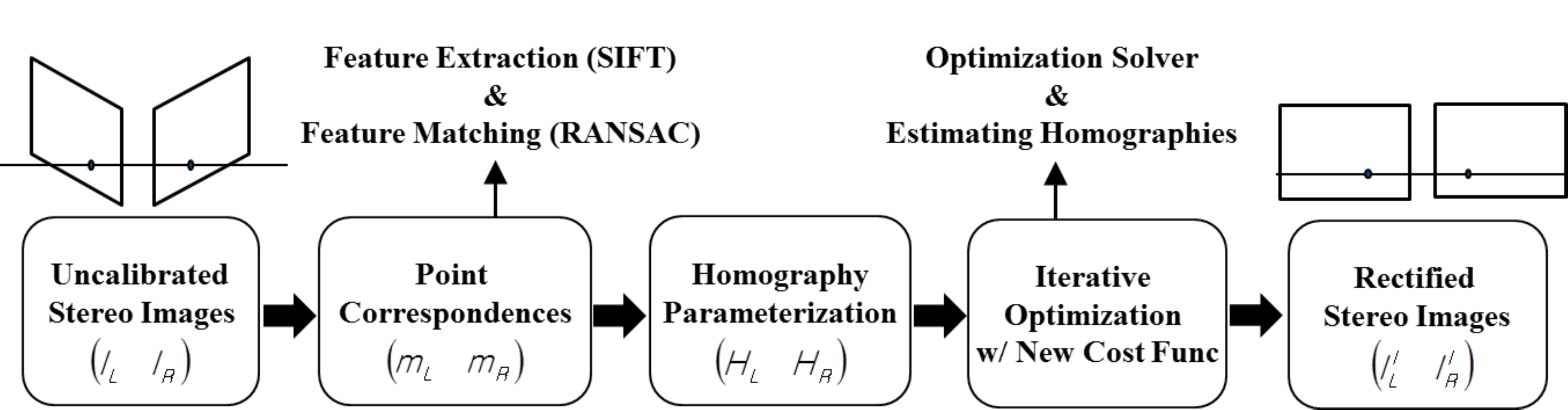}
\caption{The block-diagram of the proposed USR-CGD system.}
\label{fig:overall_system}
\end{figure}
%%%%%%%%%%%%%%%%%%%%%%%%%%%%%%%%%%%%%%%%%%%%%%%%%%%%%%%%%%

%%%%%%%%%%%%%%%%%%%%%%%%%%%%%%%%%%%%%%%%%%%%%%%%%%%%%%
\begin{figure*}[t]
	\centering
	\begin{subfigure}[h!]{1.5 in}
		\centering
		\includegraphics[width=1.6in]{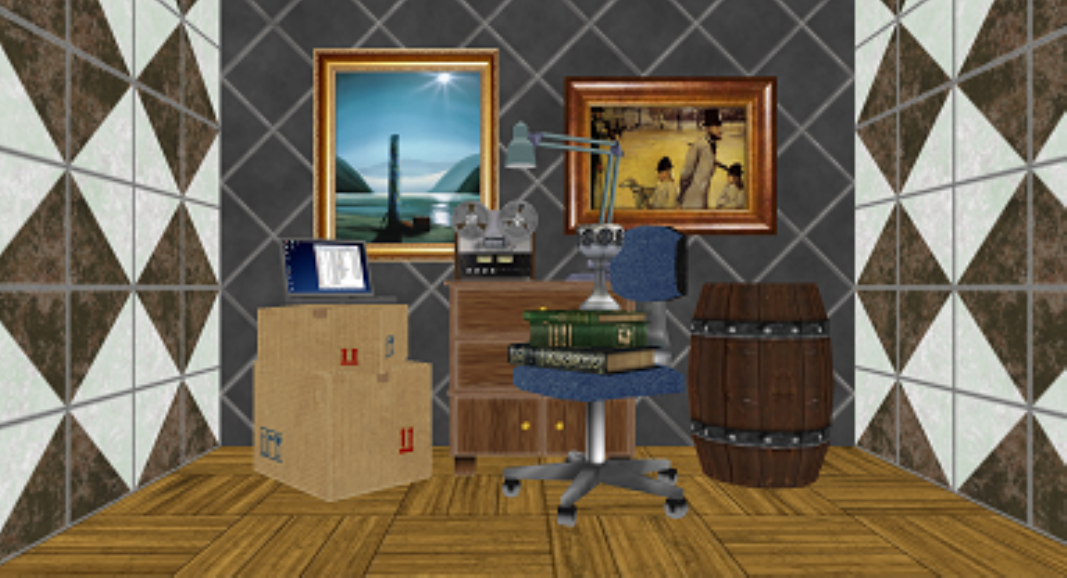} % Example image
		\caption{Interior}\label{fig:SSref1}
	\end{subfigure}\quad	
	\begin{subfigure}[h!]{1.5in}
		\centering
		\includegraphics[width=1.6in]{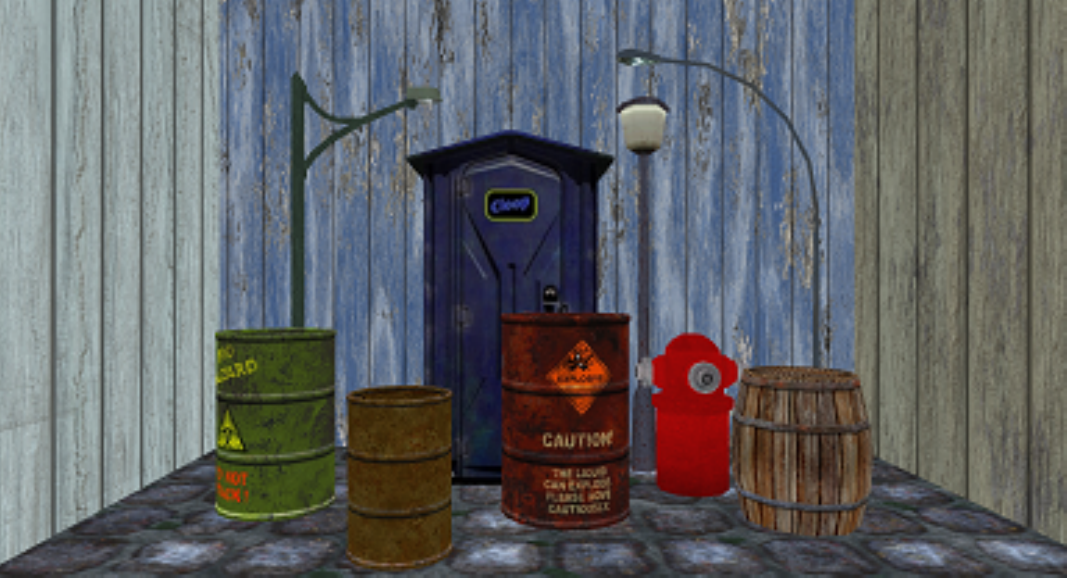} % Example image
		\caption{Street}\label{fig:SSref2}
	\end{subfigure}\quad
	\begin{subfigure}[h!]{1.5in}
		\centering
		\includegraphics[width=1.6in]{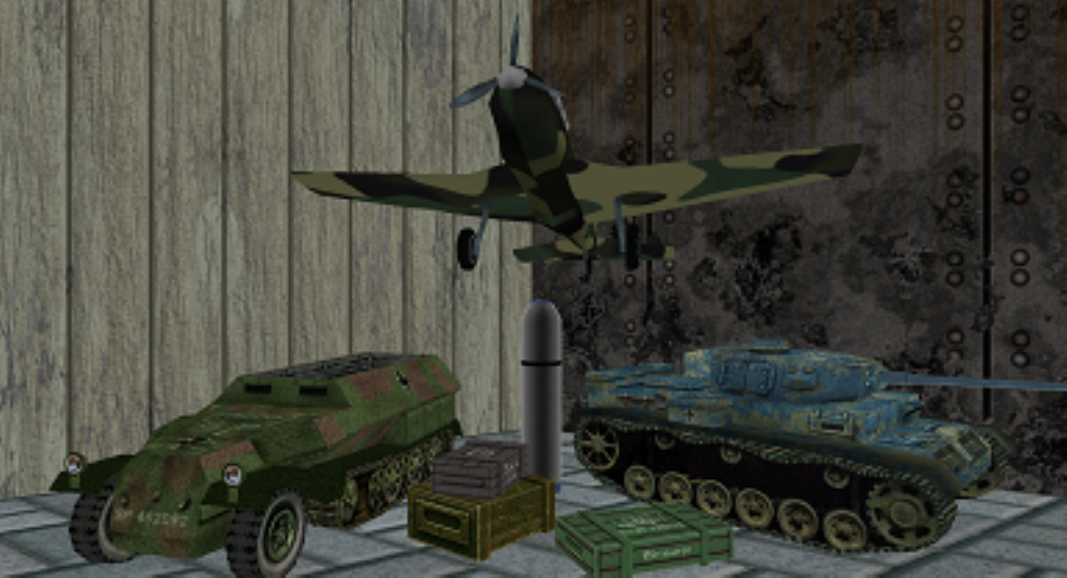} % Example image
		\caption{Military}\label{fig:SSref3}
	\end{subfigure}\quad
	\begin{subfigure}[h!]{1.5in}
		\centering
		\includegraphics[width=1.6in]{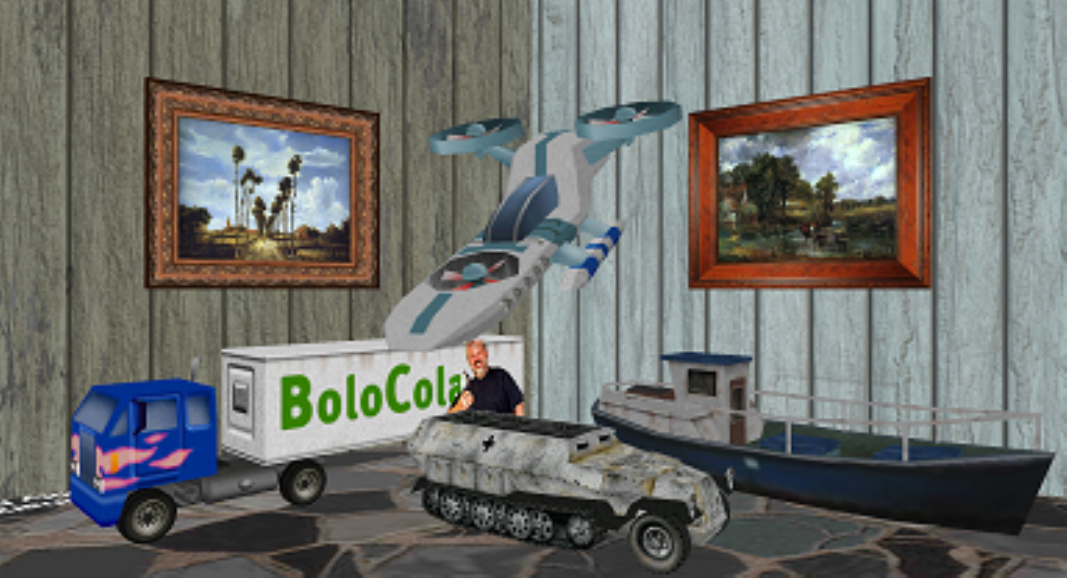} % Example image
		\caption{Vehicle}\label{fig:SSref4}
	\end{subfigure}
	\caption{The left images of 4 MCL-SS reference image pairs.}
	\label{fig:MCLSSref}
\end{figure*}
%%%%%%%%%%%%%%%%%%%%%%%%%%%%%%%%%%%%%%%%%%%%%%%%%%%%%%

%%%%%%%%%%%%%%%%%%%%%%%%%%%%%%%%%%%%%%%%%%%%%%%%%%%%%%
\begin{figure*}[t]
	\centering
	\begin{subfigure}[h!]{1.5 in}
		\centering
		\includegraphics[width=1.6in]{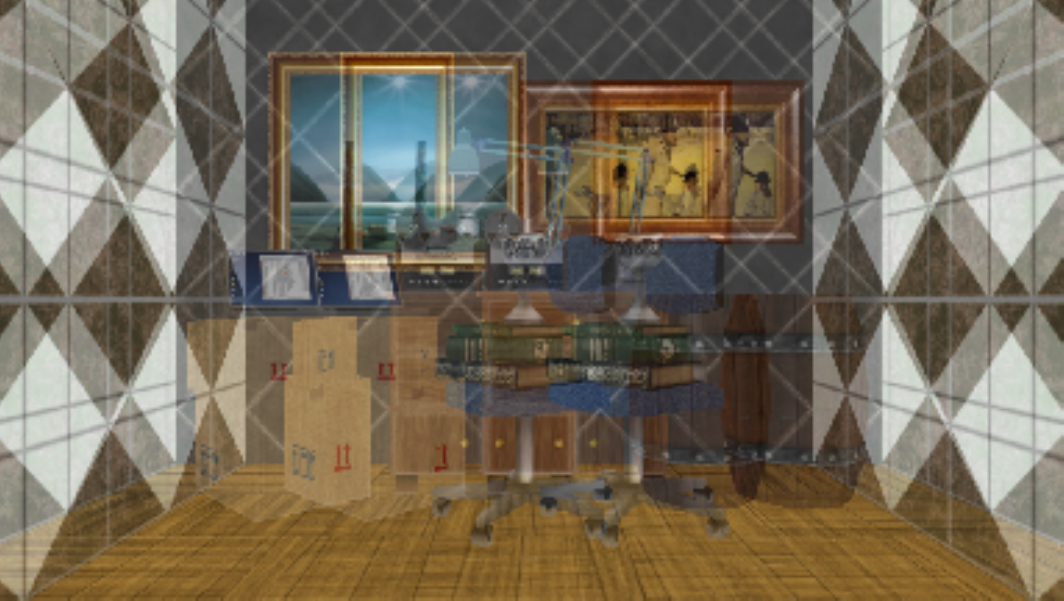} % Example image
		\caption{Trans. Error on X-axis}\label{fig:SStest1}
	\end{subfigure}\quad	
	\begin{subfigure}[h!]{1.5in}
		\centering
		\includegraphics[width=1.6in]{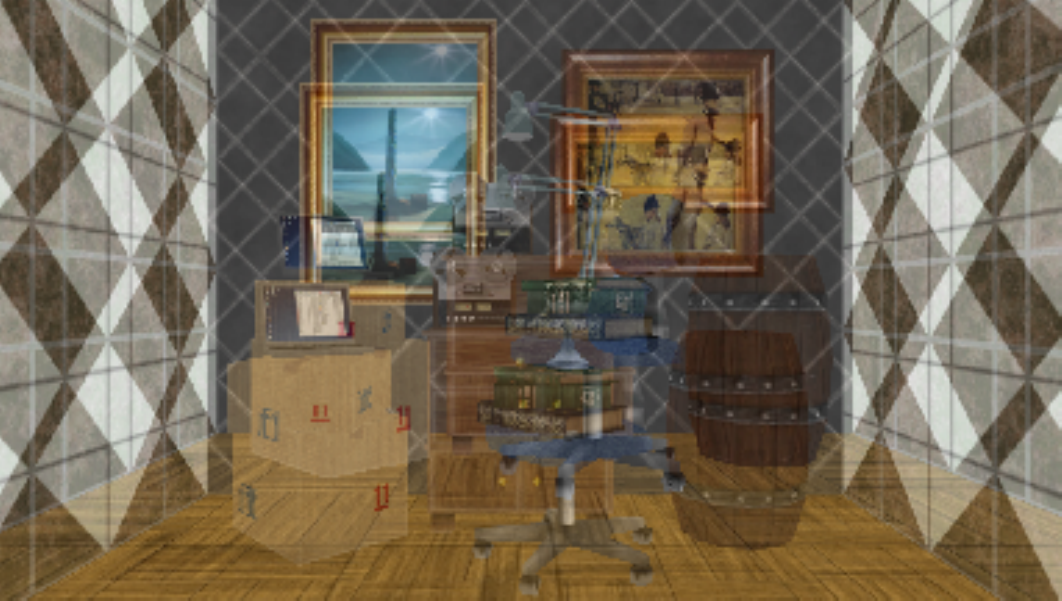} % Example image
		\caption{Trans. Error on Y-axis}\label{fig:SStest2}
	\end{subfigure}\quad
	\begin{subfigure}[h!]{1.5in}
		\centering
		\includegraphics[width=1.6in]{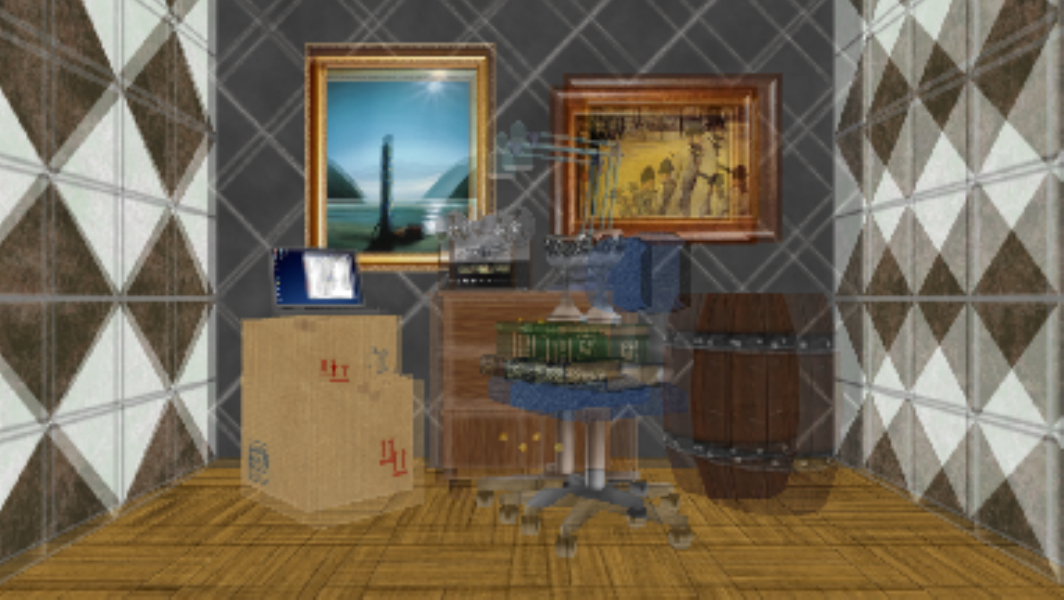} % Example image
		\caption{Trans. Error on Z-axis}\label{fig:SStest3}
	\end{subfigure}\quad
	\begin{subfigure}[h!]{1.5in}
		\centering
		\includegraphics[width=1.6in]{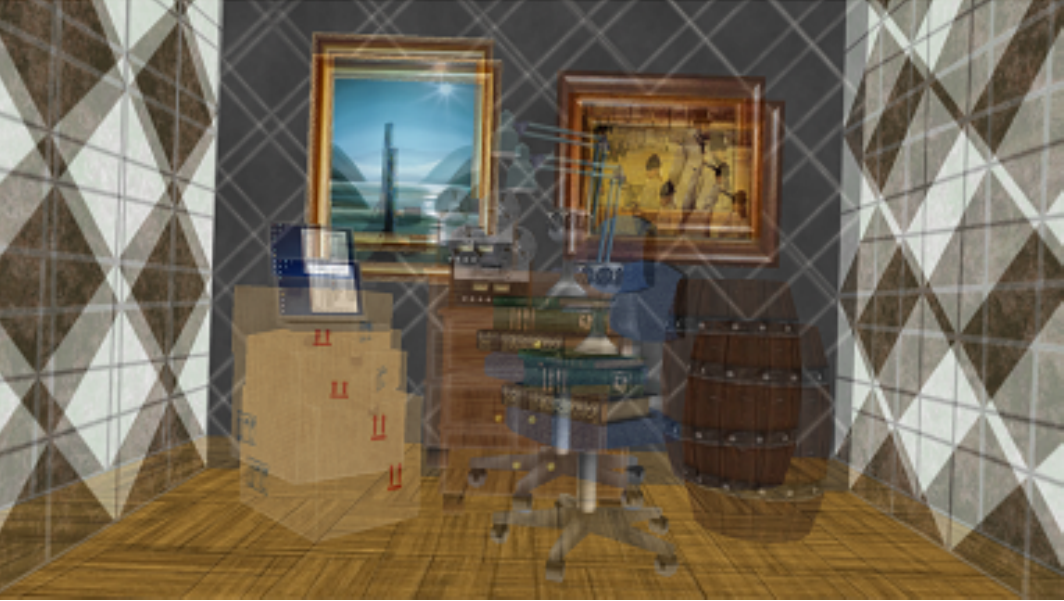} % Example image
		\caption{Compound Error 1}\label{fig:SStest4}
	\end{subfigure}\\
	\begin{subfigure}[h!]{1.5 in}
		\centering
		\includegraphics[width=1.6in]{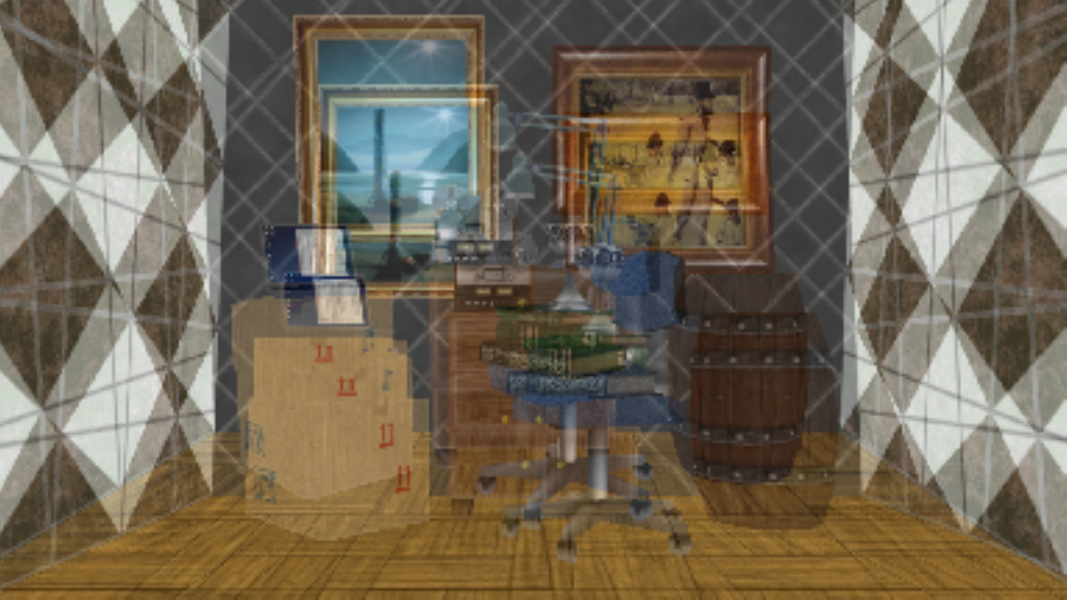} % Example image
		\caption{Rot. Error on X-axis}\label{fig:SStest5}
	\end{subfigure}\quad	
	\begin{subfigure}[h!]{1.5in}
		\centering
		\includegraphics[width=1.6in]{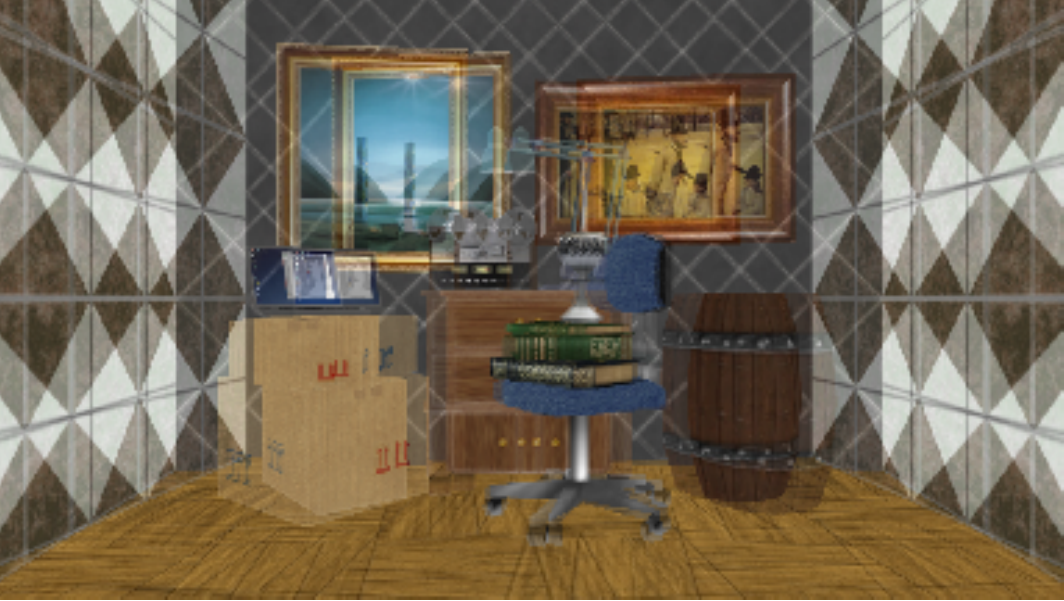} % Example image
		\caption{Rot. Error on Y-axis}\label{fig:SStest6}
	\end{subfigure}\quad
	\begin{subfigure}[h!]{1.5in}
		\centering
		\includegraphics[width=1.6in]{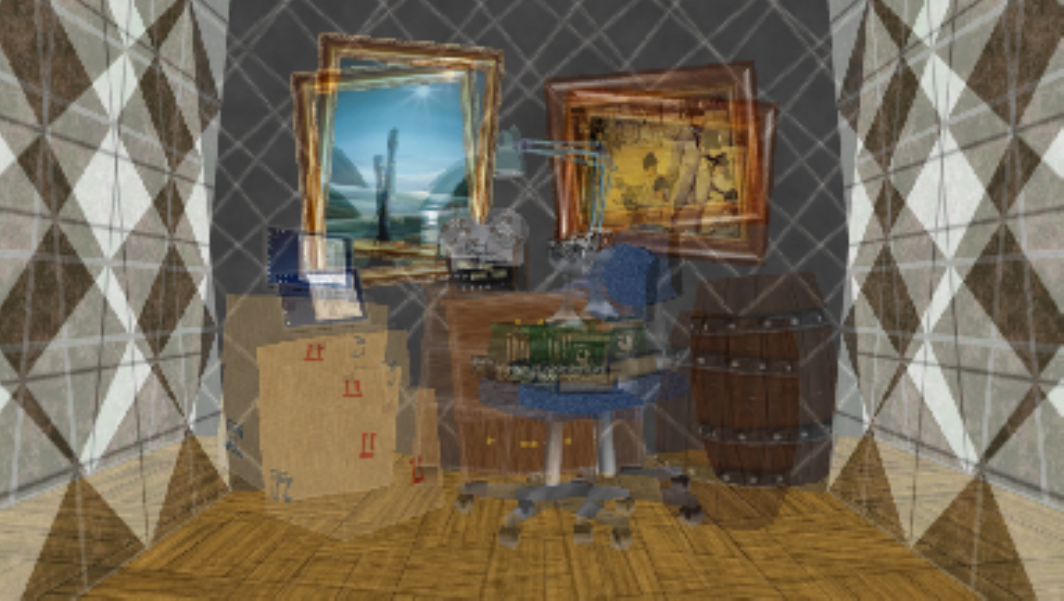} % Example image
		\caption{Rot. Error on Z-axis}\label{fig:SStest7}
	\end{subfigure}\quad
	\begin{subfigure}[h!]{1.5in}
		\centering
		\includegraphics[width=1.6in]{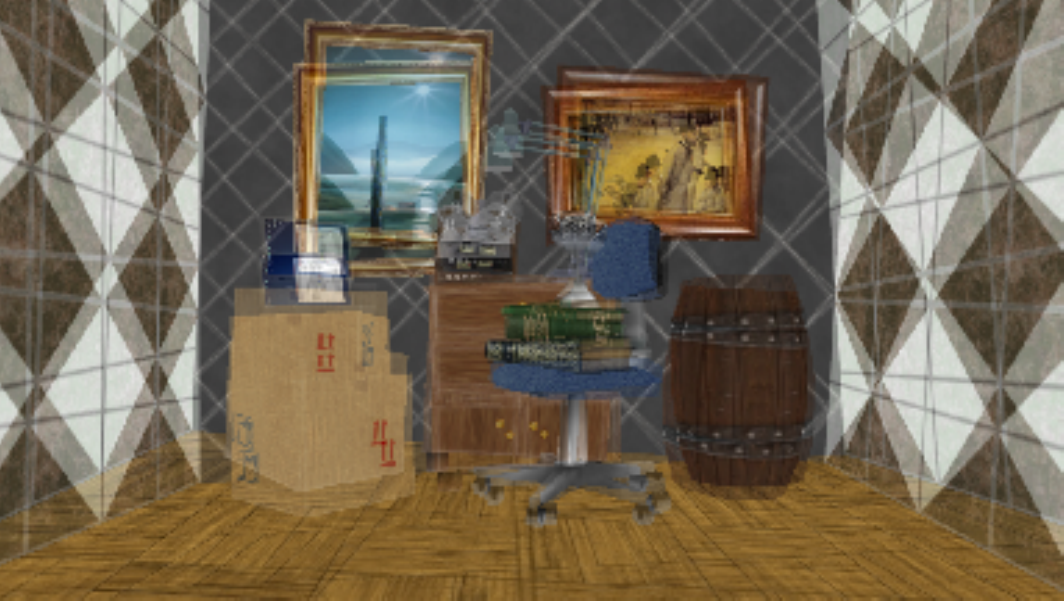} % Example image
		\caption{Compound Error 2}\label{fig:SStest8}
	\end{subfigure}
	\caption{The 8 test image pairs of ``Interior" with different geometric distortions, where 
        left and right images are overlapped for the display purpose.}
	\label{fig:MCLSStest}
\end{figure*}
%%%%%%%%%%%%%%%%%%%%%%%%%%%%%%%%%%%%%%%%%%%%%%%%%%%%%%

%%%%%%%%%%%%%%%%%%%%%%%%%%%%%%%%%%%%%%%%%%%%%%%%%%%%%%
\begin{figure*}[t]
	\centering
	\begin{subfigure}[h!]{1.5 in}
		\centering
		\includegraphics[width=1.6in]{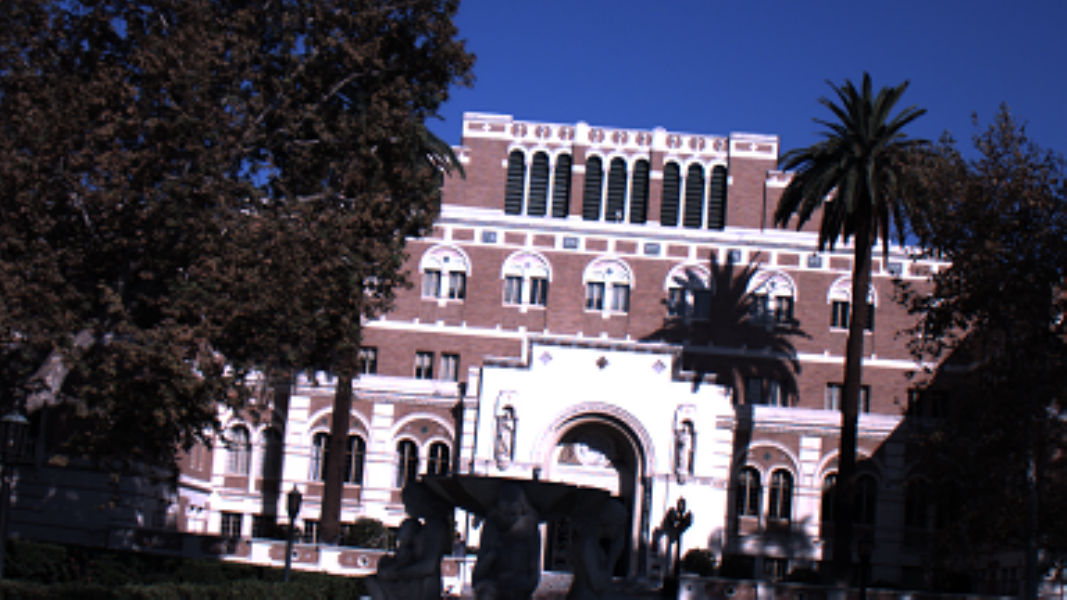} % Example image
		\caption{Doheny}\label{fig:RSref1}
	\end{subfigure}\quad	
	\begin{subfigure}[h!]{1.5in}
		\centering
		\includegraphics[width=1.6in]{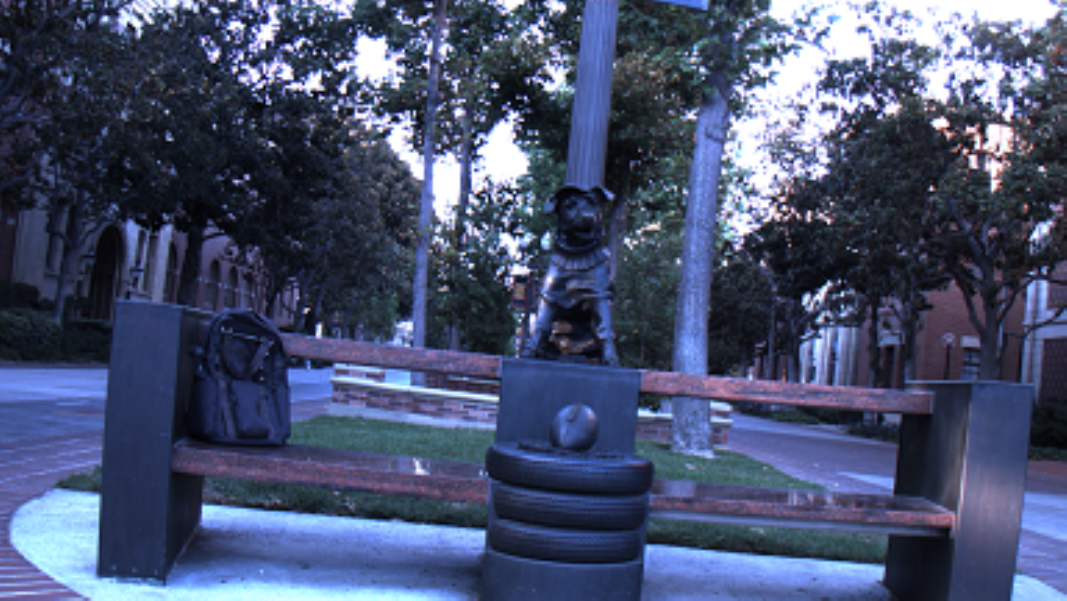} % Example image
		\caption{Dog}\label{fig:RSref2}
	\end{subfigure}\quad
	\begin{subfigure}[h!]{1.5in}
		\centering
		\includegraphics[width=1.6in]{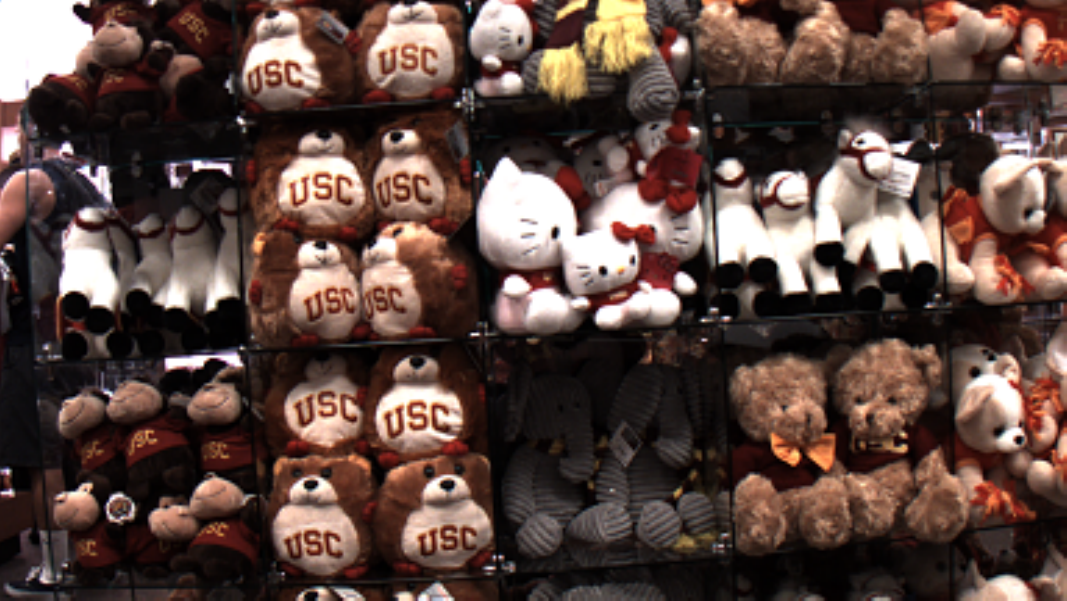} % Example image
		\caption{Dolls}\label{fig:RSref3}
	\end{subfigure}\quad
	\begin{subfigure}[h!]{1.5in}
		\centering
		\includegraphics[width=1.6in]{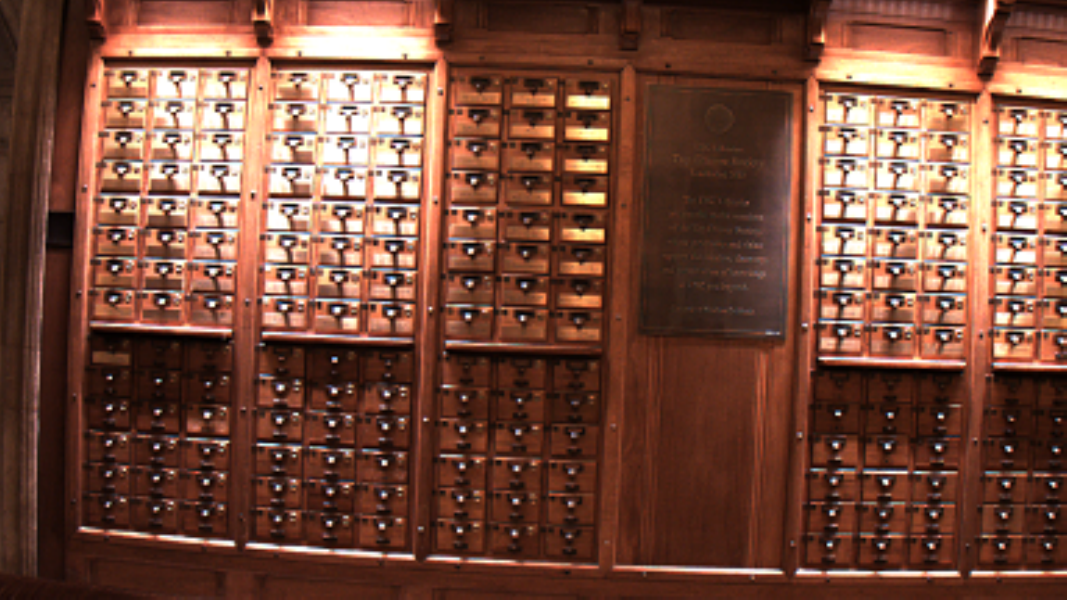} % Example image
		\caption{Drawer}\label{fig:RSref4}
	\end{subfigure}\\
	\begin{subfigure}[h!]{1.5 in}
		\centering
		\includegraphics[width=1.6in]{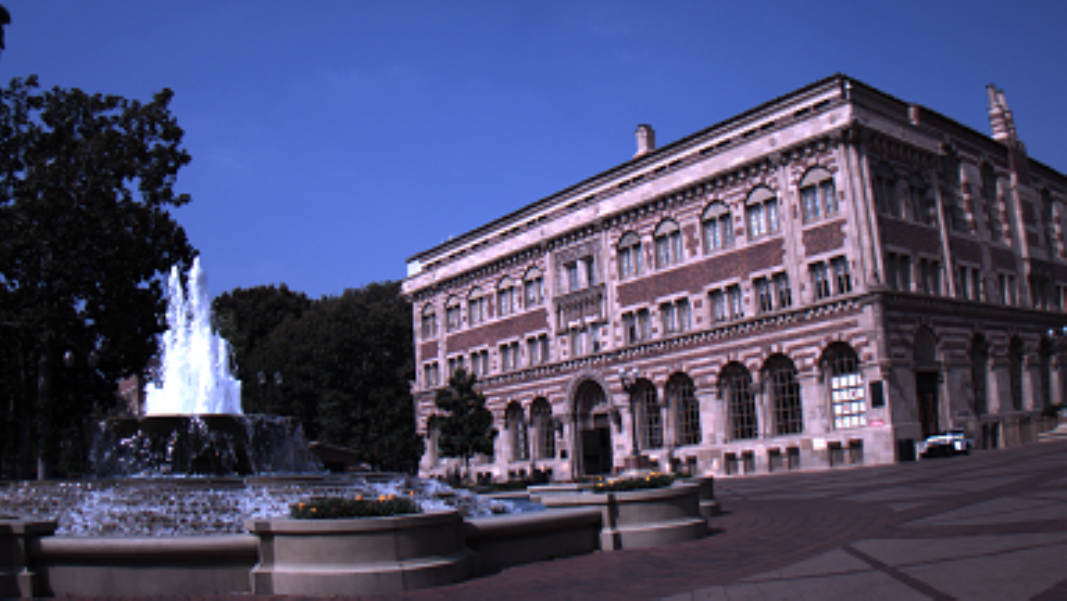} % Example image
		\caption{Fountain}\label{fig:RSref5}
	\end{subfigure}\quad	
	\begin{subfigure}[h!]{1.5in}
		\centering
		\includegraphics[width=1.6in]{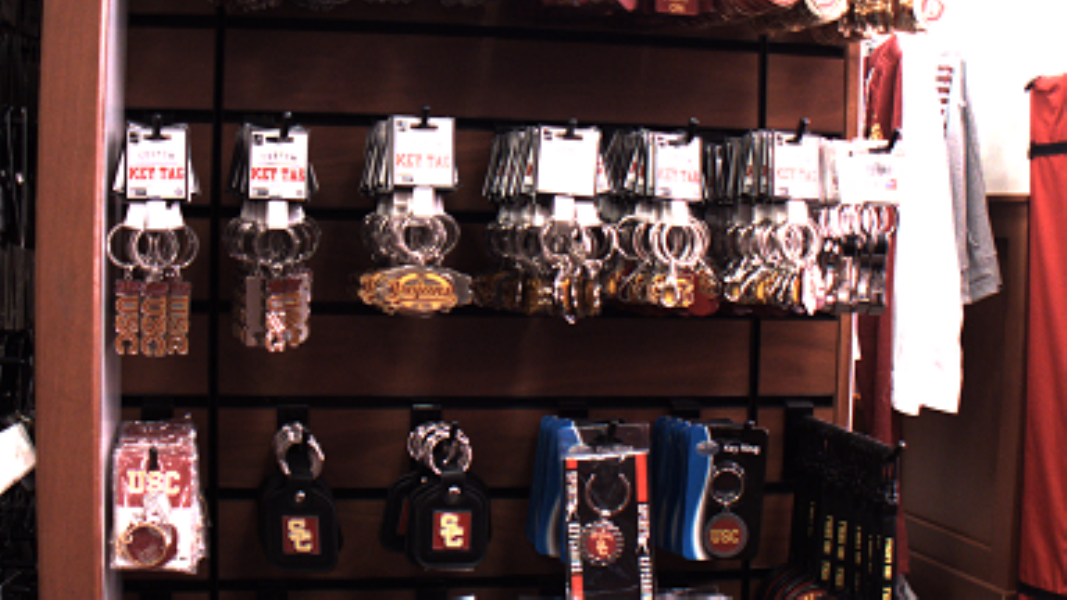} % Example image
		\caption{Keychain}\label{fig:RSref6}
	\end{subfigure}\quad
	\begin{subfigure}[h!]{1.5in}
		\centering
		\includegraphics[width=1.6in]{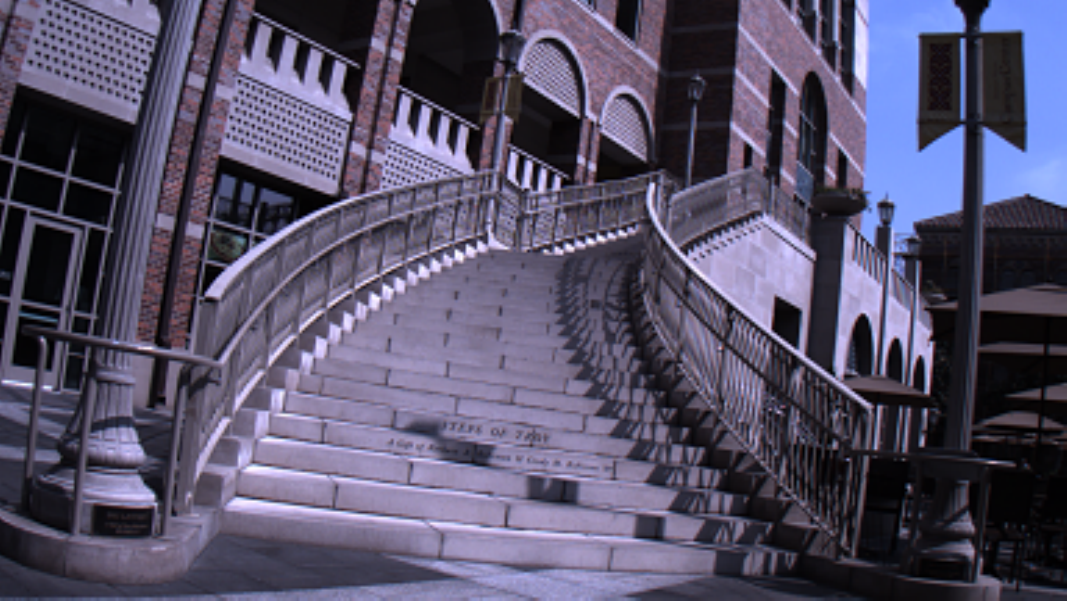} % Example image
		\caption{Step}\label{fig:RSref7}
	\end{subfigure}\quad
	\begin{subfigure}[h!]{1.5in}
		\centering
		\includegraphics[width=1.6in]{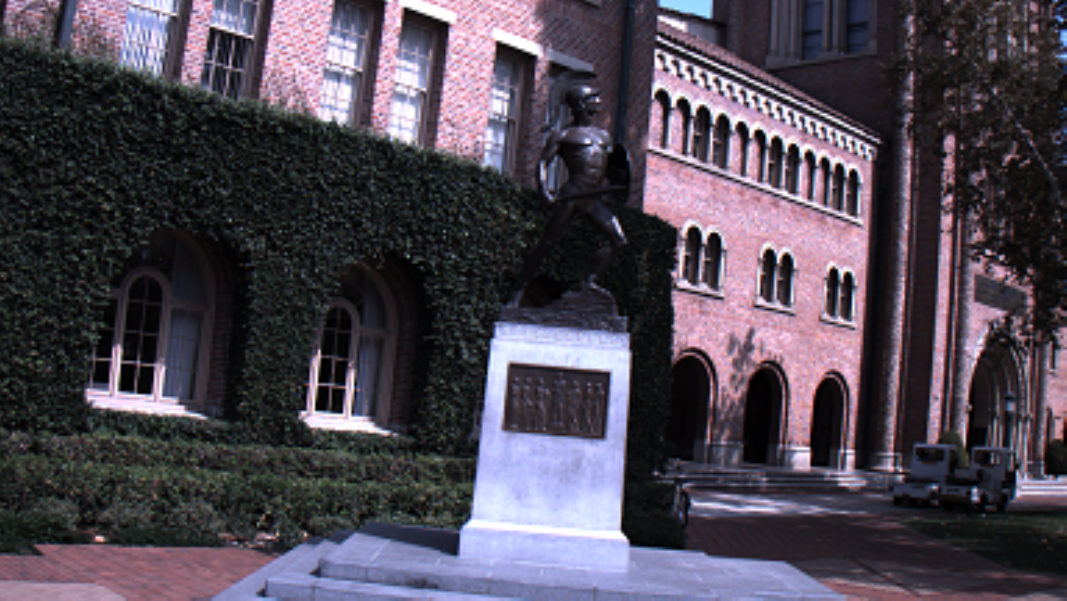} % Example image
		\caption{Tommy}\label{fig:RSref8}
	\end{subfigure}
	\caption{The eight selected left images from 20 MCL-RS image pairs.}
	\label{fig:MCLRSref}
\end{figure*}
%%%%%%%%%%%%%%%%%%%%%%%%%%%%%%%%%%%%%%%%%%%%%%%%%%%%%%

The block-diagram of the proposed USR-CGD system is shown in
Fig.~\ref{fig:overall_system}. This system is fully automatic since
there is no need to estimate the fundamental matrix. To establish the
point correspondence between the left and right images, we extract the
SIFT feature~\cite{cit:Lowe2004} and find the initial putative matching
points. We also apply RANSAC~\cite{cit:Fischler1881} to remove outliers.
It is noteworthy that the number of the correspondences strongly affects
the rectification performance because the homography is estimated based
on their errors, and the optimal number varies with the image
resolution. A special case of the USR-CGD algorithm is to turn off all
geometric distortion terms, which is called the USR algorithm.  We will
study the performance of the USR and USR-CGS algorithms in Section
\ref{sec:performance}. 

\section{Databases of Uncalibrated Stereo Images}\label{sec:databases}

%%%%%%%%%%%%%%%%%%%%%%%%%%%%%%%%%%%%%%%%%%%%%%%%%%%%%%
\begin{table*}[t]
  \centering
  \caption{The databases of uncalibrated stereo images}
    \begin{tabular}{c|cc|cc}
    \toprule
    \textbf{} & \multicolumn{2}{c|}{\textbf{MCL databases}} & \multicolumn{2}{c}{\textbf{Existing databases}} \\
    \midrule
    \textbf{Name} & \textbf{MCL-SS} & \textbf{MCL-RS} & \textbf{SYNTIM~\cite{cit:SYNTIM}} & \textbf{VGS~\cite{cit:Mallon2005}} \\
    \midrule
    \multirow{2}[0]{*}{\textbf{\# of Test Images}} & 32    & \multirow{2}[0]{*}{20} & \multirow{2}[0]{*}{8} & \multirow{2}[0]{*}{6} \\
          & \shortstack{(8 different geometric\\distortions per each reference}) &       &       &  \\
    \textbf{Resolution} & 1920x1080 & 1920x1080 & 512x512 or 768x576 & 640x480 \\
    \multirow{2}[0]{*}{\textbf{Remark}} & \multirow{2}[0]{*}{\shortstack{- CG-images generated from\\OpenGL programming}} & - 5 indoor scenes & - 7 indoor scenes & - 2 indoor scenes \\
	& & - 15 outdoor scenes & - 1 outdoor scene & - 4 outdoor scenes \\
    \bottomrule
    \end{tabular}%
  \label{tab:Databases}%
\end{table*}%
%%%%%%%%%%%%%%%%%%%%%%%%%%%%%%%%%%%%%%%%%%%%%%%%%%%%%%

The number of publicly available databases of uncalibrated stereo images
is small. The SYNTIM database \cite{cit:SYNTIM} consists of eight stereo
images (seven indoor scenes and one outdoor scene) of two resolutions,
512x512 and 768x576. The VSG database~\cite{cit:Mallon2005} consists of
six stereo images (two indoor scenes and four outdoor scenes) of
resolution 640x480.  Although these two databases contain multiple
unrectified stereo images taken by different camera poses, it is
difficult to analyze the effect of each pose on the quality of rectified
images. To allow in-depth analysis, we built two high quality databases
of uncalibrated stereo images on our own in this research.

\begin{figure}[t]
\centering
\includegraphics[width=0.4\textwidth]{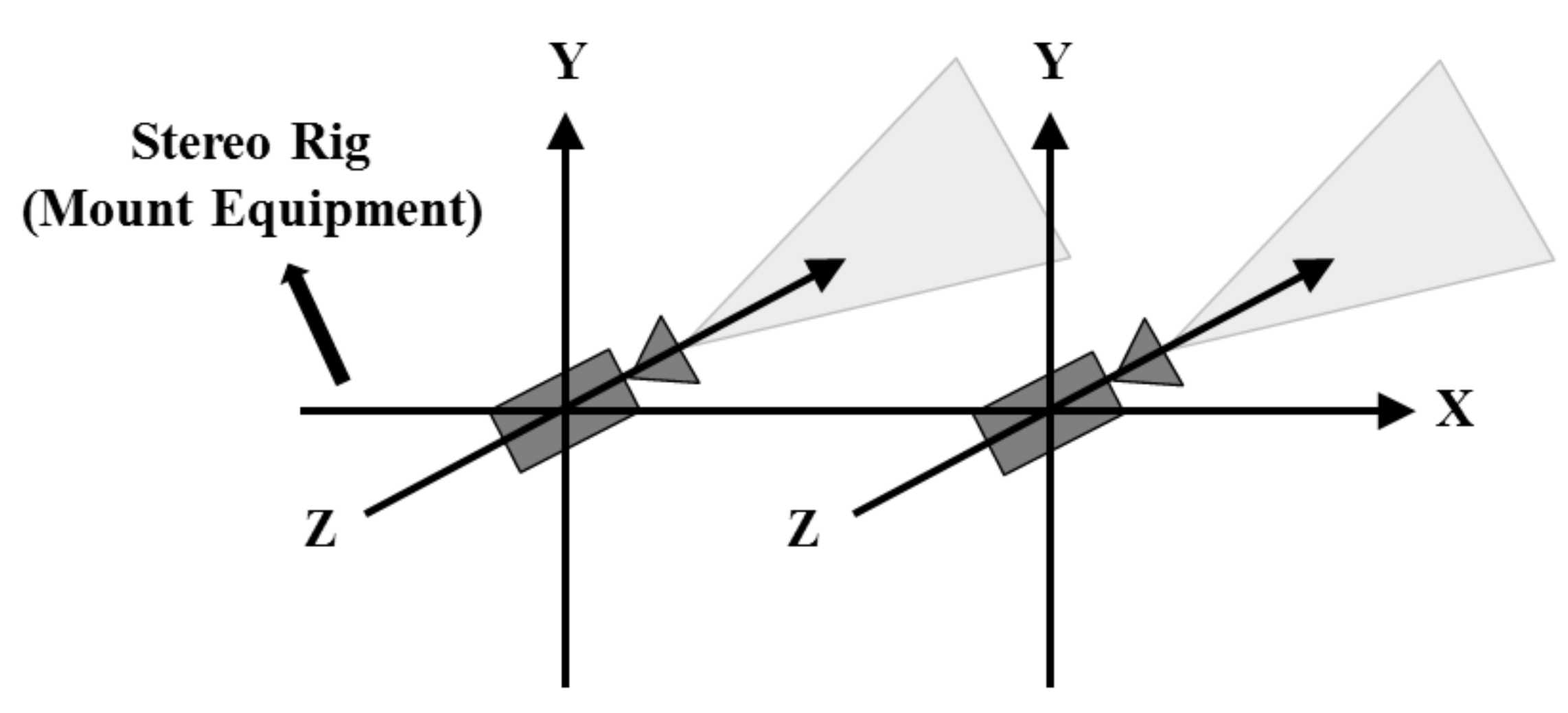}
\caption{Camera configuration used to acquire synthetic stereo images.}
\label{fig:MCLSScamera}
\end{figure}

First, we built a synthetic stereo image database consisting of 32 image
pairs using the OpenGL programming. The left images of 4 reference stereo 
pairs are shown in Fig.~\ref{fig:MCLSSref}. The advantage of using a synthetic
image is that it is easy to generate a specific geometric distortion.
Given two cameras whose initial settings are shown in
Fig.~\ref{fig:MCLSScamera}, we can translate or rotate each of cameras
along the X-/Y-/Z-axis so as to mimic real world camera configurations
such as a converged cameras, a wide baseline, vertical
misalignment, and different zoom levels.  For each of reference stereo
image pairs, we generate 8 test stereo pairs, where 6 of them are
obtained by applying a single geometric distortion while 2 are generated
by applying all six geometric distortions together.  For the latter, we
move left and right cameras along the X and Y axes and the range of
increased disparity is 0$\sim$185 pixels.  The ratio of object's sizes
between left and right images due to cameras' translation on the Z-axis
is about 11.98$\sim$31.24\%. Finally, the angle difference due to camera
rotation is $10\degree$ on each axis.  Eight such image pairs of image
``Interior" are given in Fig.~\ref{fig:MCLSStest}.  More stereo image
pairs can be found in ~\cite{cit:MCLdatabase}. 

%%%%%%%%%%%%%%%%%%%%%%%%%%%%%%%%%%%%%%%%%%%%%%%%%%%%%%
%%%%%%%%%%%%%%%%%%%%%%%%%%%%%%%%%%%%%%%%%%%%%%%%%%%%%%

Next, we built a database of unrectified stereo images by capturing
various scenes from the real world environment.  It consists of 5 indoor
scenes and 15 outdoor scenes. The left images of 8 selected scenes are
shown in Fig.~\ref{fig:MCLRSref}. The scenes were taken with different
zoom levels and various poses using two Point Grey Flea3 cameras with an
additional zoom lens. To meet the recommendation of the current
broadcasting standard, all images were captured in full-HD resolution
(1920x1080). For the rest of this paper, we call our synthetic database
MCL-SS (synthetic stereo) and the real database MCL-RS (real stereo),
where MCL is the acronym of the Media Communication Lab at the
University of Southern California.  The comparison with existing
databases is summarized in Table~\ref{tab:Databases}. The MCL-SS and
MCL-RS databases are accessible to the public~\cite{cit:MCLdatabase}. 

\section{Performance Evaluation}\label{sec:performance}

%%%%%%%%%%%%%%%%%%%%%%%%%%%%%%%%%%%%%%%%%%%%%%%%%%%%%%%%
\begin{table*}[t]
\centering
\caption{Performance comparison of six rectification algorithms in terms
of the rectification error ($E_v$) and orthogonality ($E_O$) and Skewness 
($E_{Sk}$) geometric distortions, where the best result is shown in bold. 
For the MCL-SS database, the results of four test sets (Interior, Street, 
Military, and Vehicle) are averaged.}
    \scalebox{0.7}{
    \begin{tabular}{cc|rrrrrr|rrrrrr|rrrrrr}
    \toprule
    \multirow{5}[0]{*}{DATABASE} & \multirow{5}[0]{*}{TEST SET} & \multicolumn{6}{c}{\multirow{2}[0]{*}{RECTIFICATION ERROR ($E_v$ in pixels)}} & \multicolumn{12}{|c}{GEOMETRIC DISTORTIONS (Ideal Value)} \\
          &       & \multicolumn{6}{c}{}                          & \multicolumn{6}{|c}{Othogonality (90\degree)}        & \multicolumn{6}{c}{Skewness (0\degree˚)} \\
\cline{3-20}
	   &       & & & & & &
		      & & & & & &
		      & & & & & &\\
	   &       & Hartley & Mallon & Wu & Fusiello &USR & USR-
		      & Hartley & Mallon & Wu & Fusiello &USR & USR-
		      & Hartley & Mallon & Wu & Fusiello &USR & USR-\\
	   &       & & & & & &CGD
		      & & & & & &CGD
		      & & & & & &CGD\\
\hline \hline
    \multirow{9}[0]{*}{MCL-SS} & X-translation & 0.09  & 0.13  & 144.83  & 0.10  & 0.09  & \textbf{0.09}  & 91.44  & 89.96  & 66.47  & 90.01  & 89.95  & \textbf{89.95}  & 1.63  & 0.36  & 55.97  & \textbf{0.05}  & 0.34 & 0.34  \\
          & Y-translation & \textbf{0.09}  & 0.17  & 1252.36  & 0.17  & 0.14  & 0.69  & 90.33  & 45.98  & 97.39  & 89.81  & \textbf{90.01}  & 90.19  & \textbf{0.48}  & 44.02  & 54.23  & 0.67  & 0.77  & 1.69  \\
          & Z-translation & \textbf{0.15}  & 77.26  & 960.16  & 0.94  & 0.24  & 0.79  & 100.38  & 133.94  & 116.78  & 92.58  & 97.74  & \textbf{90.95}  & 27.73  & 61.14  & 48.30  & 5.80  & 16.56  & \textbf{1.67}  \\
          & X-rotation & \textbf{0.09}  & 0.18  & 246.97  & 2.13  & 0.24  & 0.93  & 90.40  & 41.56  & 132.89  & \textbf{90.22}  & 88.17  & 89.63  & 3.75  & 48.33  & 48.48  & 6.85  & 6.25  & \textbf{3.63}  \\
          & Y-rotation & 0.11  & 0.20  & 966.68  & 0.11  & \textbf{0.08}  & 0.10  & 91.87  & 89.17  & 83.31  & 90.02  & \textbf{89.97}  & 89.90  & 4.17  & 4.14  & 35.04  & 3.74  & 3.75  & \textbf{3.74}  \\
          & Z-rotation & 0.11  & 0.18  & 109.49  & 0.10  & 0.11  & \textbf{0.09}  & 90.35  & 92.78  & 111.08  & 89.95  & \textbf{90.02}  & 90.10  & 0.93  & 3.01  & 53.95  & \textbf{0.24}  & 0.69  & 0.70  \\
          & compound1 & 37.52  & 6443.91  & 662.73  & 3.49  & \textbf{0.10}  & 0.55  & 99.26  & 110.23  & 100.38  & 88.73  & 97.10  & \textbf{89.98}  & 39.76  & 31.13  & 61.06  & 10.48  & 18.18  & \textbf{1.45}  \\
          & compound2 & \textbf{0.10}  & 4.10  & 101.37  & 1.32  & 0.12  & 0.94  & 88.36  & 118.46  & 95.07  & 87.47  & 88.06  & \textbf{89.23}  & 22.32  & 44.67  & 52.01  & 11.97  & 10.39  & \textbf{2.61}  \\
          & Mean & 4.78 & 815.76 & 555.58 & 1.04 & \textbf{0.25} & 0.50 & 92.80 & 90.26 & 100.42 & 89.85 & 91.39 & \textbf{89.99} & 12.47 & 29.60 & 51.13 & 4.98 & 7.11 & \textbf{2.18} \\
\hline
    \multirow{21}[0]{*}{MCL-RS} & Books & \textbf{0.68}  & 25.01  & 129.41  & 3.47  & 1.97  & 1.87  & 59.91  & 78.36  & 43.74  & 84.95  & 84.93  & \textbf{91.10}  & 28.47  & 35.53  & 43.31  & 6.38  & 14.99  & \textbf{4.98}  \\
          & Dogs  & 0.21  & 0.44  & 2403.43  & 5.67  & \textbf{0.16}  & 0.17  & 93.50  & 93.33  & 48.66  & 93.57  & 89.18  & \textbf{90.21}  & 4.68  & 5.57  & 67.57  & 11.72  & 2.34  & \textbf{0.99}  \\
          & Doheny & 0.10  & 0.42  & 1.75  & 0.09  & 0.09  & \textbf{0.08}  & 83.80  & 80.84  & 83.45  & \textbf{90.02}  & 88.01  & 88.96  & 15.22  & 11.85  & 68.85  & \textbf{0.38}  & 4.98  & 2.56  \\
          & Dolls & 0.17  & 1.25  & 36.09  & 0.21  & \textbf{0.16}  & 0.20  & 92.54  & 100.68  & 156.50  & 89.30  & 93.00  & \textbf{89.69}  & 7.86  & 10.27  & 60.14  & 1.59  & 8.77  & \textbf{0.92}  \\
          & Drawer & 0.21  & \textbf{0.03}  & 18.50  & 0.47  & 0.45  & 0.20  & 104.35  & 134.93  & 83.80  & 88.83  & 92.99  & \textbf{90.04}  & 34.33  & 43.71  & 76.14  & 3.08  & 27.02  & \textbf{0.14}  \\
          & Fountain & 1.72  & 36.32  & 354.99  & 0.35  & 0.20  & \textbf{0.91}  & 88.99  & 87.31  & 105.04  & 90.40  & 90.55  & \textbf{90.32}  & 2.74  & 2.85  & 52.19  & \textbf{0.49}  & 1.34  & 0.86  \\
          & Fountain2 & 0.18  & 2.66  & \textbf{0.14}  & 2.06  & 0.20  & 0.27  & 94.98  & 98.58  & 95.85  & 82.18  & 95.36  & \textbf{90.21}  & 10.90  & 18.48  & 24.89  & 15.18  & 12.31  & \textbf{0.51}  \\
          & Fountain3 & 0.27  & 5.21  & 169.22  & 0.14  & 0.11  & \textbf{0.13}  & 93.45  & 95.26  & 133.98  & \textbf{90.16}  & 93.21  & 90.65  & 8.32  & 12.58  & 53.26  & \textbf{0.32}  & 10.39  & 1.88  \\
          & Fountain4 & 2.51  & 9.30  & 355.94  & 1.04  & 1.06  & \textbf{0.45}  & 78.85  & 50.94  & 138.19  & 87.76  & 81.52  & \textbf{90.21}  & 52.38  & 43.09  & 49.54  & 5.62  & 21.62 & \textbf{0.75}  \\
          & Keychain & 0.25  & 1.63  & 391.80  & 0.48  & \textbf{0.21}  & 0.42  & 83.90  & 87.56  & 129.47  & 89.25  & 84.64  & \textbf{89.45}  & 15.12  & 12.67  & 67.26  & 1.86  & 2.65  & \textbf{1.65}  \\
          & Keychain2 & 0.38  & 0.57  & 0.63  & 12.36  & \textbf{0.23}  & 0.31  & 88.83  & 98.50  & 70.75  & 77.29  & 88.90  & \textbf{89.00}  & 2.55  & 11.76  & 70.16  & 13.67  & 4.17  & \textbf{2.30}  \\
          & Leavey & 0.20  & 1.37  & 6.72  & 0.19  & \textbf{0.11}  & 0.19  & 90.48  & 98.29  & 63.40  & 89.85  & 88.37  & \textbf{90.47}  & 2.28  & 8.40  & 71.24  & 0.48  & 4.55  & \textbf{0.06}  \\
          & Mudhall & 0.15  & 0.97  & 0.63  & 0.12  & \textbf{0.11}  & 0.13  & 91.40  & 93.48  & 116.97  & 91.42  & 91.90  & \textbf{89.50}  & 2.04  & 3.61  & 23.49  & 3.36  & 5.11  & \textbf{1.18}  \\
          & RTCC  & 3.27  & 15.39  & 0.72  & 0.27  & \textbf{0.23}  & 0.36  & 89.62  & 103.91  & 103.32  & 89.31  & 90.64  & \textbf{89.92}  & 2.76  & 14.38  & 58.07  & 5.75  & 16.40 & \textbf{1.06}  \\
          & Salvatory & 1.53  & 7.82  & 1.30  & 0.60  & \textbf{0.20}  & 0.56  & 87.73  & 87.36  & 103.62  & 87.97  & 85.65  & \textbf{90.47}  & 11.20  & 9.44  & 72.15  & 6.18  & 5.26  & \textbf{1.11}  \\
          & Step  & 1.07  & 2.98  & 82.35  & \textbf{0.20}  & 0.78  & 0.74  & 94.67  & 81.95  & 148.84  & 96.08  & 93.40  & \textbf{89.53}  & 9.47  & 14.84  & 72.25  & 9.51  & 4.46  & \textbf{1.38}  \\
          & Tommy & 0.17  & 1.41  & 452.77  & 0.13  & 0.11  & \textbf{0.12}  & 89.22  & 96.86  & 90.07  & 89.92  & 88.03  & \textbf{90.06}  & 1.97  & 8.66  & \textbf{0.00}  & 0.28  & 7.65  & 1.07  \\
          & Tommy2 & 0.14  & 5.39  & 346.31  & 0.13  & \textbf{0.10}  & 0.11  & 91.39  & 93.90  & 31.32  & 90.34  & 90.94  & \textbf{90.15}  & 5.07  & 6.54  & 51.28  & 3.22  & 9.17  & \textbf{0.20}  \\
          & Viterbi & 0.12  & 0.33  & 0.59  & 0.12  & 0.37  & \textbf{0.11}  & 93.23  & 86.30  & 64.80  & \textbf{89.98}  & 86.17  & 90.15  & 17.93  & 27.96  & 40.49  & \textbf{0.39}  & 0.56  & 1.64  \\
          & VKC   & 0.20  & 0.62  & 2527.03  & 0.17  & 0.17  & \textbf{0.15}  & 84.99  & 81.27  & 37.77  & \textbf{90.00}  & 89.73  & 89.31  & 9.27  & 9.97  & 49.77  & \textbf{0.16}  & 4.19  & 1.58  \\
          & Mean & 0.67 & 5.96 & 364.02 & 1.41 & \textbf{0.31} & 0.38 & 88.84 & 91.48 & 92.15 & 88.92 & 89.28 & \textbf{89.96} & 12.23 & 15.61 & 55.21 & 4.48 & 8.98 & \textbf{1.34} \\
\hline
    \multirow{9}[0]{*}{SYNTIM} & Aout  & \textbf{0.33}  & 2.67  & 11.43  & 4.32  & 0.53  & 1.44  & 99.68  & 110.98  & 88.99  & 86.70  & 92.94  & \textbf{91.09}  & 21.47  & 26.12  & 65.47  & 14.92  & 11.49  & \textbf{7.41}  \\
          & BalMire & 0.46  & 3.21  & 256.49  & 0.45  & \textbf{0.22}  & 0.34  & 90.77  & 96.39  & 133.71  & 91.20  & 91.24  & \textbf{89.66}  & 11.10  & 14.24  & 52.66  & 3.87  & 11.17  & \textbf{2.95}  \\
          & BalMouss & 0.19  & 0.33  & 0.19  & 0.21  & \textbf{0.18}  & 0.26  & 91.64  & 84.67  & 88.53  & 88.68  & 88.56  & \textbf{89.46}  & 6.76  & 5.29  & 6.03  & 4.89  & 5.39  & \textbf{2.42}  \\
          & BatInria & 0.15  & 0.47  & 0.15  & 0.14  & 0.13  & \textbf{0.13}  & 90.69  & 87.31  & 89.99  & 90.00  & 89.93  & \textbf{89.98}  & 3.47  & 3.35  & 6.18  & 1.26  & 4.09  & \textbf{2.35}  \\
          & Color & 0.20  & 0.32  & 0.21  & 0.42  & \textbf{0.19}  & 0.43  & 95.56  & 85.71  & \textbf{89.99}  & 88.76  & 89.95  & 90.14  & 7.91  & 5.74  & 6.29  & 7.31  & 5.06  & \textbf{2.69}  \\
          & Rubik & 0.14  & 1.24  & 1.42  & 0.41  & 0.13  & 0.38  & 87.26  & 108.58  & 58.40  & \textbf{89.84}  & 93.97  & 91.17  & 14.24  & 17.51  & 56.94  & 0.43  & 11.74  & \textbf{0.42}  \\
          & Sport & 0.13  & 0.94  & 0.56  & \textbf{0.06}  & 0.07  & 0.08  & 73.58  & 79.25  & 135.17  & 90.84  & 87.50  & \textbf{90.00}  & 49.88  & 27.26  & 64.71  & 3.18  & 9.56  & \textbf{0.10}  \\
          & Tot   & \textbf{0.14}  & 0.80  & 0.61  & 0.24  & 0.15  & 0.73  & 83.87  & 74.82  & 63.86  & 86.20  & 86.47  & \textbf{88.06}  & 16.52  & 14.99  & 54.44  & 15.10  & 13.81  & \textbf{10.78}  \\
          & Mean & 0.22 & 1.25 & 33.88 & 0.78 & \textbf{0.20} & 0.47 & 89.06 & 90.96 & 93.58 & 89.03 & \textbf{90.07} & 89.82 & 16.42 & 14.31 & 39.09 & 6.37 & 9.04 & \textbf{3.64} \\
\hline
    \multirow{7}[0]{*}{VSG} & Arch  & 0.18  & 0.55  & 67.15  & 0.17  & \textbf{0.15}  & 0.17  & 88.63  & 85.59  & 76.44  & 90.63  & 87.94  & \textbf{90.42}  & 20.28  & 13.77  & 28.67  & 2.08  & 6.88  & \textbf{1.01}  \\
          & Drive & 28.16  & 0.56  & 44.87  & 0.17  & 0.17  & \textbf{0.16}  & 97.08  & 99.16  & 130.15  & 89.22  & 87.47  & \textbf{89.98}  & 26.40  & 19.95  & 51.11  & 4.29  & 9.94  & \textbf{0.18}  \\
          & Lab   & 0.11  & 0.27  & 1.22  & 0.09  & \textbf{0.08}  & 0.10  & 86.19  & 101.45  & 86.21  & 91.34  & 91.39  & \textbf{91.14}  & 6.81  & 11.38  & 78.11  & 4.84  & 4.84  & \textbf{4.15}  \\
          & Roof  & 0.14  & 0.14  & 86.01  & 0.62  & \textbf{0.11}  & 0.32  & 111.39  & 93.25  & 91.45  & 91.64  & \textbf{90.02}  & 90.71  & 23.81  & 7.57  & 5.93  & 3.71  & 6.24  & \textbf{3.83}  \\
          & Slate & 0.20  & 0.15  & 2.21  & 0.11  & \textbf{0.11}  & 0.18  & 88.76  & 83.66  & 88.43  & 88.92  & 88.28  & \textbf{89.98}  & 7.01  & 6.30  & 76.09  & 4.33  & 6.89  & \textbf{0.42}  \\
          & Yard  & 0.20  & 0.38  & 0.14  & 0.11  & \textbf{0.10}  & 0.13  & 94.60  & 84.42  & 87.03  & 89.28  & 88.43  & \textbf{89.73}  & 10.24  & 7.81  & 14.11  & 5.02  & 7.30  & \textbf{1.52}  \\
          & Mean & 5.12 & 0.34 & 33.60 & 0.21 & \textbf{0.12} & 0.18  & 94.44 & 91.26 & 93.28 & 90.17 & \textbf{88.92} & 90.33 & 15.76 & 11.13 & 42.34 & 4.04 & 7.01 & \textbf{1.85} \\
    \bottomrule
    \end{tabular}%
  }
  \label{tab:result1}%
\end{table*}%
%%%%%%%%%%%%%%%%%%%%%%%%%%%%%%%%%%%%%%%%%%%%%%%%%%%%%%%%

%%%%%%%%%%%%%%%%%%%%%%%%%%%%%%%%%%%%%%%%%%%%%%%%%%%%%%%%
\begin{table*}[t]
\centering
\caption{Performance comparison of six rectification algorithms in terms
of aspect-ratio ($E_{AR}$). rotation ($E_{R}$) and scale-variance
($E_{SR}$) geometic distortions, where the best result is shown in bold.
For the MCL-SS database, the results of four test sets (Interior,
Street, Military, and Vehicle) are averaged.}
    \scalebox{0.7}{
    \begin{tabular}{cc|rrrrrr|rrrrrr|rrrrrr}
    \toprule
    \multirow{5}[0]{*}{DATABASE} & \multirow{5}[0]{*}{TEST SET} & \multicolumn{18}{c}{\multirow{1}[0]{*}{GEOMETRIC DISTORTIONs (Ideal Value)}} \\
          &       & \multicolumn{6}{c}{Aspect-Ratio (1)}        & \multicolumn{6}{c}{Rotation (0\degree)}        & \multicolumn{6}{c}{Scale-Variance (1)} \\
\cline{3-20}
	   &       & & & & & &
		      & & & & & &
		      & & & & & &\\
	   &       & Hartley & Mallon & Wu & Fusiello &USR & USR-
		      & Hartley & Mallon & Wu & Fusiello &USR & USR-
		      & Hartley & Mallon & Wu & Fusiello &USR & USR-\\
	   &       & & & & & &CGD
		      & & & & & &CGD
		      & & & & & &CGD\\
\hline \hline
    \multirow{9}[0]{*}{MCL-SS} & X-translation & 1.01  & 1.01  & 95.57  & 1.00  & 1.00  & \textbf{1.00}  & 0.17  & 45.13  & 50.93  & \textbf{0.02}  & 0.17  & 0.17  & 1.00  & 1.00  & 8.43  & 1.00  & 1.00  & \textbf{1.00}  \\
          & Y-translation & 1.01  & 1.01  & 248.66  & \textbf{1.01}  & 1.02  & 1.03  & 54.17  & 93.47  & 87.24  & 53.13  & 56.87  & \textbf{42.76}  & \textbf{1.01}  & 1.25  & 288.82  & 1.02  & 1.03  & 1.02  \\
          & Z-translation & 3.11  & 2.86  & 28.94  & 2.20  & 2.82  & \textbf{1.03}  & 37.37  & 73.18  & 90.10  & 2.43  & 17.58  & \textbf{2.39}  & 4.55  & 0.21  & 27.68  & 31.65  & 12.04  & \textbf{1.12}  \\
          & X-rotation & 1.16  & 1.21  & 1231.26  & 1.23  & 2.82  & \textbf{1.15}  & 56.35  & 79.69  & 71.88  & 54.72  & 60.26  & \textbf{51.57}  & 1.03  & 1.46  & 2.17  & 1.21  & \textbf{1.01}  & 1.05  \\
          & Y-rotation & 1.08  & 1.08  & 129.90  & 1.08  & 1.08  & \textbf{1.07}  & 1.68  & 16.31  & 33.63  & 1.65  & 1.64  & \textbf{1.63}  & \textbf{1.00}  & 1.07  & 78.58  & 1.04  & 1.08  & 1.07  \\
          & Z-rotation & 1.01  & 1.01  & 417.89  & 1.00  & \textbf{1.00}  & 1.01  & 5.19  & 5.22  & 100.49  & 5.18  & 5.19  & \textbf{5.17}  & 1.01  & 0.99  & 0.40  & 1.00  & 1.00  & \textbf{1.00}  \\
          & compound1 & 5.12  & 2.63  & 400.24  & 1.38  & 1.58  & \textbf{1.03}  & 64.19  & 57.39  & 87.66  & 16.91  & 23.01  & \textbf{15.87}  & 1.36  & 5.98  & 219.74  & 1.28  & 2.47  & \textbf{1.06}  \\
          & compound2 & 3.43  & 2.57  & 426.15  & 1.34  & 1.32  & \textbf{1.08}  & 48.80  & 65.84  & 67.65  & \textbf{39.71}  & 50.57  & 46.19  & 4.60  & 1.20  & 18.05  & 1.09  & 1.44  & \textbf{0.98}  \\
          & Mean & 2.11 & 1.67 & 372.33 & 1.28 & 1.38 & \textbf{1.05} & 33.49 & 54.53 & 73.70 & 22.47 & 26.91 & \textbf{21.12} & 1.94 & 1.65 & 80.48 & 4.91 & 2.01 & \textbf{1.03} \\
\hline
    \multirow{21}[0]{*}{MCL-RS} & Books & 6.21  & 10.51  & 20.05  & 1.11  & 2.94  & \textbf{1.08}  & 81.76  & 89.62  & 70.90  & \textbf{14.34}  & 86.06  & 38.32  & \textbf{0.92}  & 80.26  & 7.06  & 1.26  & 10.41  & 1.14  \\
          & Dogs  & 1.05  & 1.05  & 3.32  & 2.14  & 1.04  & \textbf{1.02}  & 8.99  & 7.51  & 90.79  & 4.72  & 8.89  & \textbf{8.88}  & 1.19  & \textbf{0.97}  & 0.93  & 2.13  & 0.83  & 0.84  \\
          & Doheny & 1.31  & 1.31  & 56.69  & \textbf{1.01}  & 1.10  & 1.05  & 7.25  & 86.51  & 16.47  & 7.36  & \textbf{6.69}  & 7.16  & 0.66  & 1.52  & 0.67  & 1.00  & \textbf{1.00}  & 0.98  \\
          & Dolls & 1.19  & 1.19  & 4.40  & 1.03  & 1.24  & \textbf{1.03}  & 8.32  & 9.43  & 148.80  & 7.22  & 8.21  & \textbf{3.26}  & 1.34  & 0.75  & 0.43  & \textbf{1.01}  & 1.57  & 0.95  \\
          & Drawer & 15.93  & 7.48  & 29.39  & 1.13  & 11.79  & \textbf{1.00}  & 22.66  & 67.89  & 82.86  & 0.80  & 64.30  & \textbf{0.05}  & 81.95  & 0.16  & 1.68  & 2.99  & 12.54  & \textbf{0.85}  \\
          & Fountain & 1.05  & 1.06  & 63.33  & 1.04  & 1.03  & \textbf{1.03}  & \textbf{4.83}  & 34.04  & 73.16  & 61.09  & 99.08  & 20.80  & 0.94  & 1.09  & 4.24  & 1.03  & 1.04  & \textbf{1.02}  \\
          & Fountain2 & 1.34  & 1.34  & 1.85  & 1.75  & 1.75  & \textbf{1.02}  & 7.26  & 87.23  & 13.88  & 7.53  & 6.95  & \textbf{2.35}  & 1.30  & 0.81  & 1.75  & 4.11  & 7.57  & \textbf{1.12}  \\
          & Fountain3 & 1.30  & 1.29  & 896.51  & \textbf{1.01}  & 2.26  & 1.04  & \textbf{5.44}  & 60.61  & 46.86  & 6.32  & 8.73  & 5.66  & 1.44  & 0.80  & 3.93  & \textbf{1.00}  & 11.96  & 1.07  \\
          & Fountain4 & 20.33  & 3.62  & 18.19  & 1.21  & 2.78  & \textbf{1.01}  & 43.01  & 163.92  & 104.06  & 4.39  & 33.87  & \textbf{4.24}  & 8.14  & 4.46  & 9.01  & 2.88  & 9.34  & \textbf{0.80}  \\
          & Keychain & 1.35  & 1.34  & 2589.48  & \textbf{1.04}  & 1.33  & 1.09  & 4.78  & \textbf{4.34}  & 115.30  & 7.53  & 4.52  & 7.42  & 0.81  & 1.63  & 101.40  & 0.99  & 1.32  & \textbf{0.98}  \\
          & Keychain2 & 1.05  & 1.05  & 138.23  & 1.17  & 1.05  & \textbf{1.04}  & 9.68  & 36.37  & 81.10  & \textbf{2.13}  & 9.68  & 9.62  & 0.77  & \textbf{1.24}  & 0.84  & 4.30  & 1.41  & 1.41  \\
          & Leavey & 1.05  & 1.06  & 35.07  & 1.01  & 1.10  & \textbf{1.00}  & 13.26  & 45.83  & 57.89  & 18.73  & 26.34  & \textbf{5.70}  & 1.04  & 0.96  & 1.91  & 1.00  & 1.03  & \textbf{1.00}  \\
          & Mudhall & \textbf{1.04}  & 1.04  & 141.11  & 1.08  & 1.10  & 1.06  & 4.76  & 4.98  & 10.76  & 4.02  & 5.43  & \textbf{3.91}  & \textbf{1.02}  & 0.96  & 0.66  & 1.13  & 1.27  & 1.04  \\
          & RTCC  & \textbf{1.05}  & 1.06  & 352.77  & 1.13  & 1.13  & 1.07  & \textbf{16.40}  & 38.97  & 87.48  & 27.42  & 29.73  & 25.09  & 1.02  & 0.96  & 0.00  & 1.09  & 1.24  & \textbf{0.98}  \\
          & Salvatory & 1.27  & 1.27  & 135.96  & 1.20  & 1.60  & \textbf{1.05}  & \textbf{6.98}  & 35.86  & 97.01  & 11.34  & 8.60  & 13.00  & 0.85  & 1.41  & 5.03  & 1.19  & 1.56  & \textbf{1.00}  \\
          & Step  & 1.24  & 1.24  & 1311.56  & 1.31  & 1.12 & \textbf{1.03}  & 12.60  & 84.80  & 133.00  & 37.01  & 38.13  & \textbf{11.94}  & 1.36  & 0.75  & 419.14  & 1.43  & 1.33  & \textbf{1.01}  \\
          & Tommy & 1.04  & 1.04  & \textbf{1.00}  & 1.01  & 1.09  & 1.04  & 13.87  & 39.54  & \textbf{0.00}  & 15.59  & 13.63  & 15.06  & 1.01  & 1.07  & 1.00  & \textbf{1.00}  & 0.96  & 0.98  \\
          & Tommy2 & 1.11  & 1.11  & 700.69  & 1.09  & 1.18  & \textbf{1.02}  & 6.52  & 7.03  & \textbf{5.39}  & 7.17  & 7.70  & 6.52  & 1.07  & 0.95  & 2.77  & 1.22  & 1.38  & \textbf{1.00}  \\
          & Viterbi & 30.28  & 28.85  & 2.06  & 1.03  & 31.69  & \textbf{1.03}  & 14.20  & 80.41  & 42.90  & 56.21  & \textbf{0.22}  & 1.96  & 32.15  & 0.85  & 1.76  & \textbf{1.00}  & 19.25  & 1.02  \\
          & VKC   & 1.20  & 1.20  & 124.81  & \textbf{1.02}  & 1.08  & 1.04  & 9.87  & 35.51  & 70.13  & 8.56  & \textbf{8.04}  & 8.39  & 0.86  & 1.28  & 0.19  & \textbf{1.00}  & 0.98  & 0.98  \\
          & Mean & 4.57 & 3.46 & 348.02 & 1.17 & 3.47 & \textbf{1.04} & 15.00 & 51.02 & 69.14 & 12.77 & 26.53 & \textbf{9.97} & 6.99 & 5.14 & 29.28 & 1.64 & 12.03 & \textbf{1.01} \\
\hline
    \multirow{9}[0]{*}{SYNTIM} & Aout  & 5.32  & 2.31  & 1114.30  & 3.64  & 2.19  & \textbf{1.65}  & 27.80  & 86.77  & 125.85  & \textbf{22.89}  & 33.56  & 25.99  & 2.54  & 1.20  & 0.02  & 1.25  & 2.01  & \textbf{1.08}  \\
          & BalMire & 1.40  & 1.40  & 3.59  & 1.12  & 1.77  & \textbf{1.08}  & 6.93  & 53.02  & 63.67  & 2.83  & 8.02  & \textbf{3.24}  & 1.12  & 0.99  & 1.50  & 1.37  & 4.97  & \textbf{1.08}  \\
          & BalMouss & 1.16  & 1.16  & 1.17  & 1.15  & 1.16  & \textbf{1.06}  & 5.67  & 48.26  & 4.88  & 4.16  & 4.40  & \textbf{3.03}  & 0.95  & 1.23  & 1.09  & 1.12  & 1.07  & \textbf{0.93}  \\
          & BatInria & 1.14  & 1.14  & 1.25  & 1.05  & 1.16  & \textbf{1.09}  & 0.95  & 45.93  & 1.31  & 1.70  & 1.38  & \textbf{1.26}  & 1.06  & 1.22  & 1.28  & \textbf{1.01}  & 1.18  & 1.07  \\
          & Color & 1.24  & 1.24  & 1.26  & 1.42  & 1.24  & \textbf{1.10}  & 5.15  & 47.16  & 5.05  & \textbf{3.84}  & 4.75  & 4.85  & 1.09  & 1.37  & 1.23  & 1.57  & 1.31  & \textbf{1.01}  \\
          & Rubik & 1.47  & 1.51  & 14.07  & 1.01  & 1.51  & \textbf{1.01}  & 17.53  & 89.82  & 87.59  & \textbf{7.70}  & 13.73  & 8.07  & 1.25  & 0.55  & 2.48  & 1.01  & 2.16  & \textbf{1.00}  \\
          & Sport & 5.86  & 4.42  & 21.63  & 1.09  & 1.34  & \textbf{1.00}  & 13.84  & 47.74  & 120.32  & 1.52  & 6.21  & \textbf{0.03}  & 1.67  & 13.77  & 7.16  & 1.03  & 1.70  & \textbf{1.00}  \\
          & Tot   & 1.66  & 1.77  & 22.56  & 1.65  & 1.67  & \textbf{1.45}  & 21.75  & 91.28  & 67.78  & 19.37  & 17.26  & \textbf{16.99}  & 0.79  & 2.71  & 46.42  & \textbf{1.04}  & 1.47  & 1.32  \\
          & Mean & 2.41 & 1.87 & 147.48 & 1.52 & 1.50 & \textbf{1.18} & 12.45 & 63.75 & 59.55 & 8.00 & 11.16 & \textbf{7.93} & 1.31 & 2.88 & 7.65 & 1.18 & 1.98 & \textbf{1.06} \\
\hline
    \multirow{7}[0]{*}{VSG} & Arch  & 10.88  & 3.70  & 1.52  & 1.06  & 1.24  & \textbf{1.03}  & 6.62  & 32.15  & 57.21  & 3.52  & \textbf{2.87}  & 3.46  & 3.08  & 13.50  & 1.07  & 1.02  & 1.30  & \textbf{1.02}  \\
          & Drive & 11.66  & 1.60  & 34.84  & 1.16  & 1.45  & \textbf{1.00}  & 32.70  & 46.59  & 93.17  & 3.86  & 5.85  & \textbf{1.62}  & 3.01  & 1.53  & 7.18  & 1.42  & 2.72  & \textbf{1.00}  \\
          & Lab   & 1.13  & 1.14  & 134.37  & 1.14  & 1.14  & \textbf{1.11}  & 16.57  & 19.62  & 86.08  & 16.86  & 16.71  & \textbf{16.44}  & 1.11  & 0.86  & 26.94  & \textbf{1.06}  & 1.16  & 1.14  \\
          & Roof  & 1.20  & 1.20  & 1.18  & 1.12  & 1.19  & \textbf{1.15}  & 1.27  & 89.40  & \textbf{1.01}  & 1.09  & 1.16  & 1.38  & 1.15  & 0.75  & 0.85  & 1.60  & 1.31  & \textbf{1.14}  \\
          & Slate & 1.20  & 1.20  & 25.26  & 1.12  & 1.21  & \textbf{1.01}  & 10.86  & 36.13  & 89.51  & 10.12  & 6.12  & \textbf{5.69}  & 0.89  & 1.32  & 2.28  & \textbf{0.99}  & 1.16  & 0.98  \\
          & Yard  & 1.27  & 1.27  & 1.42  & 1.26  & \textbf{1.03}  & 1.04  & 2.72  & 30.94  & 3.28  & \textbf{2.30}  & 2.42  & 2.49  & 0.92  & 1.46  & 1.46  & 1.09  & 1.40  & \textbf{1.01}  \\
          & Mean & 4.56 & 1.68 & 33.10 & 1.25 & 1.06 & \textbf{1.05} & 11.76 & 42.47 & 55.04 & 6.29 & 6.51 & \textbf{5.18} & 1.69 & 3.24 & 6.64 & 1.20 & 1.51 & \textbf{1.05} \\
    \bottomrule
    \end{tabular}%
  }
  \label{tab:result2}%
\end{table*}%
%%%%%%%%%%%%%%%%%%%%%%%%%%%%%%%%%%%%%%%%%%%%%%%%%%%%%%%%

To evaluate the performance of the proposed USR and USR-CGD algorithms,
we compare them with four rectification algorithms over the four
databases described in Section~\ref{sec:databases}.  The four benchmark
algorithms are Hartley~\cite{cit:Hartley2003}, Mallon and
Whelan~\cite{cit:Mallon2005}, Wu and Yu~\cite{cit:Wu2005} and Fusiello
and Luca~\cite{cit:Fusiello2011}. We adopt the following vertical
disparity error
\begin{equation} \label{eq:Ev}
E_v = \frac{1}{N}\sum_{j=1}^{N}(E_v)_j,
\end{equation}
where 
\begin{equation} 
(E_v)_j = |(H_lm_l^j)_2-(H_rm_r^j)_2|,
\end{equation}
as the objective error measure while orthogonality ($E_O$), aspect-ratio
($E_{AR}$), skewness ($E_{Sk}$), rotation ($E_{R}$), and scale-variance
($E_{SV}$) are used as subjective error measures.  For the number of
correspondences, different numbers of matching points are adopted for
different image resolutions; namely, 300 for the MCL-SS and the MCL-RS
databases (1920x1080), 80 for the SYNTIM database (512x512 or 768x576),
and 50 for the VSG database (640x480), respectively. Note that the
number of correspondences may decrease after outlier removal.  

To obtain more reliable results, we conduct experiments on each test set
with three random seeds for RANSAC, which results in different
distributions of correspondences. Then, these results are averaged for
the final performance measure. Each geometric error is the averaged
value of left and right rectified images. Extensive experimental results
are shown in Tables~\ref{tab:result1} and \ref{tab:result2}, where the
best algorithm is marked in bold. Note that these two tables contain the
same horizontal entries but different vertical entries. The
rectification error, the orthogonality and skewness distortions are
shown in Table \ref{tab:result1} while the aspect ration, rotation and
scale variance distortions are shown in Table \ref{tab:result2}. 

We first examine the advantages of the generalized homography model
without considering the geometric distortion constraints (namely, the
USR algorithm). At the bottom of each test set in Table
~\ref{tab:result1}, we show the mean performance for that particular
set.  As far as the rectificatoin error is concerned, we see that the
USR algorithm provides the smallest error for all four databases.

The results of the MCL-SS database allow us to analyze the performance
improvement. To give an example, the vertical disparity is often caused
by translational distortion in the Y-axis, rotational distortion in the
X-axis and different zoom levels in the Z-axis (emulated by
translational distortion). By comparing our generalized homography model
with that of Fuesillo and Luca~\cite{cit:Fusiello2011}, our model
provides smaller errors in all three cases since it includes additional
degrees of freedom to compensate geometric errors. Furthermore, two
stereo paris with compound distortions are challenging to existing
algorithms while the USR algorithm offers a stable and robust
performance across all test pairs since the generalized homography model
is able to cope with combined geometric distortions. 

\begin{figure}[t]
	\centering
	\begin{subfigure}[h!]{0.45\textwidth}
		\centering
		\includegraphics[width=3in]{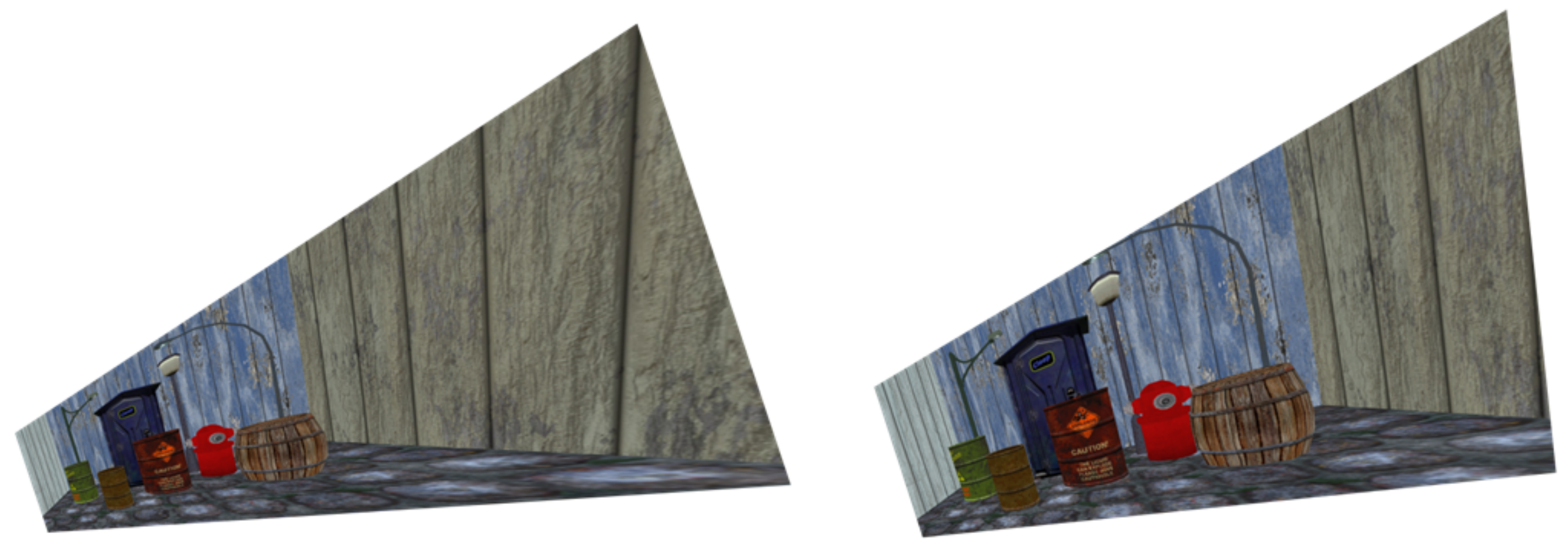} % Example image
		\caption{The rectified image pair using the USR algorithm ($E_v$=0.12)}
		\label{fig:all1}
	\end{subfigure}\\

	\begin{subfigure}[h!]{0.45\textwidth}
		\centering
		\includegraphics[width=3in]{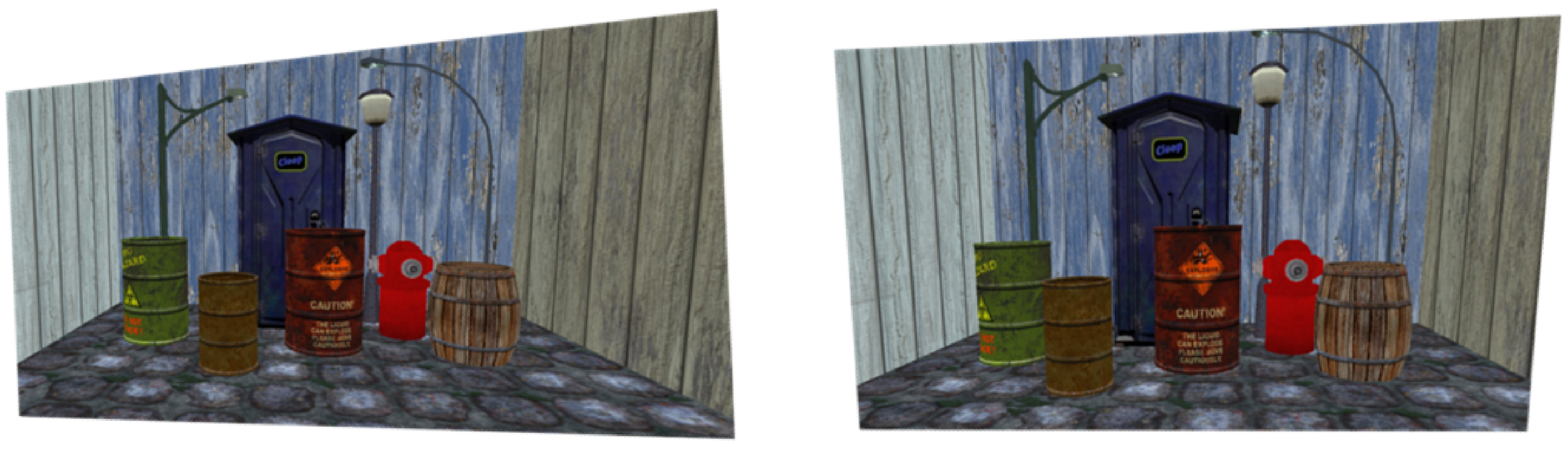} % Example image
		\caption{The rectified image pair using the USR-CGD algorithm ($E_v$=0.34)}
		\label{fig:all2}
	\end{subfigure}
\caption{Subjective quality comparison of rectified image pair 
using (a) the USR algorithm and (b) the USR-CGD algorithm.}\label{fig:all}
\end{figure}

On the other hand, the USR algorithm does not offer the best performance
in geometric distortions since it does not take them into consideration.
As shown in Tables~\ref{tab:result1} and \ref{tab:result2}, the USR-CGD
algorithm has the lowest geometric distortions in most cases.  For
subjective quality comparison, we show a rectified image pair using the
USR and USR-CGD algorithms in Fig.~\ref{fig:all}.  Although the
rectification error ($E_v$) of the USR-CGD algorithm increases slightly
(from 0.12 to 0.34 pixel), its resulting image pair look much more
natural since all geometric errors are reduced significantly. 

Generally speaking, the rectification error of the USR-CGD algorithm is
comparable with that of the USR algorithm, and the increase of $E_v$ in
the USR-CGD algorithm is around $0.06\sim0.27$ pixels.  Among the six
benchmarking algorithms, only USR and USR-CGD achieve a mean
rectification error ($E_v$) less than 0.5 pixel, which is the minimum
requirement for stereo matching to be performed along the 1D scanline. 

%%%%%%%%%%%%%%%%%%%%%%%%%%%%%%%%%%%%%%%%%%%%%%%%%%%%%%%%
%%%%%%%%%%%%%%%%%%%%%%%%%%%%%%%%%%%%%%%%%%%%%%%%%%%%%%%%

%%%%%%%%%%%%%%%%%%%%%%%%%%%%%%%%%%%%%%%%%%%%%%%%%%%%%%%%
\begin{table*}[t]
  \centering
  \caption{Performance comparison based on 100 and 300 correspondences as the input.}
    \begin{tabular}{rc|rrrrrr|rrrrrr}
    \toprule
     \multicolumn{2}{c|}{Database}       & \multicolumn{6}{c|}{MCL-SS}                    & \multicolumn{6}{c}{MCL-RS} \\
\hline
    \multicolumn{1}{c}{\multirow{2}[0]{*}{Algorithm}} & \# of & \multicolumn{1}{c}{\multirow{2}[0]{*}{$E_v$}} & \multicolumn{5}{|c|}{Geometric Errors}  & \multicolumn{1}{c}{\multirow{2}[0]{*}{$E_v$}} & \multicolumn{5}{|c}{Geometric Errors} \\
    \multicolumn{1}{c}{} & Correspondences & \multicolumn{1}{c}{} & \multicolumn{1}{|c}{$E_O$} & \multicolumn{1}{c}{$E_{Sk}$} & \multicolumn{1}{c}{$E_{AR}$} & \multicolumn{1}{c}{$E_R$} & \multicolumn{1}{c|}{$E_{SR}$} & \multicolumn{1}{c}{} & \multicolumn{1}{|c}{$E_O$} & \multicolumn{1}{c}{$E_{Sk}$} & \multicolumn{1}{c}{$E_{AR}$} & \multicolumn{1}{c}{$E_R$} & \multicolumn{1}{c}{$E_{SR}$} \\
    \midrule     \midrule
    \multicolumn{1}{c}{\multirow{2}[0]{*}{Hartley}} & 300   & 4.78  & 92.80  & 12.47  & 2.11  & 33.49  & 1.94  & 0.67  & 88.84  & 12.23  & 4.57  & 15.00  & 6.99  \\
    \multicolumn{1}{c}{} & 100   & 2202.55  & 93.12  & 15.24  & 8.90  & 38.42  & 18.59  & 110.13  & 86.55  & 16.75  & 3.07  & 24.96  & 3.24  \\
\hline
    \multicolumn{1}{c}{\multirow{2}[0]{*}{Mallon}} & 300   & 815.76  & 90.26  & 29.60  & 1.67  & 54.53  & 1.65  & 5.96  & 91.48  & 15.61  & 3.46  & 51.02  & 5.14  \\
    \multicolumn{1}{c}{} & 100   & 13266.94  & 87.72  & 31.39  & 2.30  & 59.87  & 1.94  & 19.72  & 88.98  & 19.97  & 2.27  & 51.77  & 2.23  \\
\hline
    \multicolumn{1}{c}{\multirow{2}[0]{*}{Wu}} & 300   & 555.58  & 100.42  & 51.13  & 372.33  & 73.70  & 80.48  & 364.02  & 92.15  & 55.21  & 348.02  & 69.14  & 29.28  \\
    \multicolumn{1}{c}{} & 100   & 1039.03  & 94.99  & 48.54  & 1070.51  & 82.34  & 9.47  & 1231.28  & 83.41  & 60.53  & 400.13  & 74.88  & 150.88  \\
\hline
    \multicolumn{1}{c}{\multirow{2}[0]{*}{Fusiello}} & 300   & 1.04  & 89.85  & 4.98  & 1.28  & 21.47  & 4.91  & 1.41  & 88.92  & 4.48  & 1.17  & 12.77  & 1.64  \\
    \multicolumn{1}{c}{} & 100   & 119.49  & 89.19  & 5.72  & 1.40  & 20.38  & 5.56  & 1.13  & 89.17  & 3.34  & 1.11  & 13.60  & 1.56  \\
\hline
    \multicolumn{1}{c}{\multirow{2}[0]{*}{USR-CGD}} & 300   & 0.50  & 89.99  & 2.18  & 1.05  & 24.12  & 1.03  & 0.38  & 89.96  & 1.34  & 1.04  & 9.97  & 1.01  \\
    \multicolumn{1}{c}{} & 100   & 0.52  & 90.01  & 2.88  & 1.01  & 23.28  & 1.04  & 0.26  & 90.00  & 1.36  & 1.04  & 10.17  & 1.03  \\
    \bottomrule
    \end{tabular}%
  \label{tab:result3}%
\end{table*}%
%%%%%%%%%%%%%%%%%%%%%%%%%%%%%%%%%%%%%%%%%%%%%%%%%%%%%%%%

\begin{figure*}[t]
        \centering
        \begin{subfigure}[t]{0.29\textwidth}
                \centering
                \includegraphics[width=\textwidth]{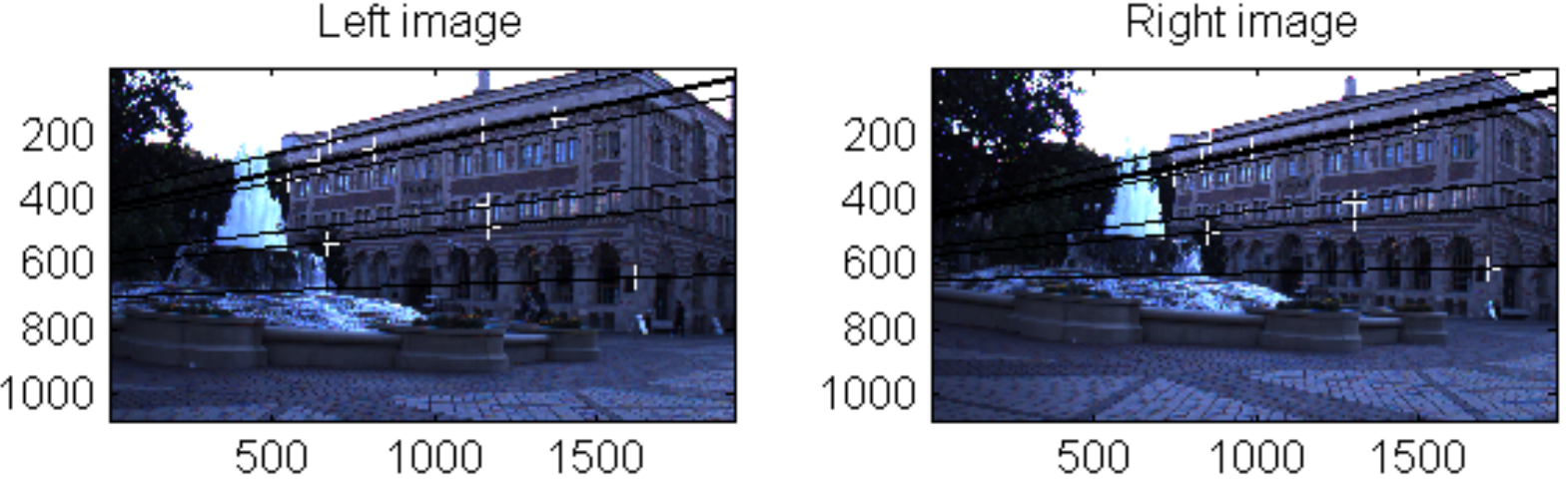}
                \caption{Original unrectified image pair}
        \end{subfigure}
		\quad
        \begin{subfigure}[t]{0.29\textwidth}
                \centering
                \includegraphics[width=\textwidth]{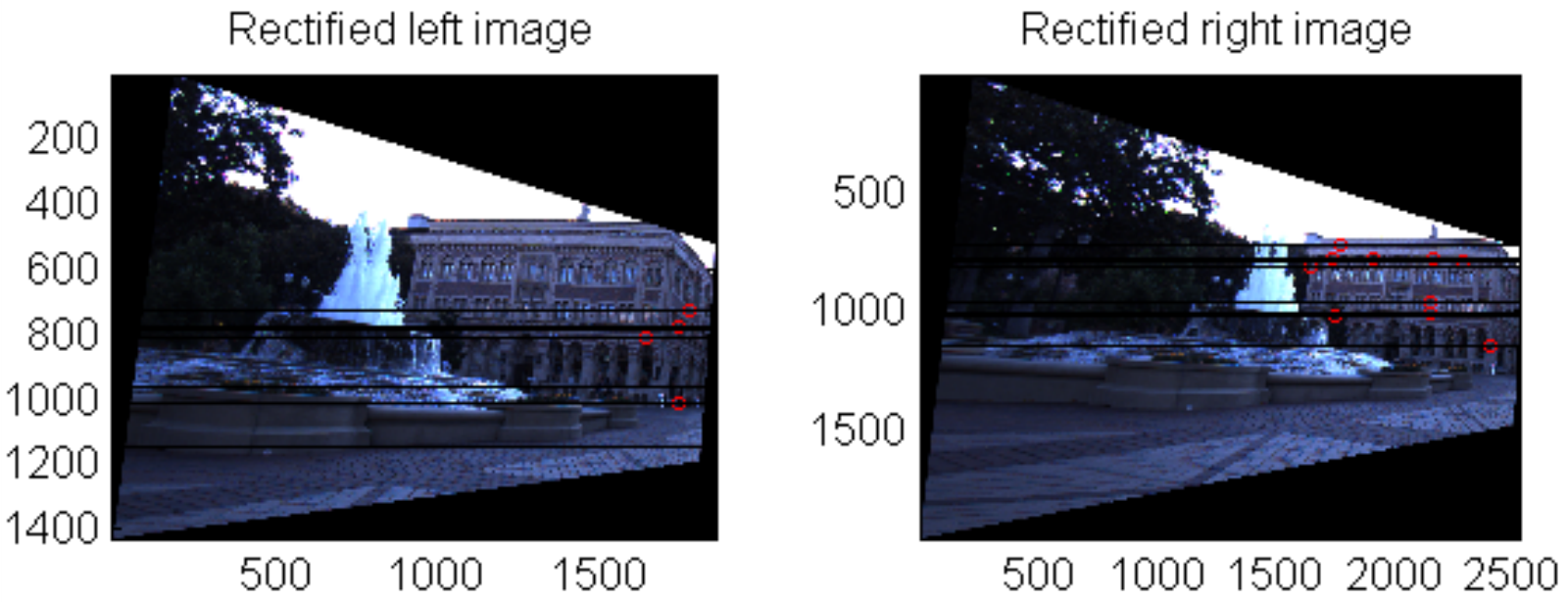}
                \caption{Hartley ($E_v$=0.146)}
        \end{subfigure}
		\quad
        \begin{subfigure}[t]{0.29\textwidth}
                \centering
                \includegraphics[width=\textwidth]{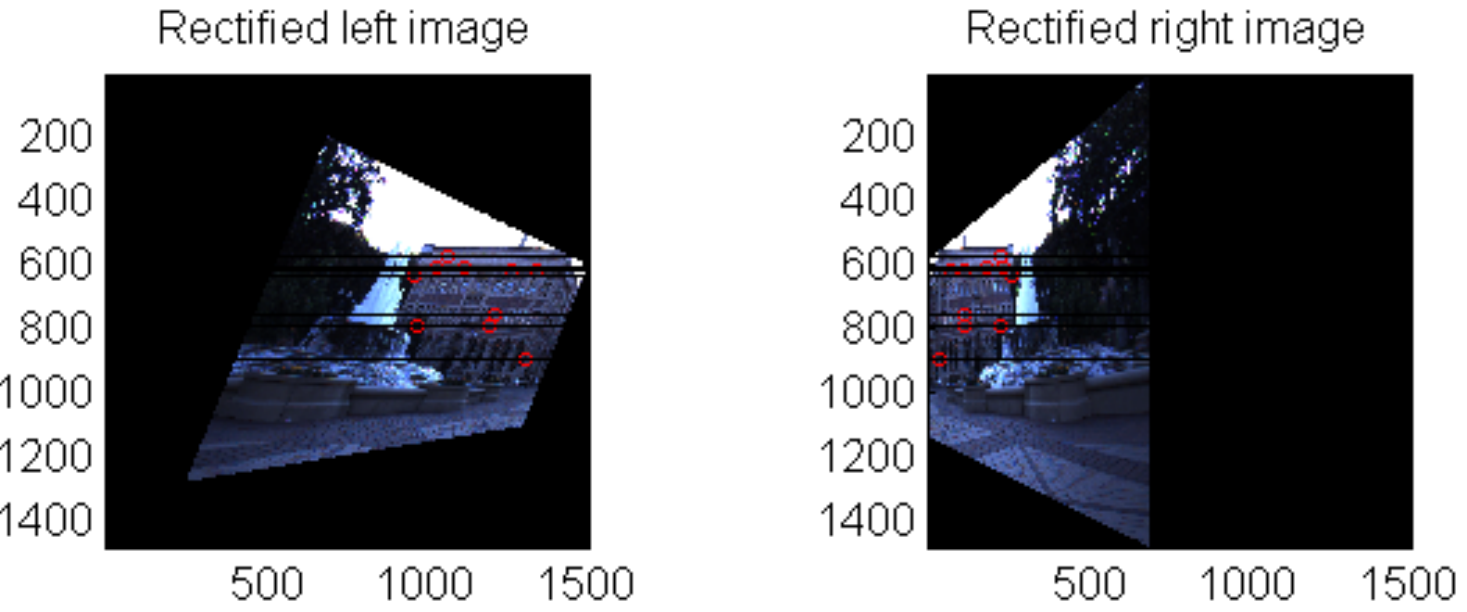}
                \caption{Mallon ($E_v$=1.504)}
        \end{subfigure}
	\\
        \begin{subfigure}[t]{0.29\textwidth}
                \centering
                \includegraphics[width=\textwidth]{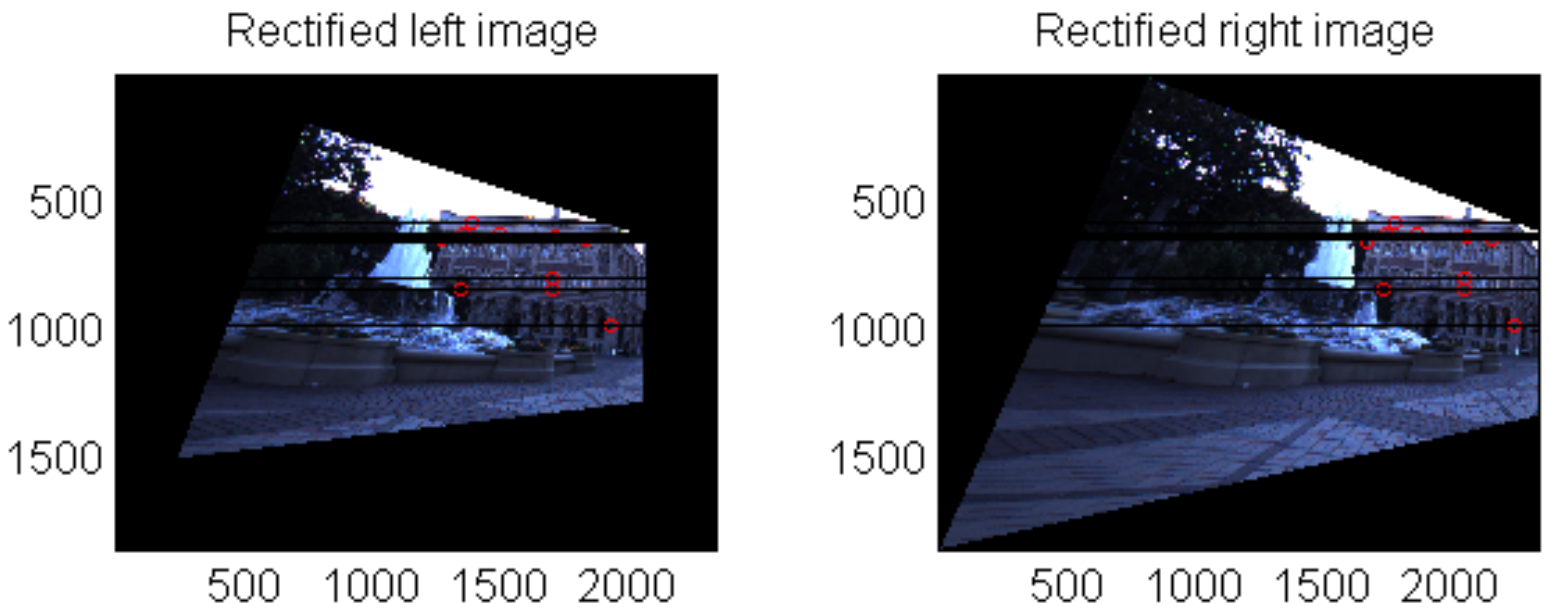}
                \caption{Wu ($E_v$=0.193)}
        \end{subfigure}
		\quad
        \begin{subfigure}[t]{0.29\textwidth}
                \centering
                \includegraphics[width=\textwidth]{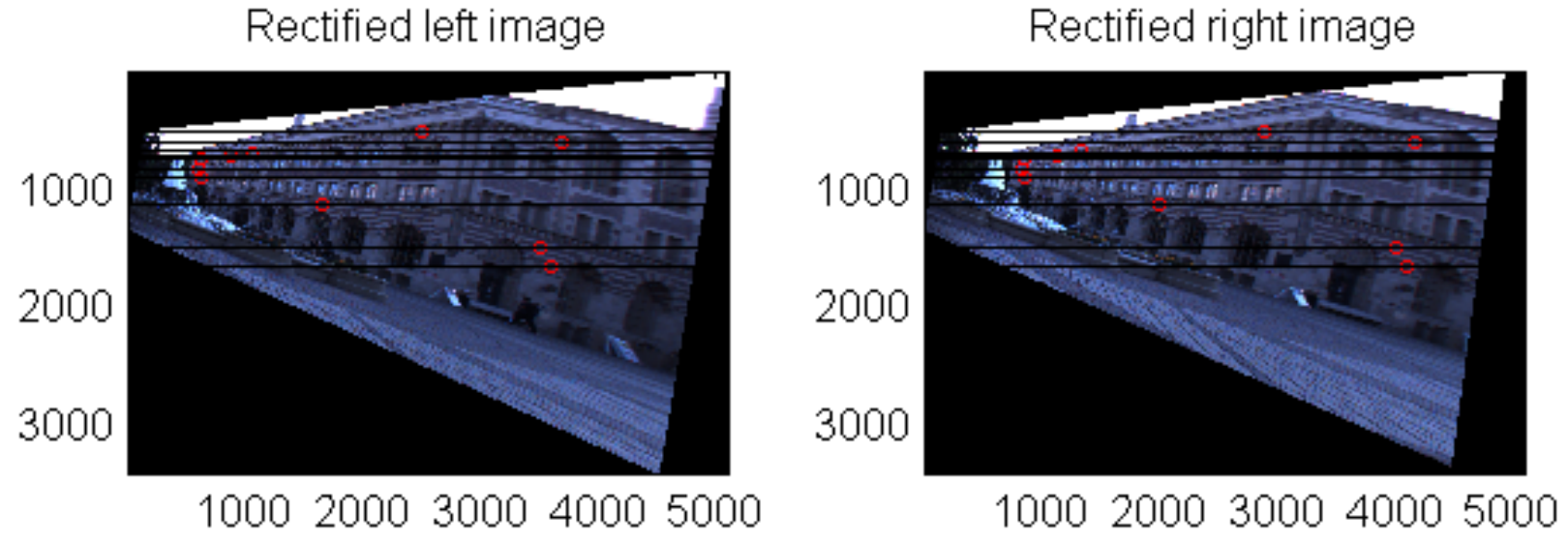}
                \caption{Fuesillo ($E_v$=2.109)}
        \end{subfigure}
		\quad
        \begin{subfigure}[t]{0.29\textwidth}
                \centering
                \includegraphics[width=\textwidth]{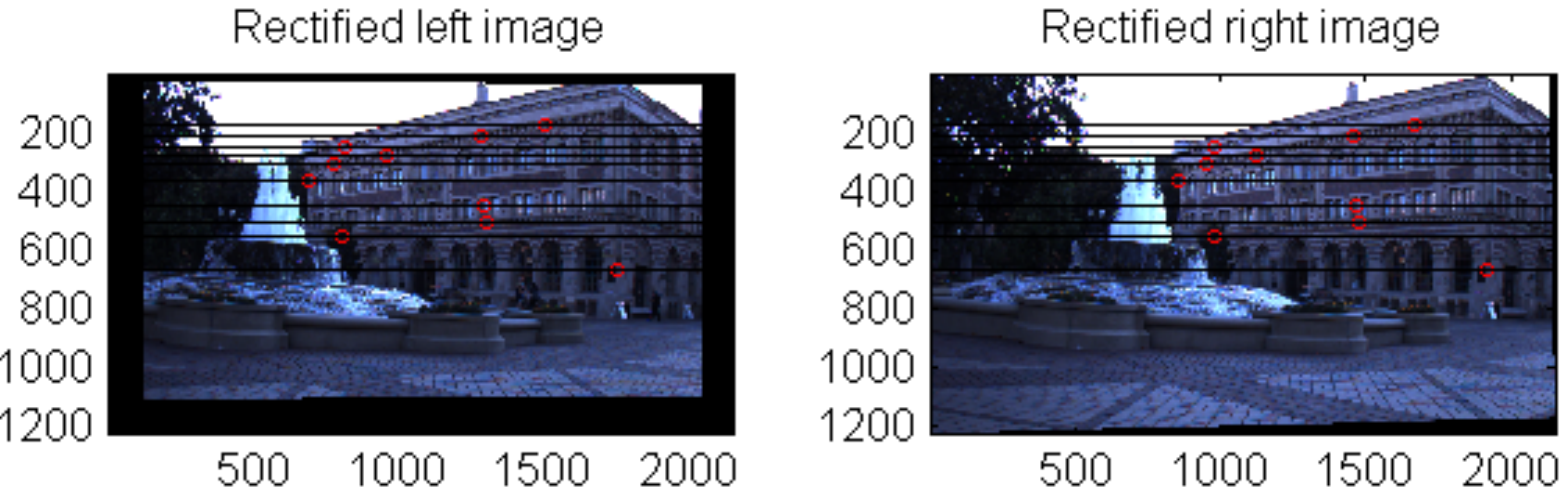}
                \caption{USR-CGD ($E_v$=0.199)}
        \end{subfigure}
\caption{Comparison of subjective quality and rectification errors for 
Fountain2 in the MCL-RS database.}\label{fig:subjectiveresult1}
\end{figure*}

\begin{figure*}[t]
        \centering
	 \begin{subfigure}[t]{0.29\textwidth}
                \centering
                \includegraphics[width=\textwidth]{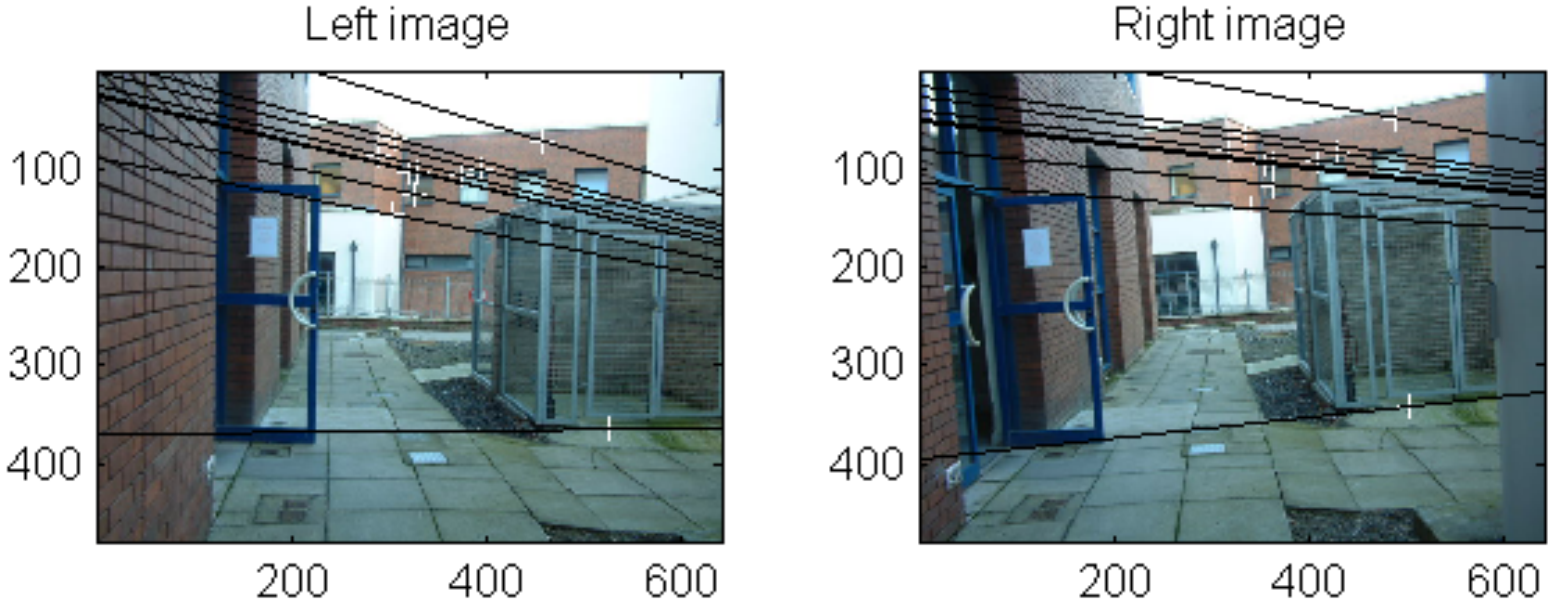}
                \caption{Original unrectified image pair}
        \end{subfigure}
		\quad
        \begin{subfigure}[t]{0.29\textwidth}
                \centering
                \includegraphics[width=\textwidth]{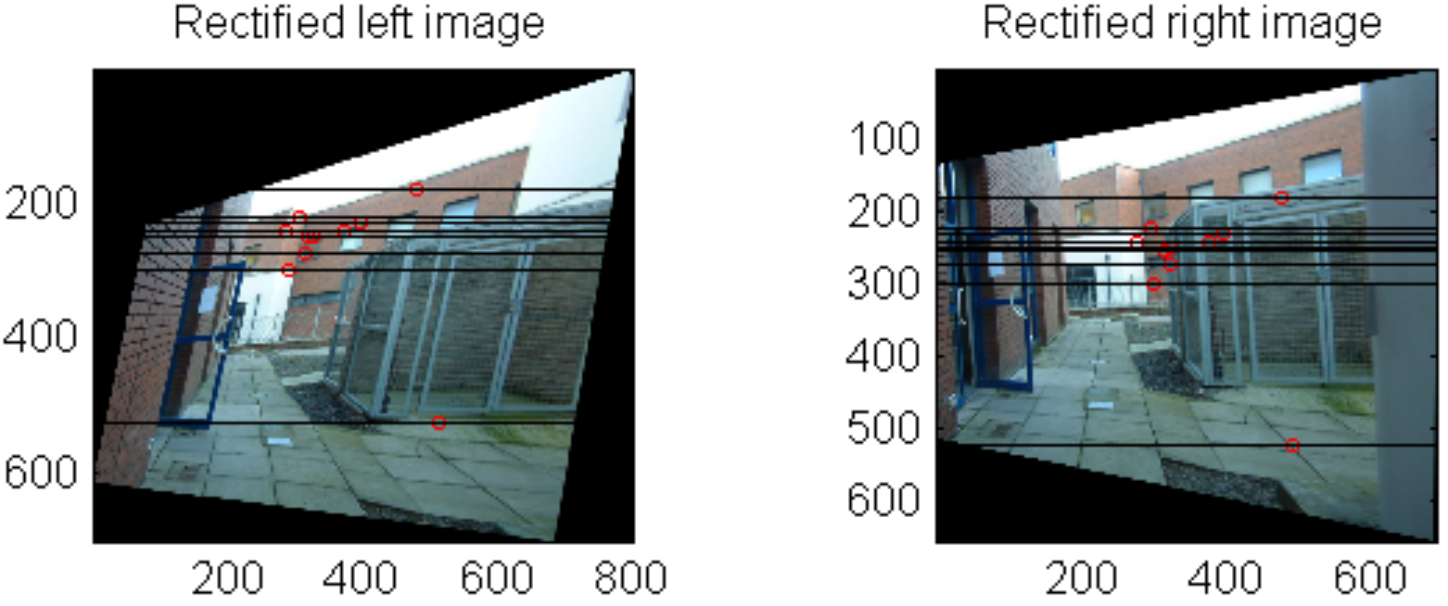}
                \caption{Hartley ($E_v$=0.098)}
        \end{subfigure}
		\quad
        \begin{subfigure}[t]{0.29\textwidth}
                \centering
                \includegraphics[width=\textwidth]{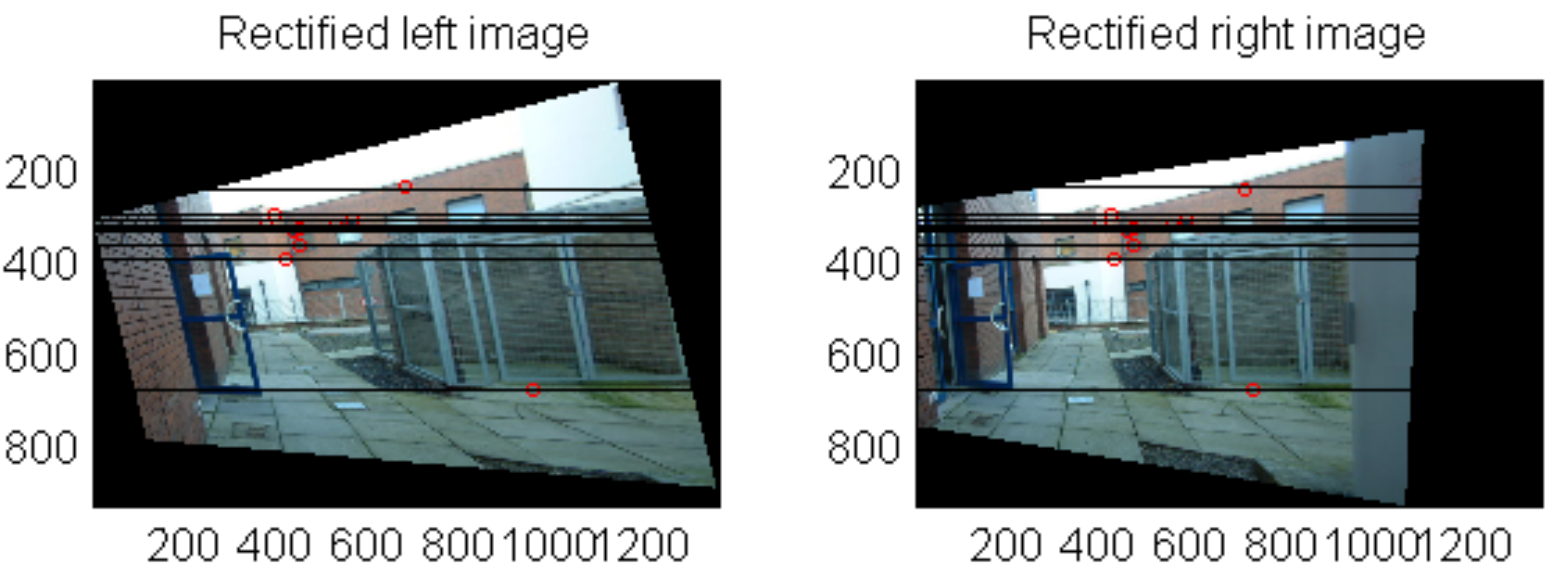}
                \caption{Mallon ($E_v$=0.385)}
        \end{subfigure}
		\\
        \begin{subfigure}[t]{0.29\textwidth}
                \centering
                \includegraphics[width=\textwidth]{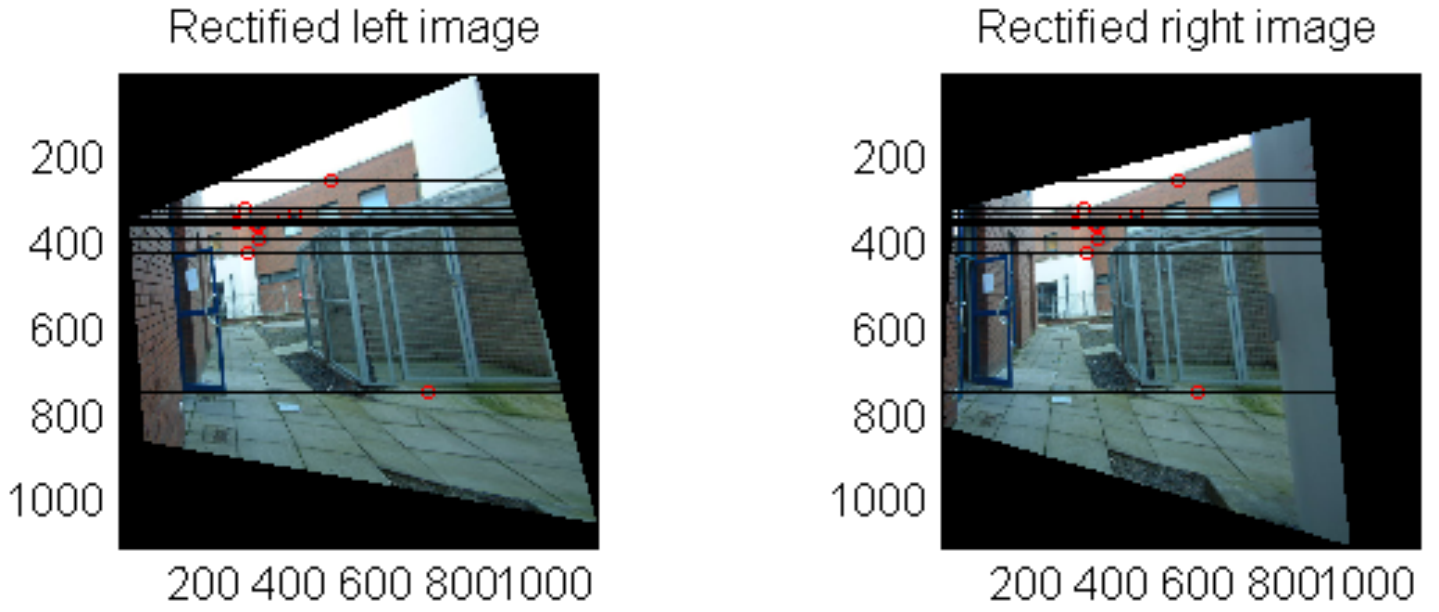}
                \caption{Wu ($E_v$=0.125)}
        \end{subfigure}
		\quad
        \begin{subfigure}[t]{0.29\textwidth}
                \centering
                \includegraphics[width=\textwidth]{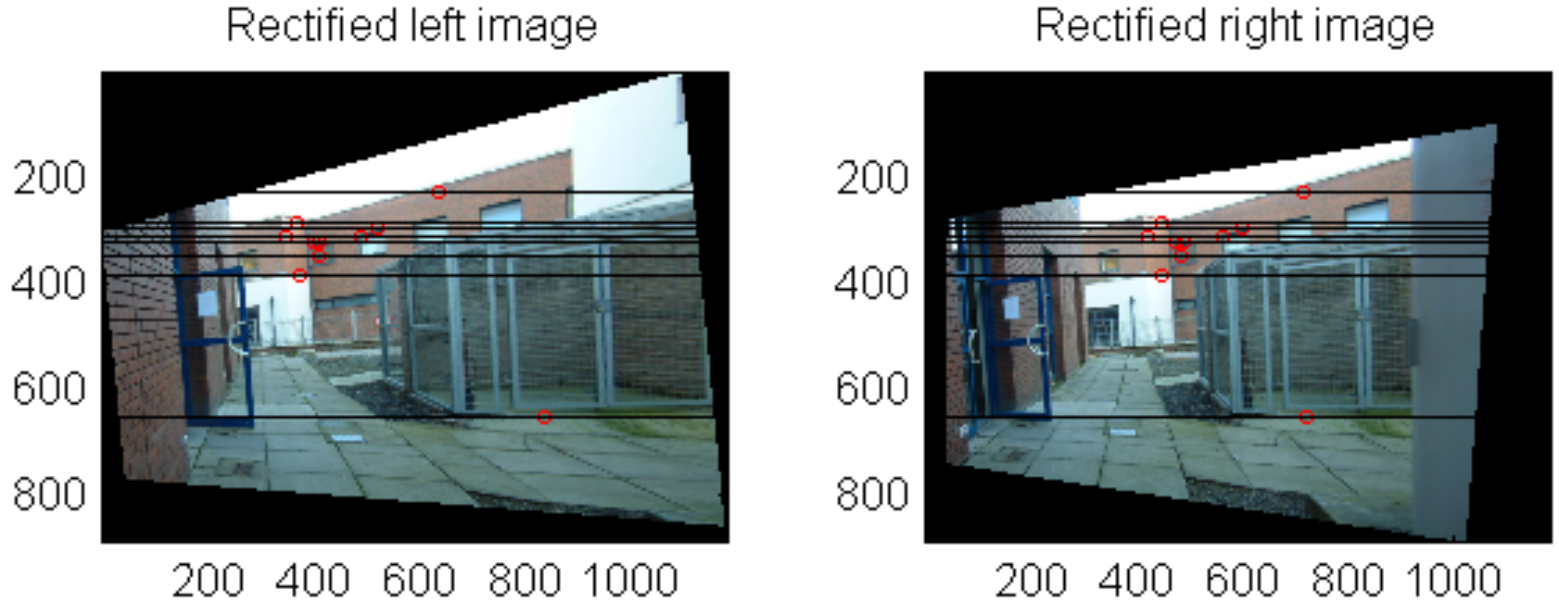}
                \caption{Fuesillo ($E_v$=0.128)}
        \end{subfigure}
		\quad
        \begin{subfigure}[t]{0.29\textwidth}
                \centering
                \includegraphics[width=\textwidth]{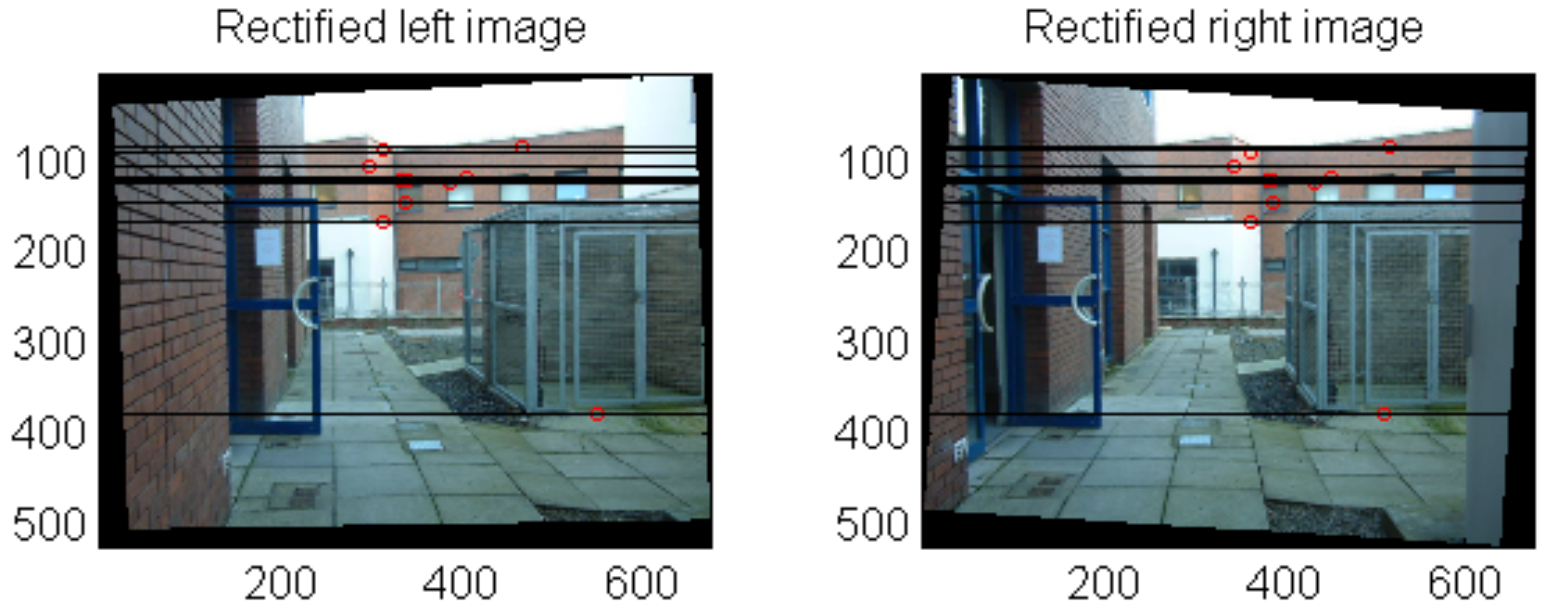}
                \caption{USR-CGD ($E_v$=0.119)}
        \end{subfigure}
\caption{Comparison of subjective quality and rectification errors for 
Yard of in the VSG database.}\label{fig:subjectiveresult6}
\end{figure*}

For the subjective quality evaluation of rectified stereo image pairs of
the proposed USR-CGD algorithm and four benchmarking algorithms, we
select six representative images pairs and show their rectified results
in Figs.~\ref{fig:subjectiveresult1}$\sim$\ref{fig:subjectiveresult6}.
The vertical disparity error $E_v$ is also listed for each subfigure.
For the purpose of displaying $E_v$, we choose ten sample matching points, 
which are evenly distributed along the vertical direction and draw corresponding
epipolar lines. It is apparent that the proposed USR-CGD algorithm have the 
best visual quality with minimal warping distortion at the cost of slightly higher 
rectification errors in some cases. 

There is another advantage of the USR-CGD algorithm. That is, it is able
to provide robust performance when the number of correspondences is
smaller. To demonstrate this, we conduct the same test with 100
correspondences for the MCL-SS and MCL-RS databases. The performance
comparison of using 300 and 100 correspondences is shown in
Table~\ref{tab:result3}. The rectification errors of all four
benchmarking algorithms increase a lot since their reliability in point
matching is more sensitive to the number of correspondences. In
contrast, the USR-CGD algorithm still maintains the rectification error
at a similar level. This is because that the USR-CGD algorithm can
adaptively optimize the homographies to meet pre-set conditions.  A
similar behavior is also observed in the degree of geometric
distortions. 

Based on the above discussion, we conclude that the USR-CGD algorithm
achieves the best overall performance in finding a proper balance
between the rectification error and geometric distortions since it
includes all of them in its objective function for optimization. 

%%%%%%%%%%%%%%%%%%%%%%%%%%%%%%%%%%%%%%%%%%%%%%%%%%%%%%%%%%%%%%%%%
%%%%%%%%%%%%%%%%%%%%%%%%%%%%%%%%%%%%%%%%%%%%%%%%%%%%%%%%%%%%%%%%%

%%%%%%%%%%%%%%%%%%%%%%%%%%%%%%%%%%%%%%%%%%%%%%%%%%%%%%%%%%%%%%%%%
\begin{figure*}[t]
        \centering
	 \begin{subfigure}[t]{0.29\textwidth}
                \centering
                \includegraphics[width=\textwidth]{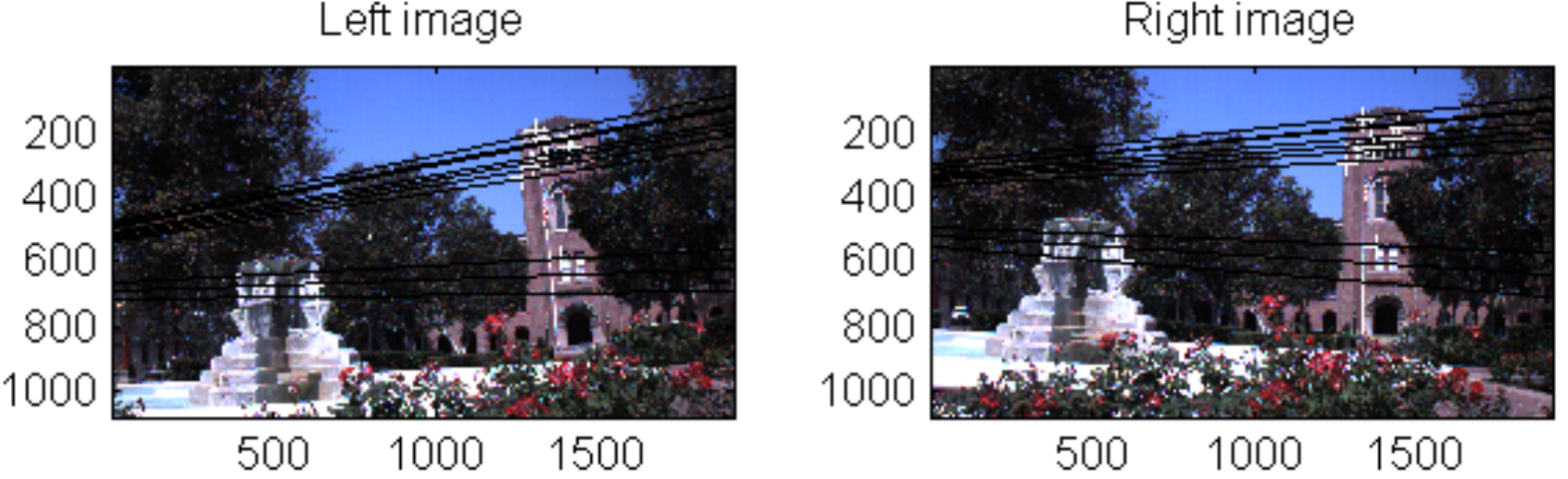}
                \caption{Original unrectified image pair}
        \end{subfigure}
		\quad
        \begin{subfigure}[t]{0.29\textwidth}
                \centering
                \includegraphics[width=\textwidth]{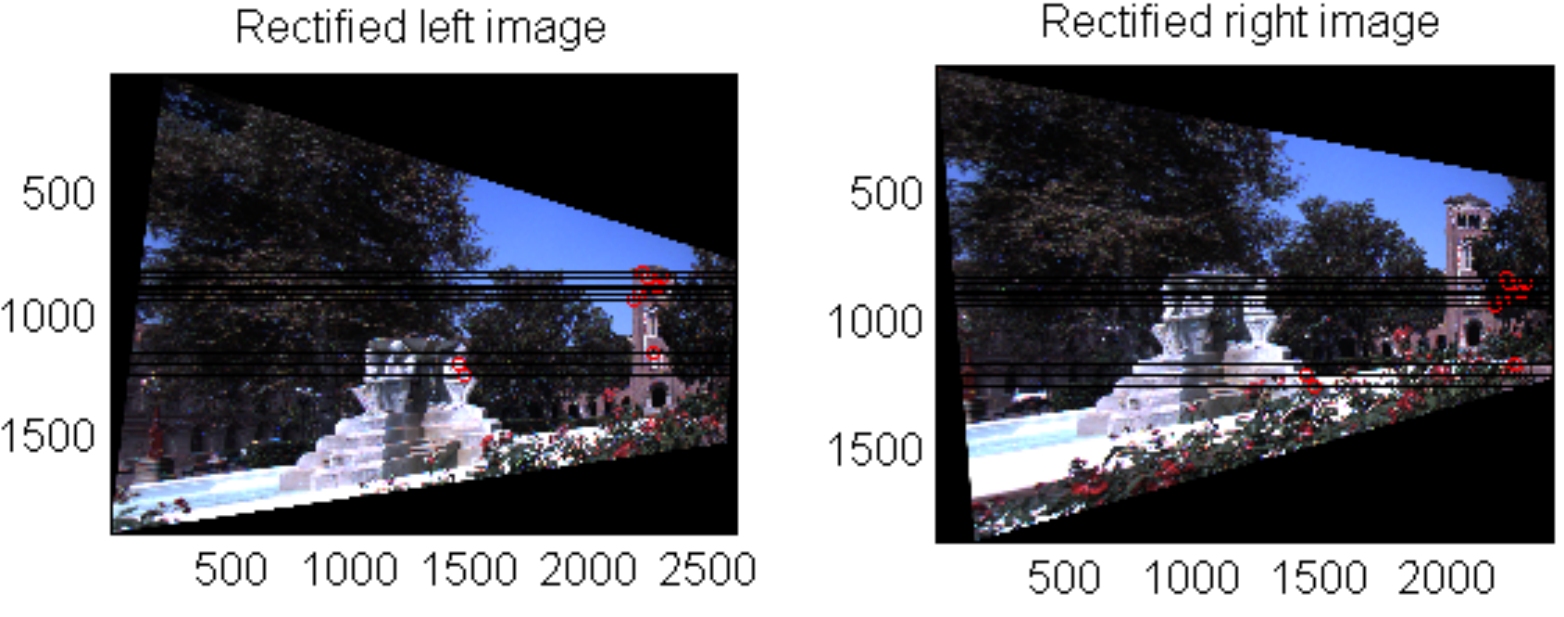}
                \caption{Hartley ($E_v$=0.214)}
        \end{subfigure}
		\quad
        \begin{subfigure}[t]{0.29\textwidth}
                \centering
                \includegraphics[width=\textwidth]{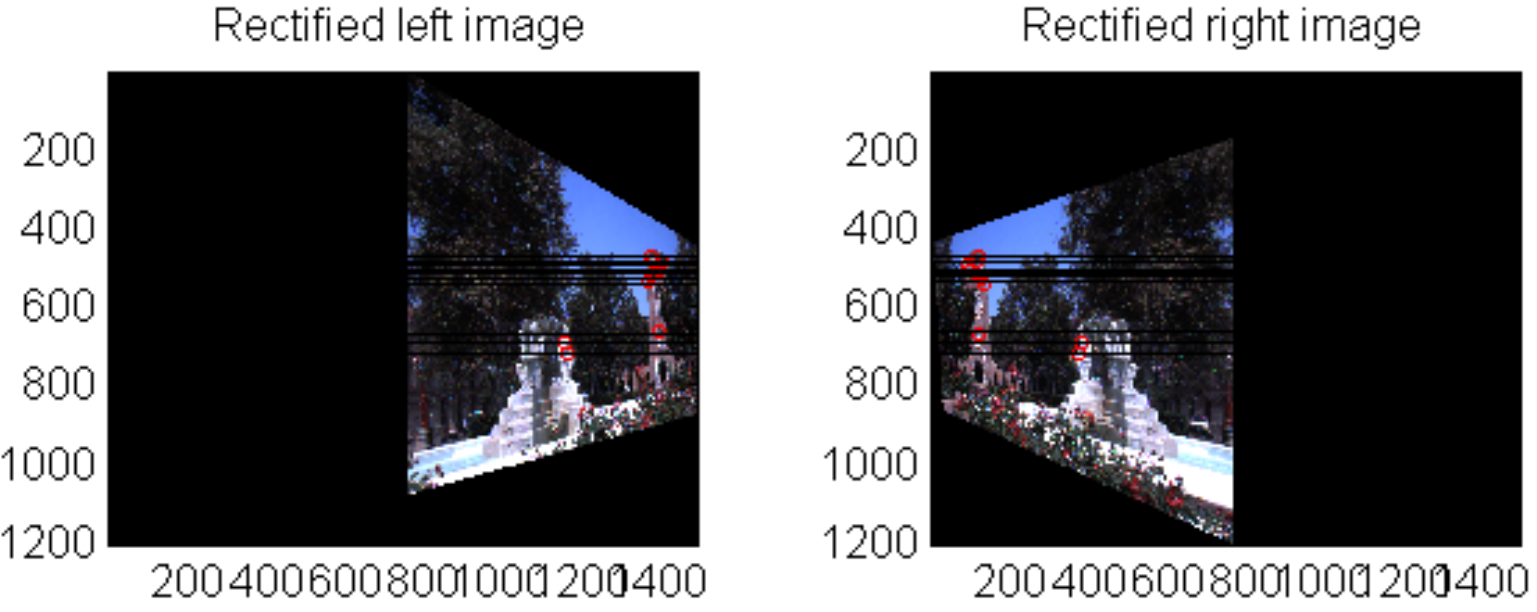}
                \caption{Mallon ($E_v$=5.484)}
        \end{subfigure}
		\\
        \begin{subfigure}[t]{0.29\textwidth}
                \centering
                \includegraphics[width=\textwidth]{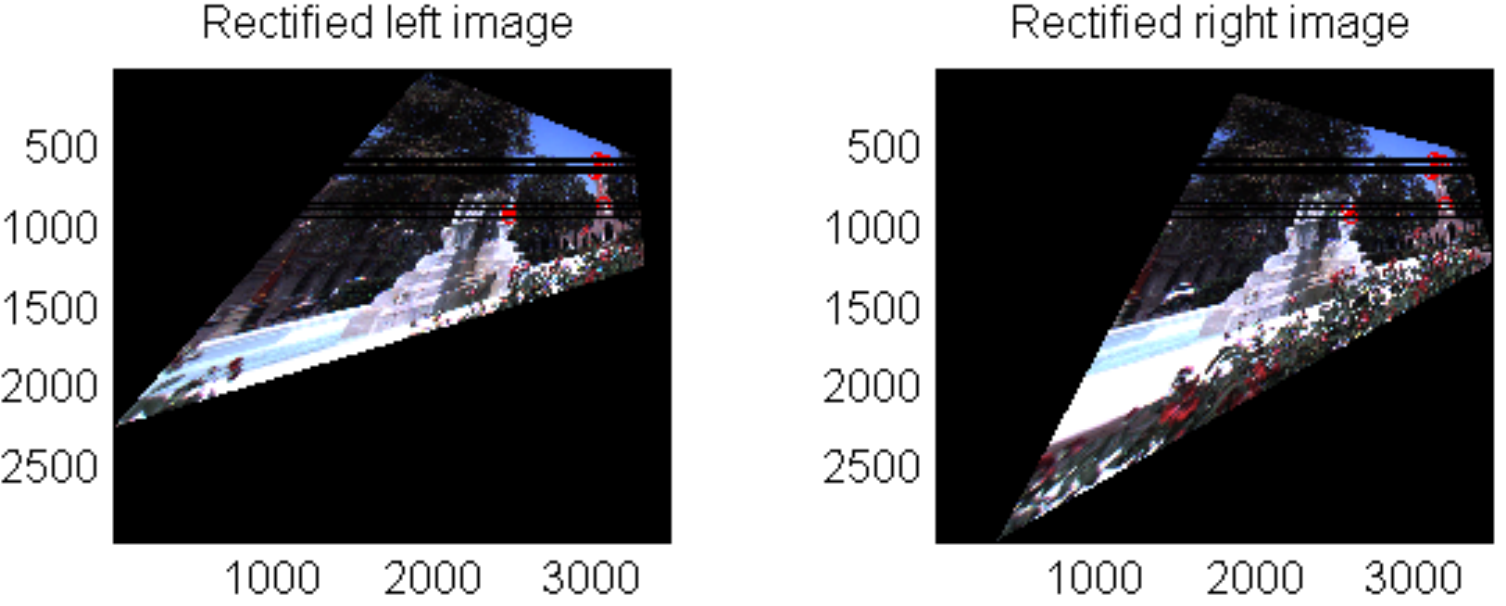}
                \caption{Wu ($E_v$=4.520)}
        \end{subfigure}
		\quad
        \begin{subfigure}[t]{0.29\textwidth}
                \centering
                \includegraphics[width=\textwidth]{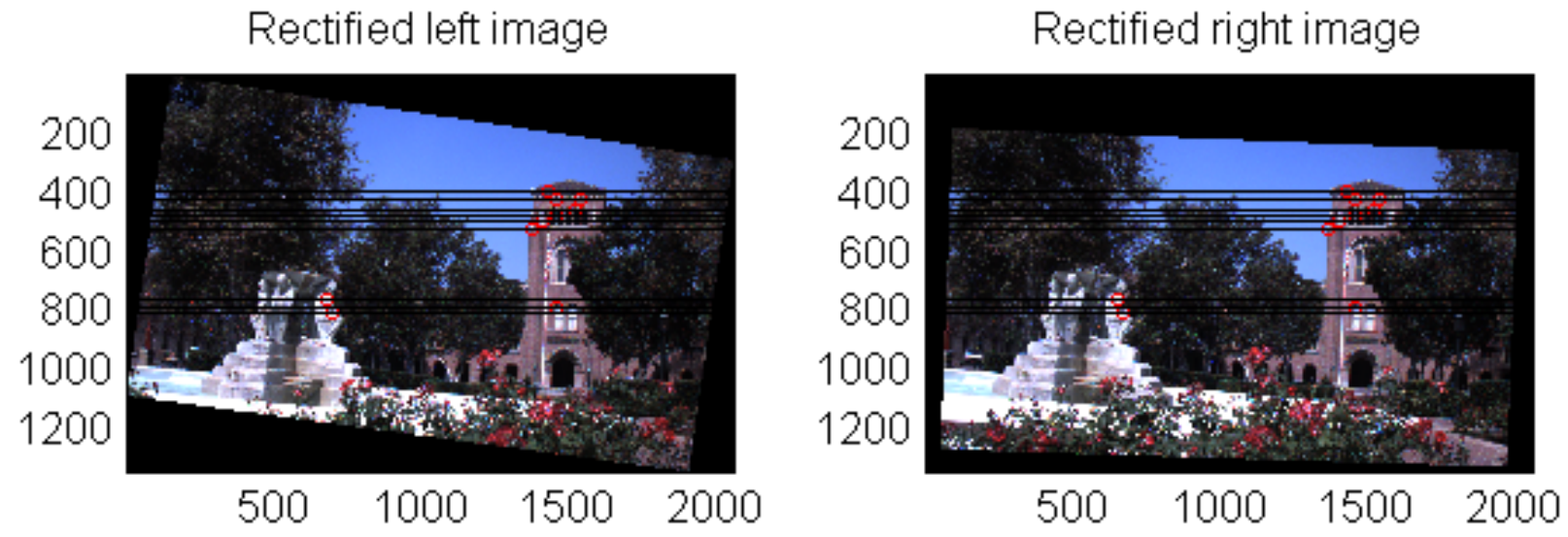}
                \caption{Fuesillo ($E_v$=0.120)}
        \end{subfigure}
		\quad
        \begin{subfigure}[t]{0.29\textwidth}
                \centering
                \includegraphics[width=\textwidth]{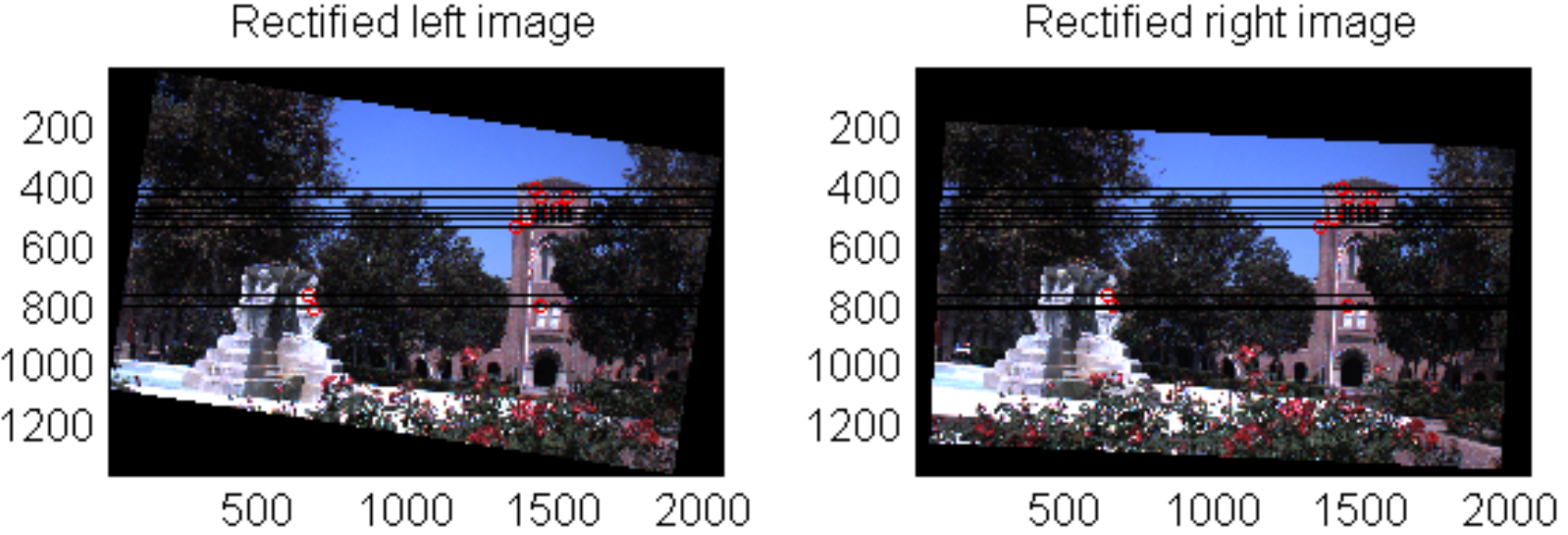}
                \caption{USR-CGD ($E_v$=0.101)}
        \end{subfigure}
\caption{Comparison of subjective quality and rectification errors for 
for Fountain3 in the MCL-RS database.}\label{fig:subjectiveresult2}
\end{figure*}
%%%%%%%%%%%%%%%%%%%%%%%%%%%%%%%%%%%%%%%%%%%%%%%%%%%%%%%%%%%%%%%%%

%%%%%%%%%%%%%%%%%%%%%%%%%%%%%%%%%%%%%%%%%%%%%%%%%%%%%%%%%%%%%%%%%
\begin{figure*}[t]
        \centering
	 \begin{subfigure}[t]{0.29\textwidth}
                \centering
                \includegraphics[width=\textwidth]{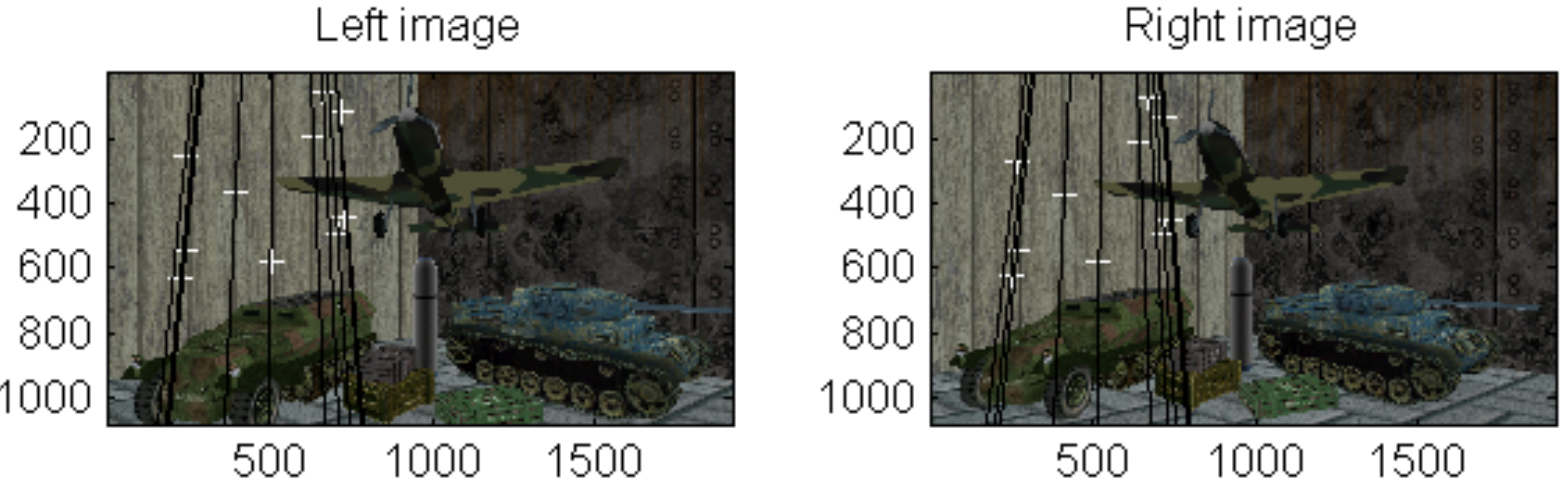}
                \caption{Original unrectified image pair}
        \end{subfigure}
		\quad
        \begin{subfigure}[t]{0.29\textwidth}
                \centering
                \includegraphics[width=\textwidth]{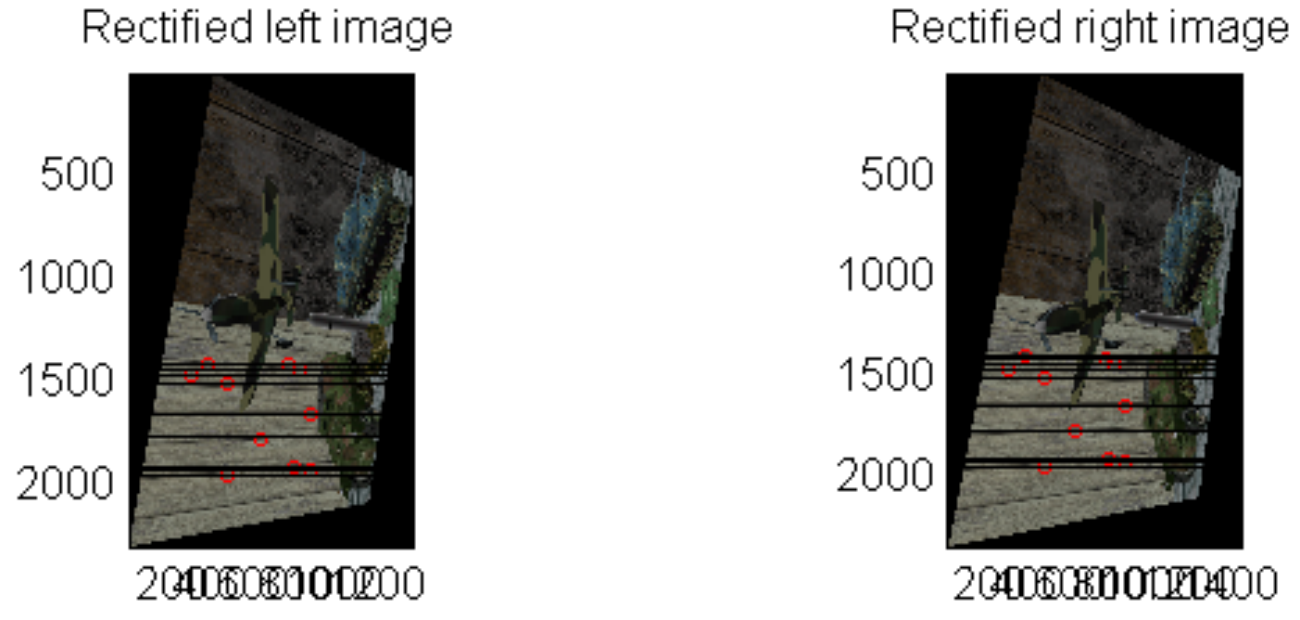}
                \caption{Hartley ($E_v$=0.061)}
        \end{subfigure}
		\quad
        \begin{subfigure}[t]{0.29\textwidth}
                \centering
                \includegraphics[width=\textwidth]{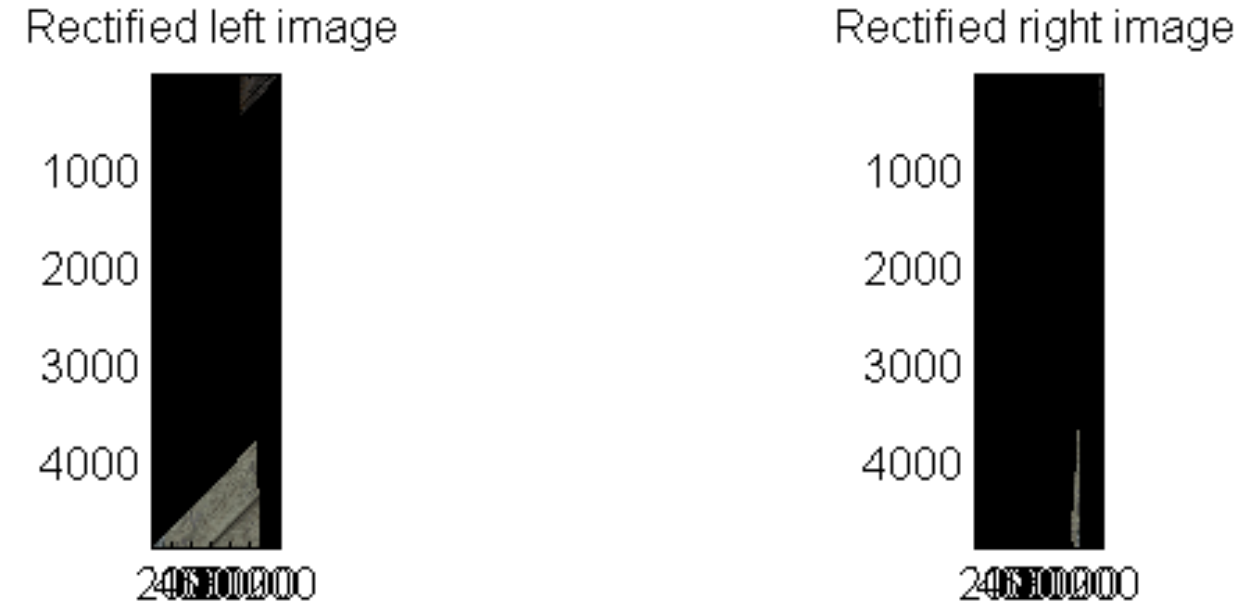}
                \caption{Mallon ($E_v$=625.942)}
        \end{subfigure}
		\\
        \begin{subfigure}[t]{0.29\textwidth}
                \centering
                \includegraphics[width=\textwidth]{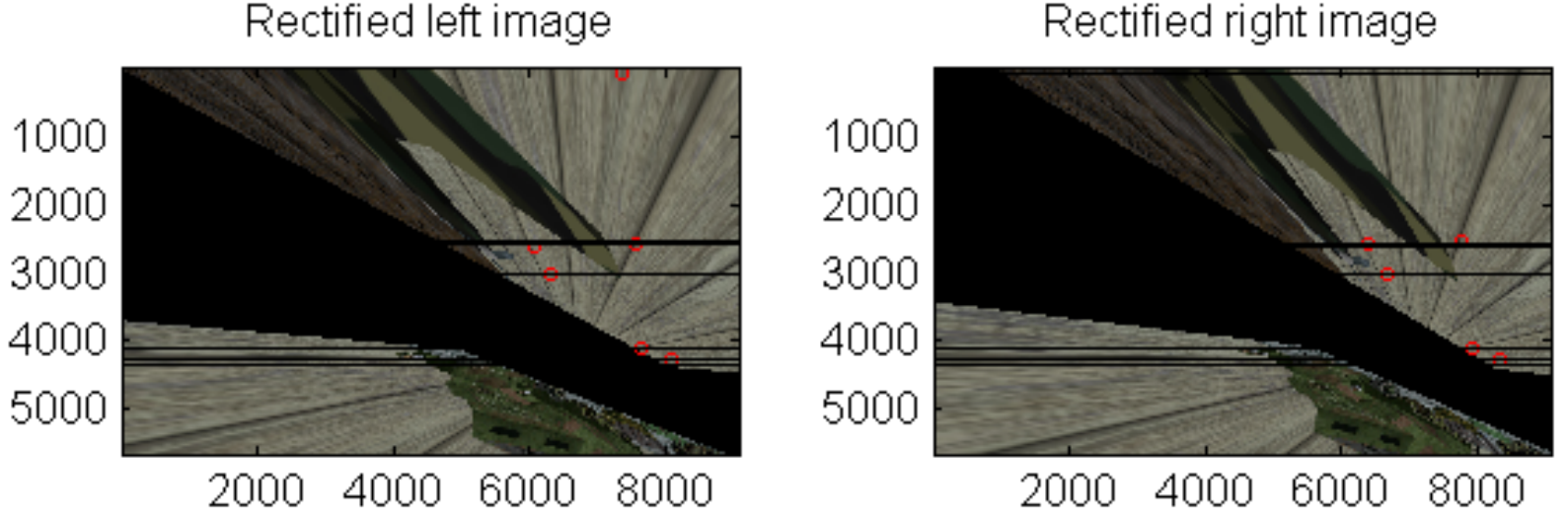}
                \caption{Wu ($E_v$=99.432)}
        \end{subfigure}
		\quad
        \begin{subfigure}[t]{0.29\textwidth}
                \centering
                \includegraphics[width=\textwidth]{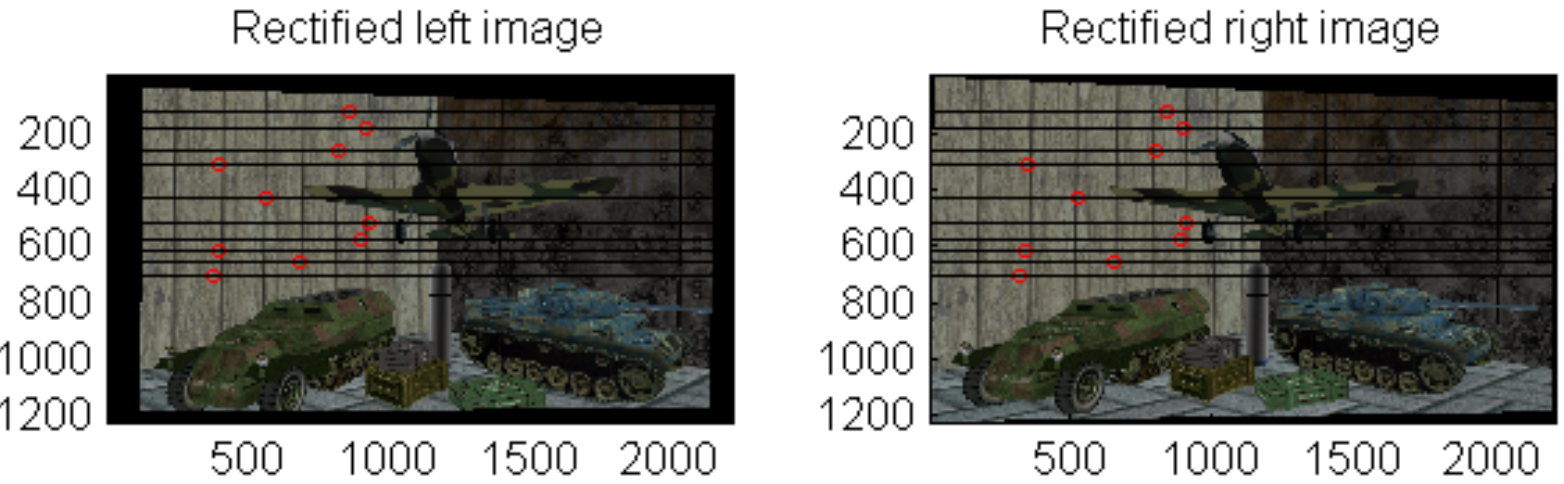}
                \caption{Fuesillo ($E_v$=0.133)}
        \end{subfigure}
		\quad
        \begin{subfigure}[t]{0.29\textwidth}
                \centering
                \includegraphics[width=\textwidth]{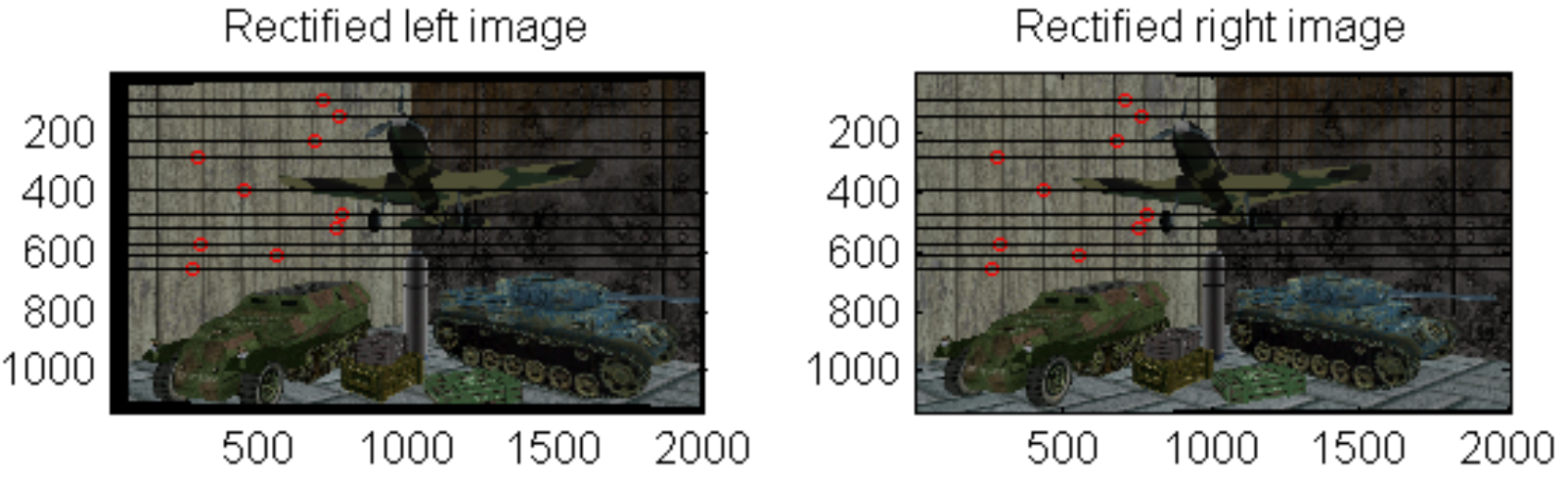}
                \caption{USR-CGD ($E_v$=0.127)}
        \end{subfigure}
\caption{Comparison of subjective quality and rectification errors for 
Military in the MCL-SS database.}\label{fig:subjectiveresult3}
\end{figure*}
%%%%%%%%%%%%%%%%%%%%%%%%%%%%%%%%%%%%%%%%%%%%%%%%%%%%%%%%%%%%%%%%%

%%%%%%%%%%%%%%%%%%%%%%%%%%%%%%%%%%%%%%%%%%%%%%%%%%%%%%%%%%%%%%%%%
\begin{figure*}[t]
        \centering
	 \begin{subfigure}[t]{0.29\textwidth}
                \centering
                \includegraphics[width=\textwidth]{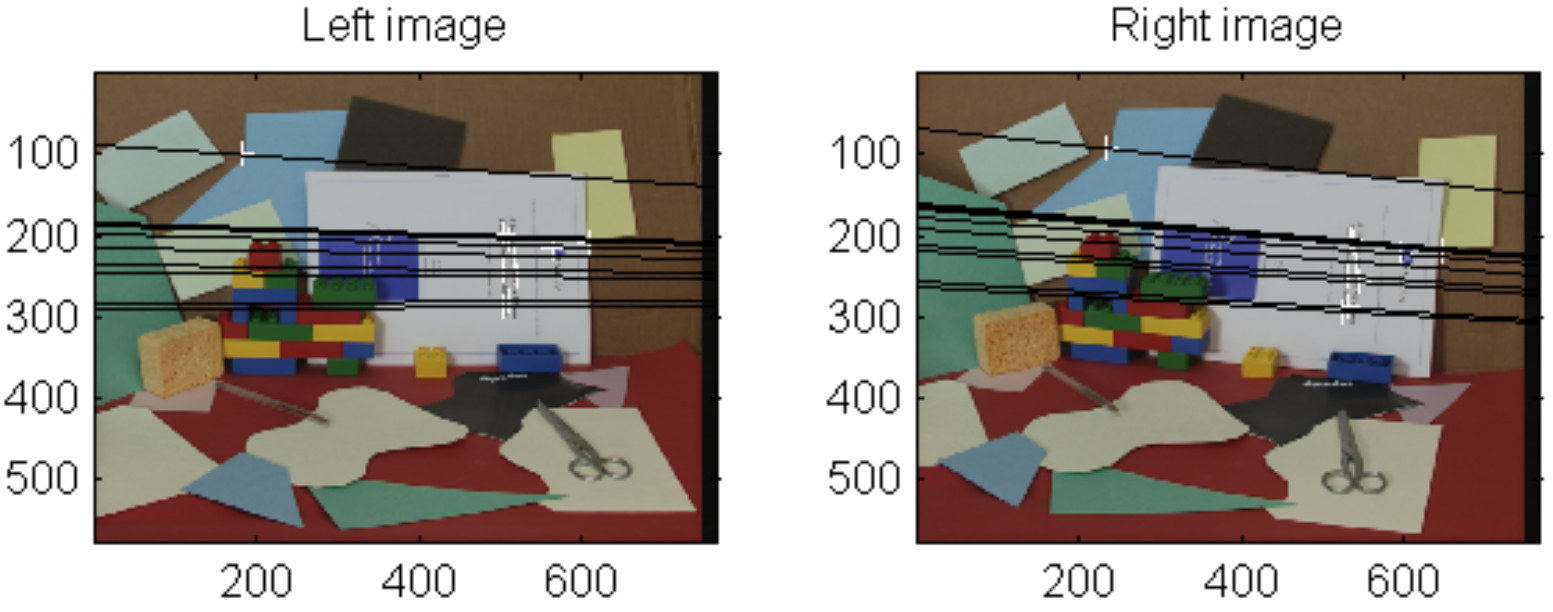}
                \caption{Original unrectified image pair}
        \end{subfigure}
		\quad
        \begin{subfigure}[t]{0.29\textwidth}
                \centering
                \includegraphics[width=\textwidth]{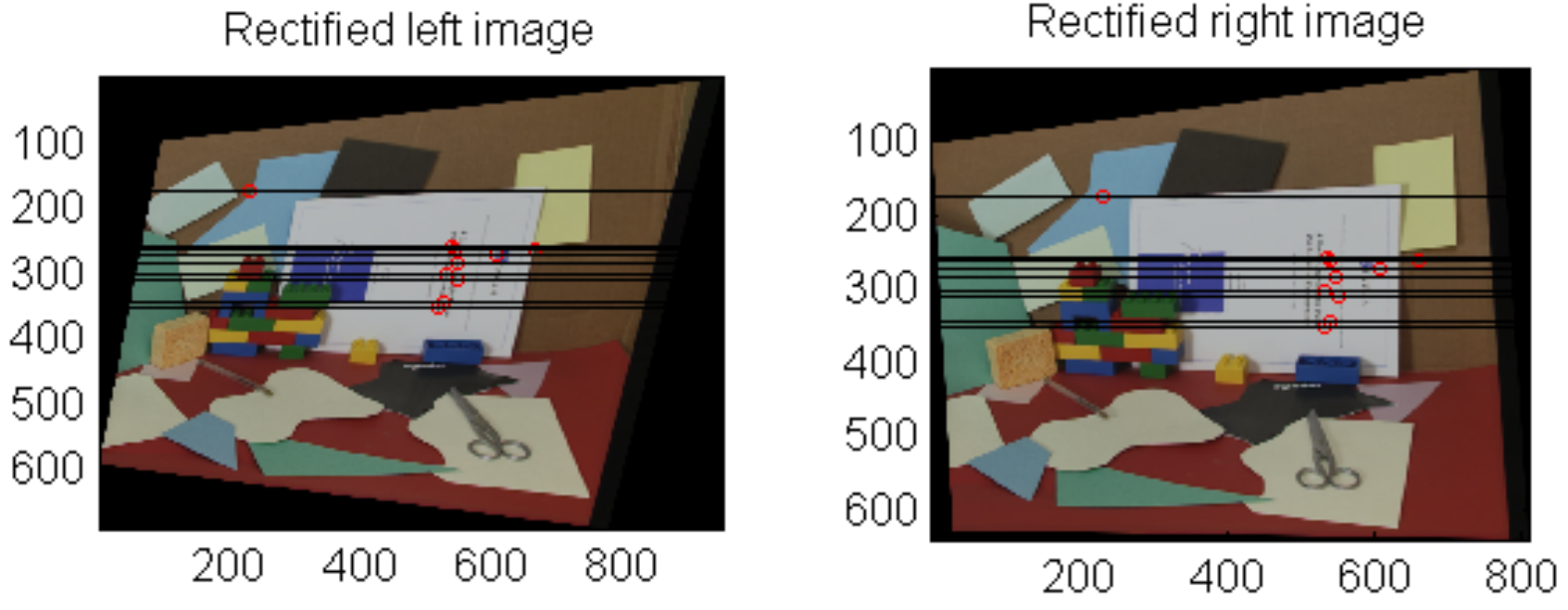}
                \caption{Hartley ($E_v$=0.141)}
        \end{subfigure}
		\quad
        \begin{subfigure}[t]{0.29\textwidth}
                \centering
                \includegraphics[width=\textwidth]{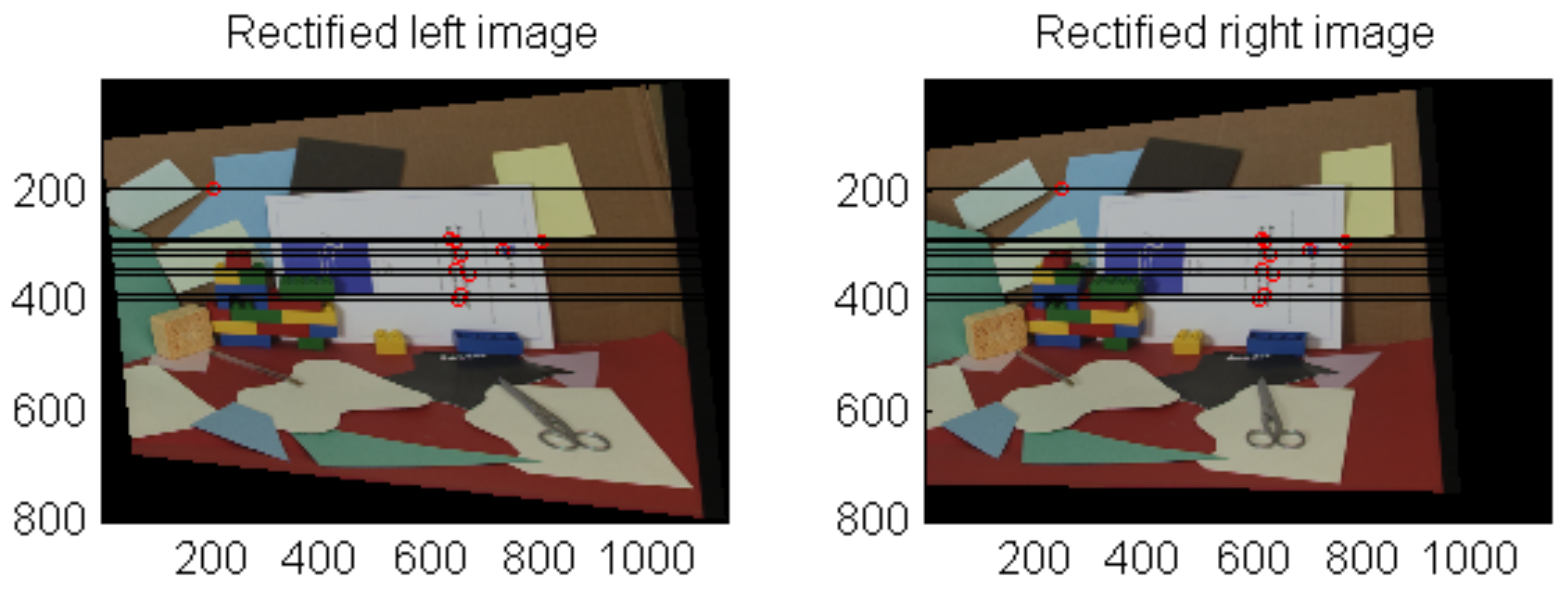}
                \caption{Mallon ($E_v$=0.796)}
        \end{subfigure}
		\\
        \begin{subfigure}[t]{0.29\textwidth}
                \centering
                \includegraphics[width=\textwidth]{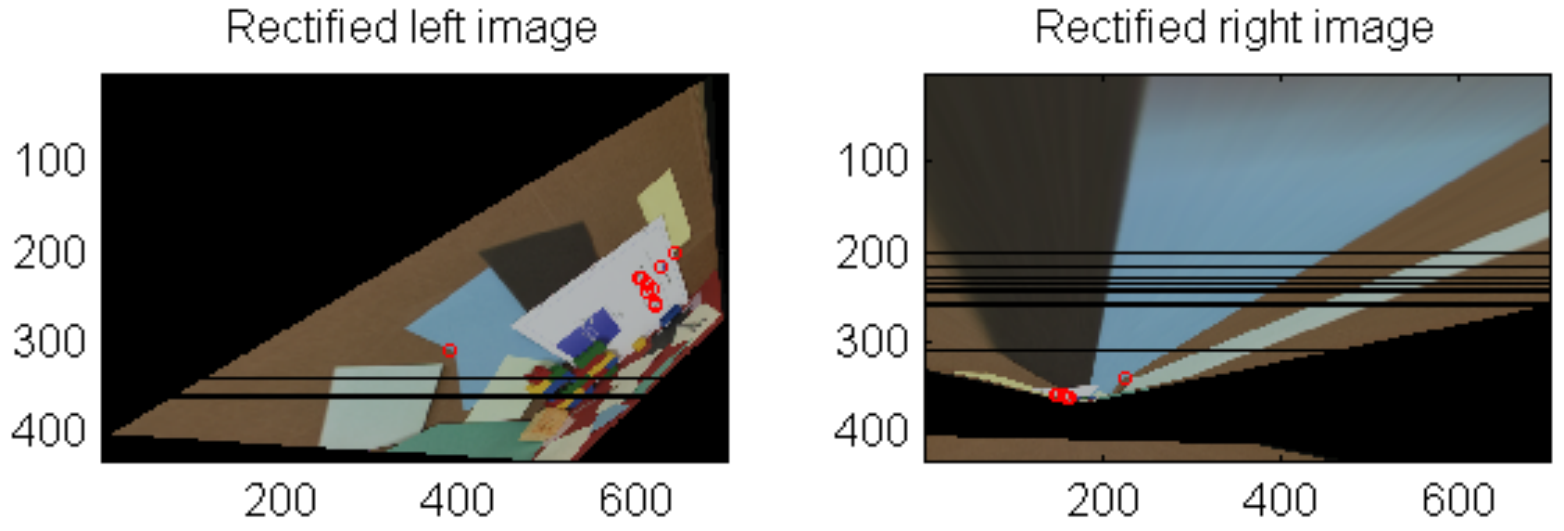}
                \caption{Wu ($E_v$=0.607)}
        \end{subfigure}
		\quad
        \begin{subfigure}[t]{0.29\textwidth}
                \centering
                \includegraphics[width=\textwidth]{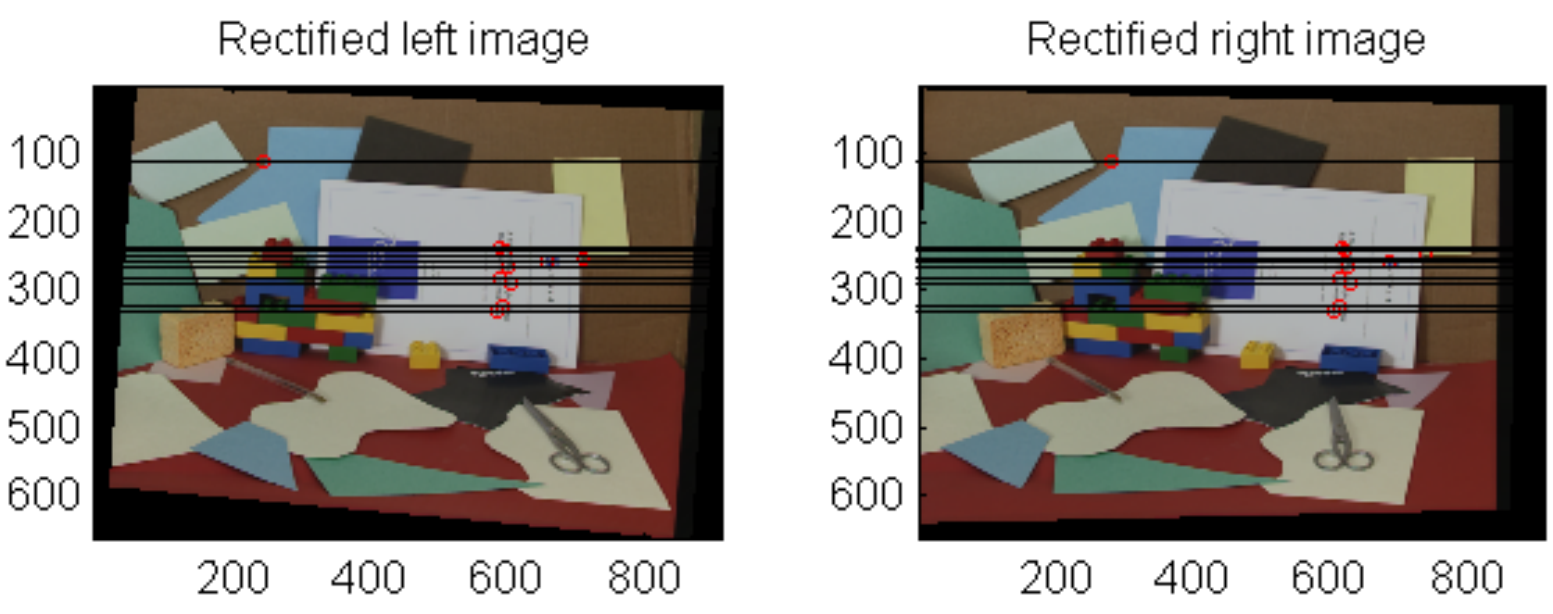}
                \caption{Fuesillo ($E_v$=0.240)}
        \end{subfigure}
		\quad
        \begin{subfigure}[t]{0.29\textwidth}
                \centering
                \includegraphics[width=\textwidth]{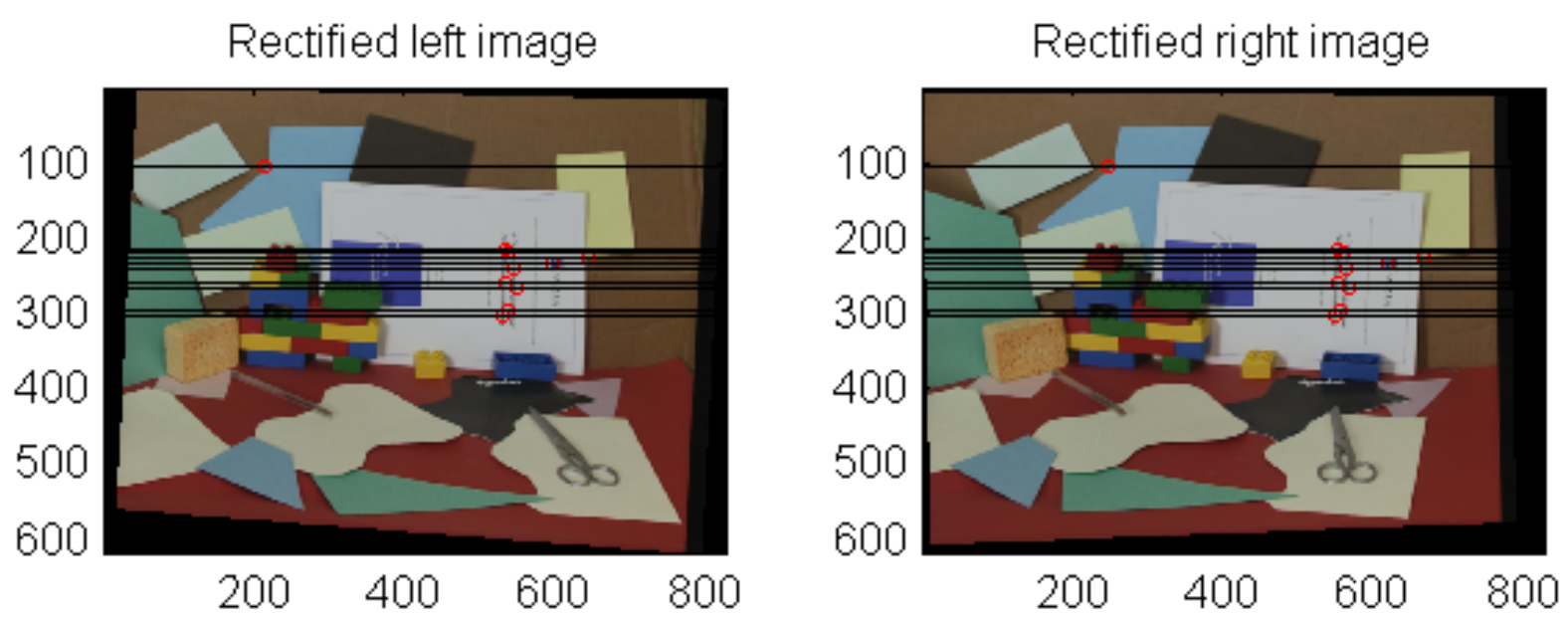}
                \caption{USR-CGD ($E_v$=0.308)}
        \end{subfigure}
\caption{Comparison of subjective quality and rectification errors for 
Tot in the SYNTIM database.}\label{fig:subjectiveresult4}
\end{figure*}
%%%%%%%%%%%%%%%%%%%%%%%%%%%%%%%%%%%%%%%%%%%%%%%%%%%%%%%%%%%%%%%%%

%%%%%%%%%%%%%%%%%%%%%%%%%%%%%%%%%%%%%%%%%%%%%%%%%%%%%%%%%%%%%%%%%
\begin{figure*}[t]
        \centering
	 \begin{subfigure}[t]{0.29\textwidth}
                \centering
                \includegraphics[width=\textwidth]{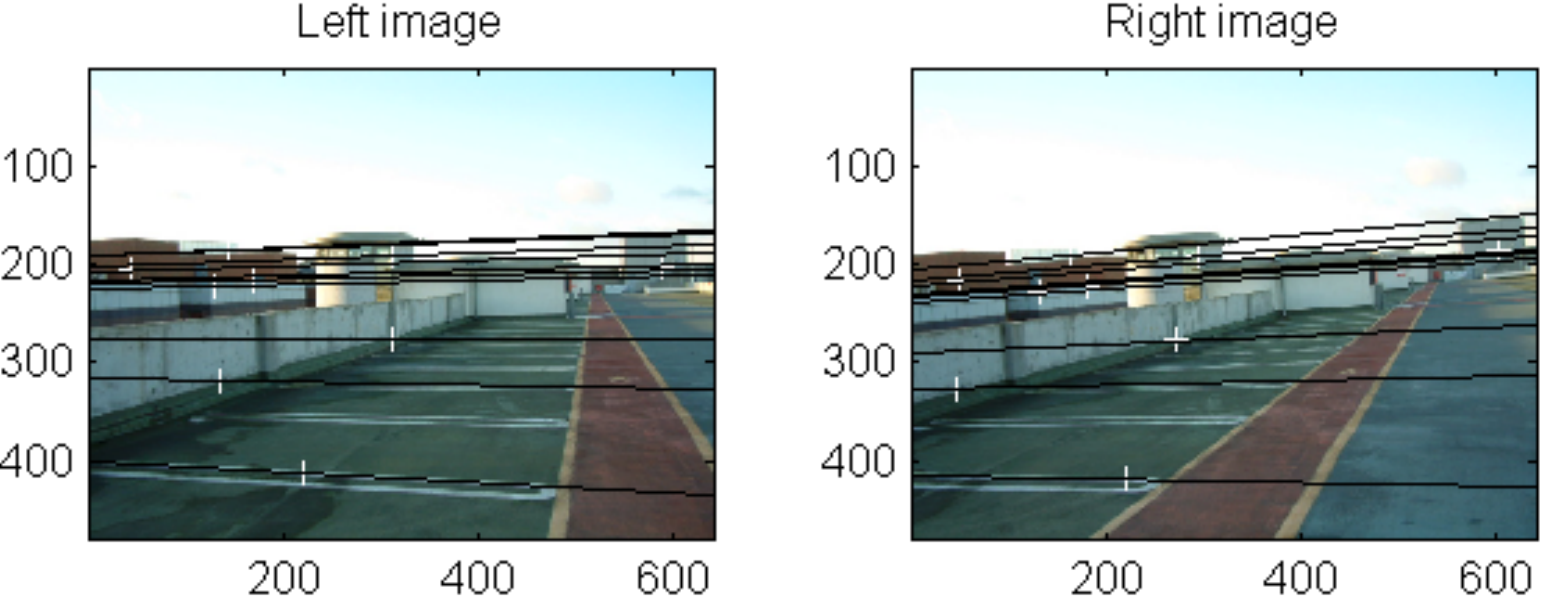}
                \caption{Original unrectified image pair}
        \end{subfigure}
		\quad
        \begin{subfigure}[t]{0.29\textwidth}
                \centering
                \includegraphics[width=\textwidth]{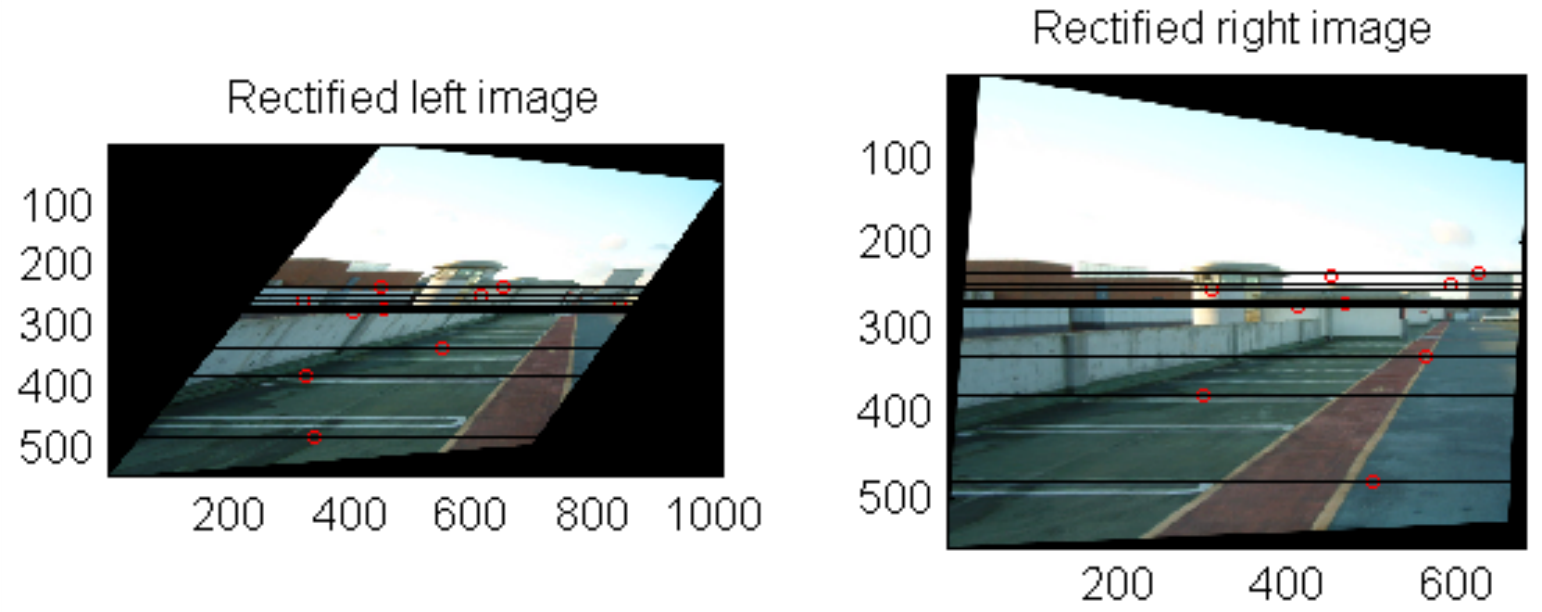}
                \caption{Hartley ($E_v$=0.108)}
        \end{subfigure}
		\quad
        \begin{subfigure}[t]{0.29\textwidth}
                \centering
                \includegraphics[width=\textwidth]{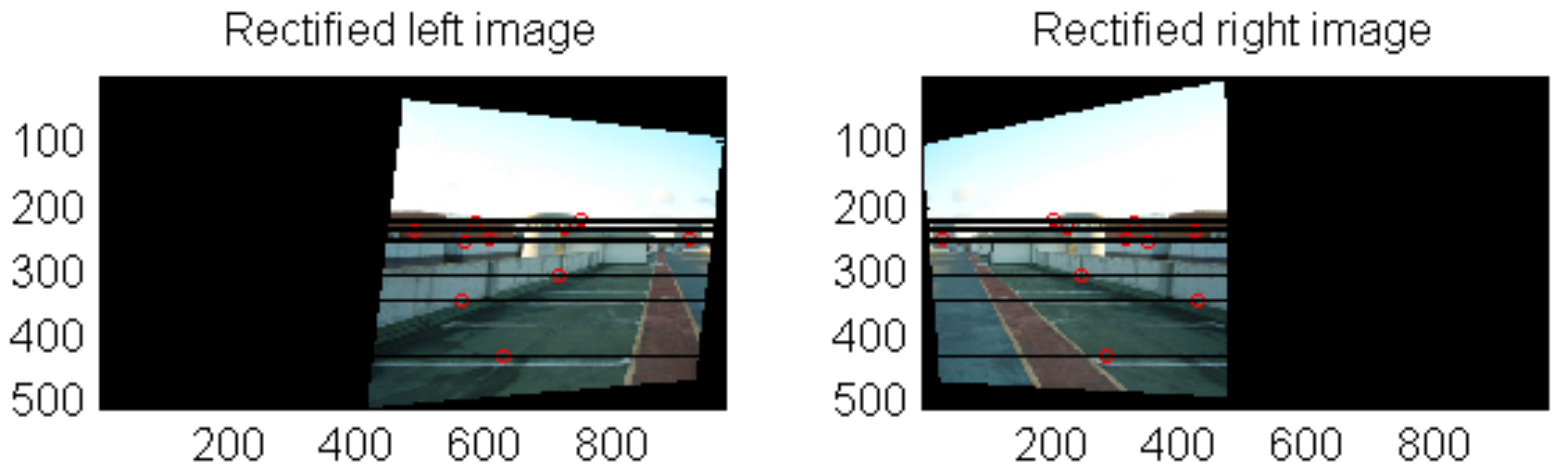}
                \caption{Mallon ($E_v$=0.127)}
        \end{subfigure}
		\\
        \begin{subfigure}[t]{0.29\textwidth}
                \centering
                \includegraphics[width=\textwidth]{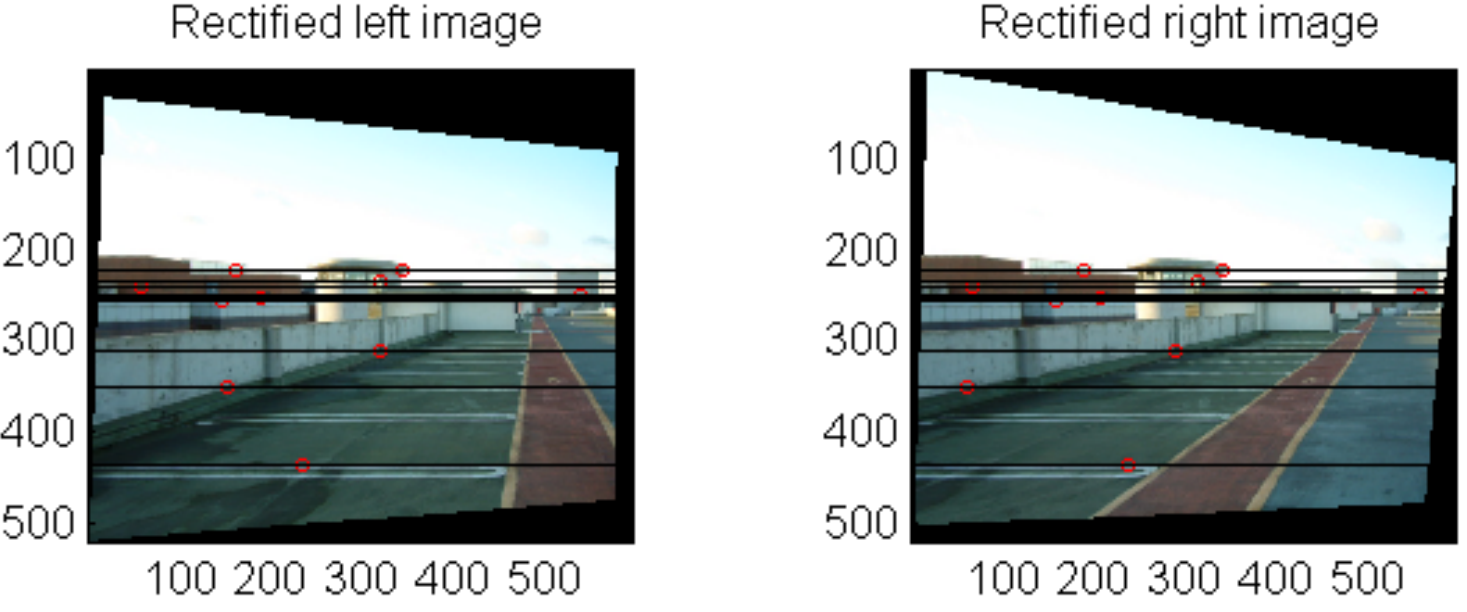}
                \caption{Wu ($E_v$=0.091)}
        \end{subfigure}
		\quad
        \begin{subfigure}[t]{0.29\textwidth}
                \centering
                \includegraphics[width=\textwidth]{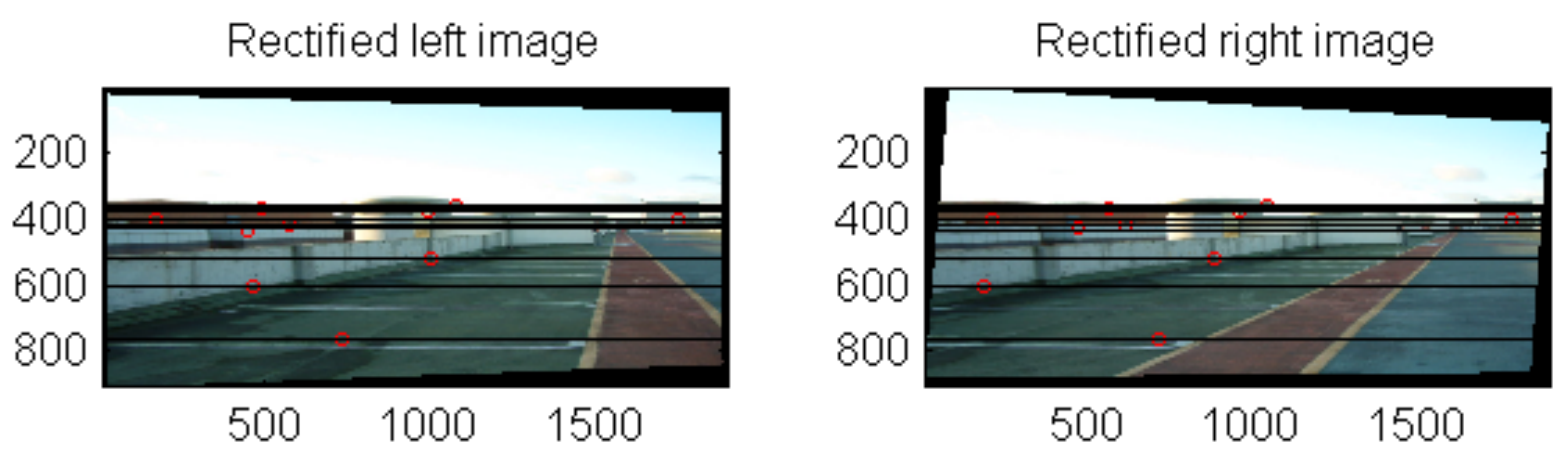}
                \caption{Fuesillo ($E_v$=1.068)}
        \end{subfigure}
		\quad
        \begin{subfigure}[t]{0.29\textwidth}
                \centering
                \includegraphics[width=\textwidth]{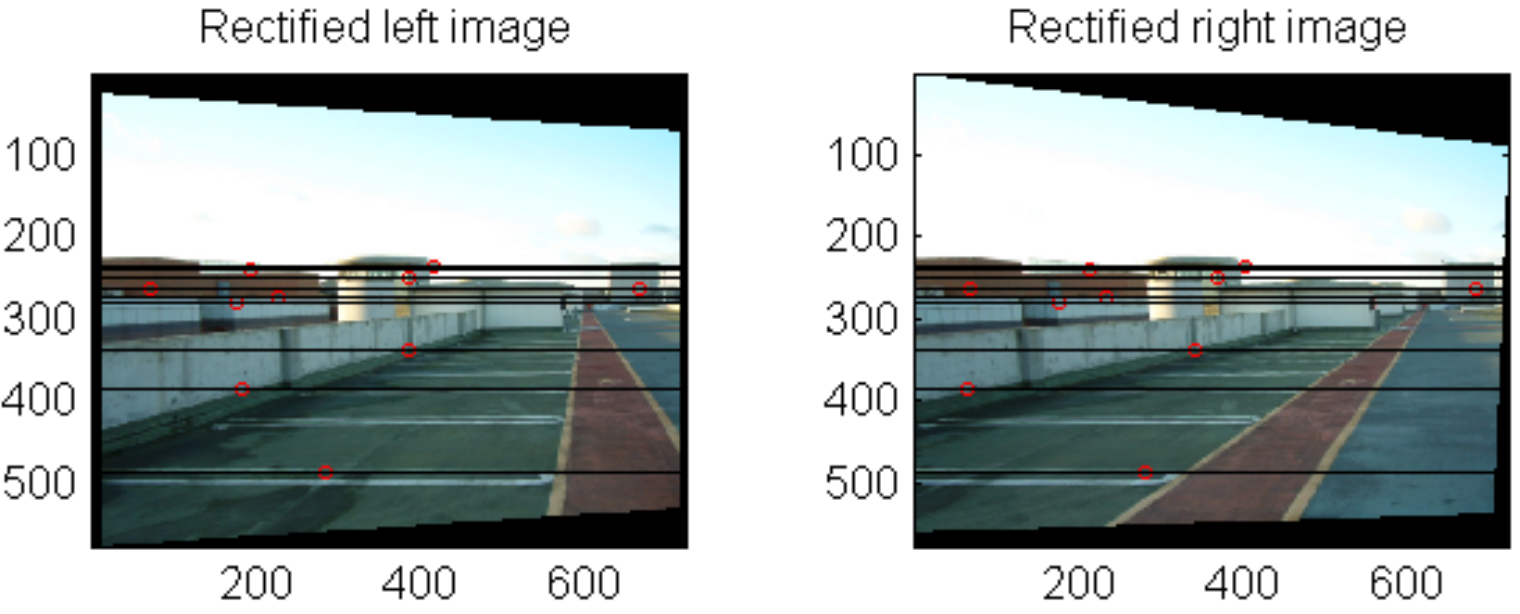}
                \caption{USR-CGD ($E_v$=0.282)}
        \end{subfigure}
\caption{Comparison of subjective quality and rectification errors for 
Root in the VSG database.}\label{fig:subjectiveresult5}
\end{figure*}
%%%%%%%%%%%%%%%%%%%%%%%%%%%%%%%%%%%%%%%%%%%%%%%%%%%%%%%%%%%%%%%%%

%%%%%%%%%%%%%%%%%%%%%%%%%%%%%%%%%%%%%%%%%%%%%%%%%%%%%%%%%%%%%%%%%
%%%%%%%%%%%%%%%%%%%%%%%%%%%%%%%%%%%%%%%%%%%%%%%%%%%%%%%%%%%%%%%%%

\section{Conclusion and Future Work}\label{sec:conclusions}

A new rectification algorithm, called USR-CGD, was proposed for
uncalibrated stereo images in this work. It adopts a generalized
homography model to reduce the rectification error and incorporates
several practical geometric distortions in the cost function as
regularization terms, which prevent severe perspective distortions in
rectified images. It was shown by experimental results that the
proposed USR-CGD algorithm outperforms existing algorithms in both
objective and subjective quality measures.  In the future, we would like
to study the stereo matching problem for depth estimation based on the
current work on uncalibrated stero image rectification. 

\section{Acknowledgements}\label{sec:Ack}

This work was funded by SAIT (Samsung Advanced Institute of Technology)
and supported in part by the University of Southern California Center
for High-Performance Computing and Communications.  The use of Matlab
functions by Dr. A. Fusiello, SYNTIM (INRIA, Rocquencourt) images
(http://www-rocq.inria.fr/$\sim$tarel/syntim/paires.html) and VSG
(Vision Systems Group, Dublin CityUniversity) images
(http://www.vsg.dcu.ie/code.html) is gratefully appreciated.


\begin{thebibliography}{99}

\bibitem{cit:Ayache1991}
N. Ayache, and L. Francis, ``Trinocular stereo vision for robotics.'' 
 in {\it IEEE Transactions on Pattern Analysis and Machine Intelligence},
vol. 13, no. 1, pp 73-85, 1991.

\bibitem{cit:Fusiello2000}
A. Fusiello, T. Emanuele, and V. Alessandro, ``A compact algorithm 
for rectification of stereo pairs.'' in {\it Machine Vision and Applications},
vol.12, no. 1, pp. 16-22, 2000.

\bibitem{cit:Hartley1999}
R. Hartley, ``Theory and practice of projective rectification.'' in {\it Internat. 
J. Comput. Vision}, vol. 35, no. 2, pp. 115-127, 1999.

\bibitem{cit:Hartley2003}
R. Hartley, and Z. Andrew, ``Multiple view geometry in computer vision.''
in {\it Cambridge university press}, 2003.

\bibitem{cit:Loop1999}
C. Loop, and Z. Zhang, ``Computing rectifying homographies for stereo
vision.'' in {\it Proceedings of the IEEE Conference on Computer
Vision and Pattern Recognition}, vol. 1, pp. 125–131, Jun, 1999.

\bibitem{cit:Pollefeys1999}
M. Pollefeys, K. Reinhard, and G. Luc, ``A simple and efficient rectification 
method for general motion.'' in {\it Proceedings of the IEEE Conference on Computer
Vision and Pattern Recognition}, Vol. 1. pp. 496-501, 1999.

\bibitem{cit:Gluckman2001}
J. Gluckman, and S. Nayar. ``Rectifying transformations that minimize 
resampling effects'' in {\it Proceedings of the IEEE Conference on Computer
Vision and Pattern Recognition}, vol. 1,  pp. 111–117, Dec, 2001.

\bibitem{cit:Mallon2005}
J. Mallon, and P. Whelan, ``Projective rectification from the fundamental
matrix.'' in {\it Image and Vision Computing}, vol. 23, no.7, pp. 643-650, 2005.
(VSG images can be downloadable at: http://www.vsg.dcu.ie/code.html)

\bibitem{cit:Isgrò1999}
F. Isgrò, and T. Emanuele, ``On robust rectification for uncalibrated images.''
in {\it IEEE International Conference on Image Analysis and Processing}, 
pp. 297-302, 1999.

\bibitem{cit:Wu2005}
H. Wu, and Y. Yu. ``Projective rectification with reduced geometric 
distortion for stereo vision and stereoscopic video.'', in {\it Journal of 
Intelligent and Robotic Systems}, vol. 42, no. 1, pp. 71-94, 2005.

\bibitem{cit:Fusiello2011}
A. Fusiello, and I. Luca, ``Quasi-Euclidean epipolar rectification of 
uncalibrated images.'' in {\it Machine Vision and Applications}, vol. 22,
no. 4, pp. 663-670, 2011.

\bibitem{cit:Zilly2010}
F. Zilly, M. Müller, P. Eisert, and P. Kauff, ``Joint estimation of epipolar 
geometry and rectification parameters using point correspondences 
for stereoscopic TV sequences.'' in {\it Proceedings of 3DPVT},
Paris, France, 2010.

\bibitem{cit:Georgiev2013}
M. Georgiev, G. Atanas, and H. Miska, ``A fast and accurate re-calibration
technique for misaligned stereo cameras.'' in {\it IEEE International conference 
on Image Processing}, pp. 24-28, 2013.

\bibitem{cit:Fuageras1993}
O. Faugeras, ``Three-dimensional computer vision: a geometric viewpoint'', 
MIT press, Cambrigde, MA, 1993.

\bibitem{cit:Sampson}
Q. Luong, and D. Olivier, ``The fundamental matrix: Theory, algorithms, 
and stability analysis.'' in {\it International Journal of Computer Vision}
Vol. 17, No. 1, pp. 43-75, 1996.

\bibitem{cit:Yuan2010}
Y. Yuan, ``A review of trust region algorithms for optimization.'' in {\it ICIAM}, 
Vol. 99, 2000.

\bibitem{cit:Lowe2004}
G. Lowe, ``Distinctive image features from scale-invariant keypoints.'' in {\it 
International journal of computer vision}. Vol. 60, no. 2, pp. 91-110, 2004.

\bibitem{cit:Fischler1881}
M. Fischler, and R. Bolles, ``Random sample consensus: a paradigm for model 
fitting with applications to image analysis and automated cartography.'' in {\it
Communications of the ACM}, Vol. 24, no. 6, pp. 381-395, 1881.

\bibitem{cit:SYNTIM}
SYNTIM images can be downloadable at: 
``http://perso.lcpc.fr/tarel.jean-philippe/syntim/paires.html''

\bibitem{cit:MCLdatabase}
MCL-SS \& MCL-RS databases can be downloadable at :
http://mcl.usc.edu/mcl-ss-database/ and http://mcl.usc.edu/mcl-rs-database/

\end{thebibliography}
\end{document}